\documentstyle[11pt,epsf,ut-thesis,named,headerfooter,my-macros]{report}

\begin{document}

\pagenumbering{roman}
\pagestyle{empty}
\null
\vfill

\begin{singlespace}
\begin{centering}
{\huge A Formalism and an Algorithm for \\ 
\vspace*{0.5ex}
Computing Pragmatic Inferences and \\
\vspace*{0.7ex}
Detecting Infelicities} \\
\vspace*{2.5cm}
by \\
\vspace*{2.5cm}
{\LARGE Daniel Marcu} \\
\vspace*{4cm}
Department of Computer Science \\
University of Toronto \\
Toronto, Canada \\
September 1994 \\
\vspace*{1cm}
A thesis submitted in conformity with the requirements \\
for the degree of Master of Science \\
Graduate Department of Computer Science \\ 
University of Toronto \\
\vspace*{1.5cm}
Copyright \copyright\ 1994 by Daniel Marcu \\
\end{centering}
\end{singlespace}

\vfill

\newpage
\pagestyle{plain}
\null
\vfill
\begin{center}\Large\bf Abstract \end{center}

\noindent Since Austin introduced the term {\em infelicity}, the
linguistic literature has been flooded with its use.  Today, not only
performatives that fail are considered infelicitous but also
utterances that are syntactically, semantically, or pragmatically
ill-formed. However, no formal or computational explanation has been
given for infelicity. This thesis provides one for those
infelicities that occur when a pragmatic inference is cancelled.

We exploit a well-known difference between pragmatic and semantic
information: since implicatures and presuppositions, i.e., the
carriers of pragmatic information, are not specifically uttered,
pragmatic inferences are defeasible, while most of semantic inferences
are indefeasible. Our contribution assumes the existence of a finer
grained taxonomy with respect to pragmatic inferences. It is shown
that if one wants to account for the natural language expressiveness,
she should distinguish between pragmatic inferences that are
felicitous to defeat and pragmatic inferences that are infelicitously
defeasible. Thus, it is shown that one should consider at least three
types of information: indefeasible, felicitously defeasible, and
infelicitously defeasible. The cancellation of the last of these 
determines the pragmatic infelicities.

A new formalism has been devised to accommodate the three levels of
information, called {\em stratified logic}. Within it, we are able to
express formally notions such as {\em utterance $u$ presupposes $p$}
or {\em utterance $u$ is infelicitous}. Special attention is paid to
the implications that our work has in solving some well-known
existential philosophical puzzles.  The formalism yields an algorithm
for computing interpretations for utterances, for determining their
associated presuppositions, and for signalling infelicities. Its
implementation is a Lisp program that takes as input a set of
stratified formulas that constitute the necessary semantic and
pragmatic knowledge and the logical translation of an utterance or set
of utterances and that computes a set of {\em optimistic}
interpretations for the given utterances. The program computes for
each set of utterances the associated presuppositions and signals when
an infelicitous sentence has been uttered.

\vfill

\newpage
\vspace*{2cm}
\begin{center}\Large\bf Acknowledgments \end{center}

%\begin{singlespace}

\noindent First, I thank my supervisor, Graeme Hirst, for his
competent guidance, patience, and humor in explaining and re-explaining so
many things; and confidence that I can delineate and solve problems on
my own. I also thank him for the way he taught me to weight and glue
words into sentences, sentences into paragraphs, to put down ideas
in a way that other people can understand them. The flaws in this thesis
show that I still have a long way to go until I come to know well
everything he taught me.

I want to thank Hector Levesque, my second reader, for the helpful
comments he gave me. I thank Hector Levesque and Ray Reiter for
teaching me how to tackle a logical problem and how to find formal
explanations for the phenomena around us. I thank Jeff Siskind for
reminding me that the ultimate proof is a program that {\em does}
something.

I thank my colleagues in the natural language, cognitive robotics, and
knowledge representation groups who shared with me their questions,
problems, opinions, and thoughts.

I am grateful for the financial support I have received from the
University of Toronto and from the Natural Sciences and Engineering
Research Council of Canada.

Because a thesis is not only words, programs, or formulas I thank Ed
Klajman for being the friend with whom I was able to get into a
world free of syntactic, semantic, or pragmatic constraints. 

I thank my parents for everything they have taught me, for the
understanding and trust they have shown me every day since I
have started this work.
 
But most of all, I want to thank my wife, Oana, for her endless
love, support, and strength in believing that one day, we will be again
together, in a better world, every single minute.

%\end{singlespace}

\vfill

\newpage
\tableofcontents
\newpage
\listoffigures

\newpage
\pagenumbering{arabic}

\chapter{Pragmatic inferences, infelicities, and logic} 
\label{relevant-research}

Much of the research done in linguistic pragmatics makes extensive use
of the term {\em infelicity}. However, it seems that no approach
offers a formal or computational definition for this phenomenon. It is
the goal of this thesis to provide one. Unfortunately, the scope of
the phenomenon of infelicity is extremely large and not well-defined.
Apparently, the word was introduced by Austin~\shortcite{austin62} to
characterize performative utterances that fail.  But since then, it
has become fashionable to characterize anything that goes wrong as
infelicitous.  If a sentence is not syntactically or semantically
well-formed it is often said to be infelicitous; if an utterance is
not pragmatically well-formed it is said to be infelicitous too.  This
thesis is about pragmatic infelicities: those that occur when some
Gricean inferences are cancelled. 

In this chapter we review the relevant work in implicature and
presupposition from a perspective concerned primarily with aspects of
felicity.  The aim of the first two sections is to familiarize the
reader with the vocabulary and to give her a {\em feeling} of what
pragmatic infelicities are. The third section analyzes pragmatic
inferences in detail and shows why it is inappropriate to equate
infelicity to inconsistency. The last section studies representative
approaches to default reasoning and shows that none of them is able to
formalize adequately the infelicity phenomenon.

In order to capture pragmatic infelicities, we introduce a new
formalism called {\em stratified logic}. Chapter 2 contains the informal
intuitions that lie behind this new mathematical tool, and a
characterization of its syntax and semantics for the propositional and
first-order case.

Chapter 3 sheds some light on a fundamental difference between the
notions of lexical semantics and lexical pragmatics. It is shown that
an appropriate semantic and pragmatic characterization of the lexical
items and syntactical constructs in terms of stratified logic not only
provides a formal characterization for pragmatic infelicities but also
an adequate explanation for cancellations that implicatures and
presuppositions are subject to. Chapter 4 extends this work and shows
how the solution developed for simple utterances is appropriate
without any modifications for handling complex utterances. Thus, we
obtain a formalism and a computational method that signal infelicitous
utterances and determine the presuppositions that are associated with
an utterance independent of its complexity. Our method works even for
sequences of utterances, a case that is omitted by previous research in
the area.

The existential presuppositions carried by definite references are one
of the most problematic linguistic and philosophical issues. Chapter 5
is dedicated to a review of the work done in this area. It is shown
that most of the philosophical approaches are more concerned in
dealing with nonexistence and they fail to capture the existential
presuppositions, while most of the linguistic approaches are incapable
of accommodating nonexistent entities. We use a rich ontology \`{a} la
Hirst and a set of methodological principles that embed the essence of
Meinong's philosophy and Grice's conversational principles into a
stratified logic, under an unrestricted interpretation of the
quantifiers. The result is a logical formalism that yields a tractable
computational method that uniformly calculates all the presuppositions
of a given utterance, including the existential ones. A comparison
with Parsons's, Russell's, and Hobbs's work emphasizes the superiority
of our approach.

Chapter 6 is concerned with some implementation issues. Two annexes
are provided: one that contains the Lisp code of our implementation,
and one that contains the results computed by our program on a
representative set of examples.

\section{Implicatures}
\label{implicatures}

Although Grice~\shortcite{grice75,grice78} sketched his theory of
logic and conversation only briefly, his ideas had an enormous impact
on the research on pragmatics. Grice wanted to avoid any ambiguity
between what is said in an utterance and the implications, intentions,
or beliefs that are used to interpret the utterance, or that are
conveyed by it.  The beauty of his theory emanates from a very small
set of conversational principles or maxims that are the linguistic
expression of a more general form of human cooperative
behavior~\mbox{\cite[p.~47]{grice75}}.  Here follow these principles:

\vspace{5mm}
\ul{Maxims of quality}

\vspace{3mm}

Try to make your contribution one that is true.
\begin{enumerate}
\item Do not say what you believe to be false.

\item Do not say that for which you lack adequate evidence.
\end{enumerate}

\vspace{5mm}
\ul{Maxims of quantity}

\begin{enumerate}
\item Make your contribution as informative as is required (for the
current purposes of the exchange).

\item Do not make your contribution more informative than is
required\footnote{The validity of this maxim is disputable, because
one may be enticed to subsume it under the maxim of relevance; this is
admitted even by Grice himself. General inquiries into this matter are
made by Green~\shortcite{green89}. A study related to its impact on
redundancy in collaborative dialogues is elaborated by
Walker~\shortcite{walker92}.}.

\end{enumerate}

\vspace{5mm}
\ul{Maxim of relation}

\begin{enumerate}
\item Be relevant.
\end{enumerate}

\vspace{3mm}
\ul{Maxims of manner}

\begin{enumerate}
\item Avoid obscurity of expression.

\item Avoid ambiguity.

\item Be brief (avoid unnecessary prolixity).

\item Be orderly.
\end{enumerate}

By observing these maxims, one is able to explain inferences that do
not belong to the conventional meaning of an utterance. These
inferences are called {\em conversational implicatures}.  It is
important to notice that Gricean maxims are not intended to be
prescriptive rules that conversants must obey. Rather, they are
default rules of the conversants' intentions or knowledge.  At
the end of his paper, Grice~\shortcite[pp.~57--58]{grice75} provides a
list of the features he attributes to conversational implicatures:

\begin{enumerate}
\item They are cancellable.
\item They are non-detachable, i.e., it will be impossible to find a
way of saying the same thing that lacks the corresponding implicature. 
\item They are not part of the semantic meaning of the linguistic
expressions to which they attach. 
\item They are carried by the saying of what is said, rather than by what
is said.
\item Usually, each implicature is a disjunction of possible specific
explanations in a given context.
\end{enumerate}
The first and third property are those that most favours their
cancelability or defeasibility. But if one examines implicatures more
carefully, she will notice that sometimes it is {\em felicitous} to
cancel or defeat them, while at other times it is not.

Following Grice and Levinson~\shortcite{levinson83}, one may
distinguish four categories of implicatures: generalized,
particularized, floating, and conventional implicatures.

{\em Generalized implicatures} are  ones which do not require
a particular context in order to be inferred. The following
generalized implicatures can be inferred if the corresponding maxim is
observed.

\vspace{5mm}
\noindent \ul{Quality} 
\bexample{John bought a new car.}
\bexample{ $\rhd$ I believe he did it and I have adequate evidence for
this.}
Notice that cancelling a quality implicature is infelicitous: 
\bexample{ $\star$ John bought a new car, but I don't believe it. \name{infel1}}

\vspace{5mm}
\noindent \ul{Quantity} 
\bexample{Jane has three exams this term.}
\bexample{$\rhd$ Jane has {\em only} three exams this term.}
Although it would be compatible with the truth of the above utterance
that Jane has four exams, the implicature is that Jane has {\em only}
three exams because if she had  had four (or more), then by the maxim
of quantity (say as much as is required) one should have said so. 
However, one can felicitously defeat the implicature if she utters: 
\bexample{Jane has three exams this term; in fact, she has four. \name{quantity-maxim}}

\vspace{5mm}
\noindent \ul{Relevance} 
\bexample{Give me that book, please.}
\bexample{$\rhd$ Give me that book now.}
In this case, the implicature can be felicitously defeated if one
utters: 
\bexample{Give me that book when you finish it. \name{relevance-maxim}}

\vspace{5mm}
\noindent \ul{Manner} (order) 
\bexample{Ed went to the airport and flew to California.}
\bexample{$\rhd$ Ed performed the above actions in the given order.}
Cancelling the above implicature is infelicitous:  
\bexample{$\star$ Ed went to the airport and flew to California, but he went to
California first. \name{infel2}}

{\em Particularized implicatures} are valid only in a
particular context. For example, one would normally infer
\bexample{$\rhd$ Perhaps the dog ate the roast beef. \name{A}}
from
\bexample{The dog looks happy. \name{B}}
only in a context where~\bsent{B}  was uttered as an answer to
\bexample{Where is the roast beef?~\cite[p.~2]{horton87} \name{C}}
This particularized implicature can be felicitously cancelled if
instead of~\bsent{B}, one utters:
\bexample{The dog looks happy, but it was here all the time.
\name{particularized-implicature}}

Other kinds of implicature come about by blatantly not following
some maxims. These give  the {\em floating implicatures}. For
example, by answering with
\bexample{{\bf A:} My car's not working.} 
to a question such as 
\bexample{{\bf B:} Can I get a ride with you?~\cite[p.~92]{green89}} 
one presumably implicates that because his car is not working, he
cannot provide a ride for {\bf B}.  However, the implicature is
felicitously defeasible if {\bf A} utters:
\bexample{{\bf A:} My car's not working, but I'll take my wife's. \name{floating-implicature}}

Further work in the field of quantity
implicatures~\cite{horn72,gazdar79} gave rise to two important
sub-cases of what today are called {\em conventional implicatures}.

The first is {\em scalar implicatures}.  For any linguistic scale
arranged in a linear order by degree of informativeness or semantic
strength $\langle e_1, e_2, \ldots , e_n \rangle$, the utterance of a
well-formed sentence $A(e_i)$ will implicate $\neg A(e_{i-1})$, $\neg
A(e_{i-2})$, $\ldots$, $\neg A(e_1)$. For example on the scale
$\langle all, most, many, some, few \rangle$, the utterance of
\bexample{John says that some of the boys went to the theatre.
\name{scalar-ex}} 
will implicate the following conventional implicatures:
\bexample{$\rhd$ Not many/most/all of the boys went to the theatre.
\name{scalar-ex-implic}} 
Normally, conventional implicatures are felicitously defeasible.
Consider the following repair for the previous example:
\bexample{ John says that some of the boys went to the theatre. John,
Mike, and Jeff were there. Fred was there too. In fact {\em all} of
them went to the theatre. \name{scalar-implicature}}

The second type of conventional implicature, the {\em clausal
implicature} was formulated by Gazdar~\shortcite{gazdar79}. For
example, if one utters 
\bexample{John believes Margaret to be unfaithful.}
the clausal implicatures are:
\bexample{$\rhd$ It is possible for Margaret to be unfaithful.}
and
\bexample{$\rhd$ It is possible for Margaret to be faithful.}
It seems that cancelling the clausal implicatures is felicitous:
\bexample{John believes Margaret to be unfaithful, but it is impossible
for her to be like this. \name{clausal-implicature}}

There are a number of attempts to accommodate Grice's conversational
principles with syntactic and semantic theories. For example, Gordon
and Lakoff~\shortcite{gordon75} try to formalize the conversational
principles, to incorporate them into a theory of generative semantics,
and to determine if there are rules of grammar that depend on these
conversational principles. The conversational implicatures are defined
in terms of logical entailment derived from a logical translation of
the Gricean maxims, a class of contexts, and the given utterance.
Logical entailment is essentially monotonic. Therefore it cannot
explain the cancelability of some implicatures.  However, Gordon and
Lakoff's suggestion that an appropriate formalism should distinguish
between relations such as entailment, equivalence, presupposition,
assumption, and conversational entailment would be valuable for our
purposes.

Hirschberg is interested in a theory of conversational implicature and
provides a set of criteria for recognizing
them~\shortcite[p.~38]{hirschberg85}.
Green~\shortcite{green90,green92} uses Hirschberg's theory as the
foundation for the introduction of another type of implicature,
{\em normal state implicature}, and for the study of implicature in
indirect replies. Normal state implicature is used to offer an
explanation for the circumscription and qualification problems on
linguistic grounds, but her solution circumvents many important
aspects, such as the representation of the preconditions of a plan, or
the recognition by the hearer of the speaker's plans. We believe that
in most of the cases, is unreasonable to assume that a hearer is aware
not only of the Gricean conversational principles and the conversants'
beliefs but also of the speaker's plans in order to infer the
expected implicatures.

A good analysis of the interaction between the conversational
principles and a natural language generator is presented by
Reiter~\shortcite{reiter90}. His concern is the generation of natural
language utterances that are free of false implicatures, i.e., those
that are maximal under the rules of local brevity, unnecessary
components, and lexical preference. The proposed algorithms prevent
the generation of a sentence such as {\em Sit by the brown wooden
table} when {\em Sit by the table} would suffice. Complexity results
for the various algorithms are also established.

Dealing with implicatures is not purely a theoretical issue:
understanding them can substantially improve many day-to-day tasks,
such as interrogating a database. A computational account of Grice's
conversational principles with respect to a natural language database
interface is presented by Kaplan~\shortcite{kaplan82}. The interface
is supposed to provide cooperative responses for the queries.  For
questions such as {\em Which students got a grade of F in CS105 in
Spring 1980?} there may be situations in which it will be better to
answer {\em CS105 was not given in Spring 1980} instead of {\em nil}.

The examples we have given show a clear delimitation between the
implicatures that are felicitous to cancel and those that are not.
However, we are not aware of any research concerning implicatures that
tries to exploit this difference.  Such an approach seems interesting to
explore because the same dichotomy can be found when one studies
presuppositions: some of them are felicitous to defeat, while others
are not.

\section{Presuppositions}

\subsection{What are presuppositions?}

The main property that seems to delineate presuppositions from
implicatures is the fact
that they are {\em implied} in both positive and negative environments. It
is common knowledge that utterances $a$ and $b$ both presuppose $c$ in the
following two examples.
\bexample{ a.  Peter regrets that he got a speeding ticket. \\
\hspace*{12mm} b.  Peter does not regret that he got a speeding ticket. \\
\hspace*{12mm}	   c.  Peter got a speeding ticket.}
\bexample{ a.  The King of Buganda is bald. \\
\hspace*{12mm}	   b.  The King of Buganda is not bald. \\
\hspace*{12mm}	   c.  There exists a king of Buganda. \name{buganda-ex}}

Soames~\shortcite{soames89} proposes two classes of questions that are
supposed to be answered by any theory of presupposition. Analogous
questions should be addressed by a theory of implicature as well.

\hspace*{10mm} \ul{Foundational questions}
\begin{enumerate}
\item What is presupposition --- what does it mean to say $X$
presupposes $Y$?
\item Why are there linguistically expressed presuppositions at all ---
what functions do presuppositions have in the representation and
communication of information?
\item How are presuppositions of utterances affected by the semantic
rules that determine the information encoded by a sentence relative to
its context, and the pragmatic rules that specify the manner in which
utterances increment sets of assumptions common among conversational
participants? 

\noindent \ul{Descriptive questions}

\item What presuppositions do various constructions give rise to?
\item Which constructions allow utterances to inherit the
presuppositions of their constituents and which do not?  This question
is often called the {\em projection problem}.
\item What do utterances of arbitrary sentences presuppose?
\end{enumerate}

Three different classes of answers have been given for the first question. 
The first  is due to Frege~\shortcite{frege92}, who
sees presupposition as a logical relation between two sentences.   
This category contains what are usually called semantic theories.
Semantic theories are usually built on a definition that captures the
fact that presuppositions are implied in both positive and negative
environments, and that is  similar to the one given below, 

\begin{definition}
Sentence $S_1$ \ul{semantically presupposes} sentence $S_2$ if
and only if $S_1 \entails S_2$ and $\neg S_1 \entails S_2$.
\end{definition} 
This definition does not survive a logical examination because as
Gazdar~\shortcite[p.~90]{gazdar79} has shown, the definition says that
presuppositions {\em are always} true. Several variations on this
definition have been developed using modal logic~\cite{karttunen71} or
a heavy-parentheses notation~\cite{katz76}.  A critique of these
theories is given by Gazdar~\shortcite[pp.~90--103]{gazdar79}.
Revivals of semantic theories are observed even later~\cite{wilson79},
where in order to circumvent the problems an ordered relation on the
entailments is used.  On the basis of this ordering, Sperber and
Wilson distinguish between {\em focalized} and {\em peripheral}
entailments. A critique of their proposal is given by
Horton~\shortcite[pp.~27--28]{horton87}.

The second class of answers is related mainly to
Strawson~\shortcite{strawson50}, who sees presupposition as a relation
between a sentence and its use.  In a Strawsonian world, the existence
of the King of Buganda is a necessary precondition that should be
satisfied before any truth value can be assigned to the sentence {\em
The King of Buganda is bald}.

The third class of answers, the one that encompasses the pragmatic
approaches, is the one that looks most successful.  Pragmatic
approaches offer a variety of definitions for presupposition that
differ primarily in the role played by context~\cite{karttunen74}, by
the conversational participants~\cite{stalnaker73} or by the logic and
pragmatic relation between these factors and
presupposition~\cite{gazdar79,horton87}. We shall look at several of
these approaches in detail in the following sections.

\subsection{Karttunen and Peters}

Karttunen and Peters's evolution of understanding in dealing with
pragmatic inferences reflects a desire for simplification: in their
1979 paper for example, an important class of inferences is labelled
not as presuppositions, as in their earlier work, but rather as
implicatures. The inferences that are triggered by the use of word
{\em even} are an example.  We think this reflects the idea that
presupposition and implicature do not differ too much in their
intrinsic status.  Essentially, they both constitute information added
to the context beyond what is specifically uttered. Furthermore, they
are both cancellable: sometimes felicitously, sometimes infelicitously.

 Karttunen and Peters's theory addresses primarily the projection
problem by introducing the {\em plugs, holes,} and {\em filter}
mechanism. Plugs are constructs that do not allow the
presuppositions of the underlying structure to become presuppositions
of the matrix sentence. The verbs of saying ({\em say, ask, tell,}
etc.) are categorized as plugs:
\bexample{Bill told me that the present King of France is bald.}
\bexample{ $\rhd$ There is a king of France.}
Holes are  those constructs in which presuppositions always
survive embedding. Factives ({\em know, regret}), aspectuals ({\em
begin, stop}), implicatives ({\em manage, remember}) are the most
common examples: 
\bexample{John managed to open the door.}
\bexample{ $\rhd$ John opened the door.}
The filters are  the constructs in which presuppositions act
strangely: they sometimes survive and sometimes fail to survive. The
filtering conditions provide the rules by which presuppositions are
inherited. For example, a conditional having the form {\em if $A$
then $B$} will presuppose $P$ if $B$ presupposes $P$ unless there is
some set $X$ of assumed facts such that $X \cup \{A \}$ semantically
entails $P$. According to this rule, \bsent{mary1} will presuppose
\bsent{mary2}, but \bsent{mary3} will not presuppose \bsent{mary4}.
\bexample{If Mary came to the party, then John would regret that Sue came to
the party. \name{mary1}}
\bexample{ $\rhd$ Sue came to the party. \name{mary2}}
\bexample{If Mary came to the party, then John would regret that she did.
\name{mary3}} 
\bexample{ $\rhd \! \! \! \!/$ Mary came to the party. \name{mary4}} 
A formalization of these rules is developed in Montaque's PTQ grammar.

The theory has been attacked by Gazdar~\shortcite{gazdar79} and Soames
~\shortcite{soames79,soames82} for reasons related
mainly to contradictory presuppositions such as that exhibited in
{\em My teacher is a bachelor or a spinster}, and presuppositions
cancelled by conversational implicatures, e.g. {\em It is possible that
John has children and it is possible that his children are away}.
Another problem is a methodological one: it is very difficult to
construct a taxonomy of plugs, filters, and holes because the
criteria needed to define these notions are ambiguous.

\subsection{Gazdar}
\label{gazdar}

A major step taken by Gazdar~\shortcite{gazdar79} is the shift from
explaining presuppositions in terms of entailment towards explaining
them in terms of consistency. For Gazdar, implicatures and
presuppositions work together towards the enhancement of context.  His
theory is an elegant exploration of how pragmatic rules can explain
the inheritance of presupposition. Gazdar's system is composed of three
parts:
\begin{enumerate}
	\item A set of presupposition-bearing functions, which, when
applied to a sentence, generate all the presuppositions afferent to
that sentence.

	\item A trivial projection rule that projects all the presuppositions
generated in part 1 as the potential presuppositions for the
sentence.

	\item A Gricean filter that cancels some  of the potential
presuppositions. 
\end{enumerate}

	The critics of his theory were mostly concerned with the
ad-hoc order in which the conversational context is enhanced, and the
lack of grounds for this.  However, they do not provide an insight
into the logical properties that Gazdar's presupposition has.
Mercer~\shortcite{mercerphd} and Horton~\shortcite{horton87}
note that Gazdar's method does not allow presuppositions to be
cancelled by knowledge added to the context by later utterances. This
gives presuppositions a very interesting status, assigning them a dual
life: they are cancellable in the first stage (the stage of their
generation and their evaluation against the current context) but they
become indefeasible knowledge in the second stage, after they have
been accepted as presuppositions for the given utterance. Consider
that one utters~\bsent{seq1}. The first sentence in the sequence
presupposes~\bsent{prep1}. Gazdar's theory will appropriately reflect
this.  The second sentence in~\bsent{seq1} comes to cancel the
presupposition but Gazdar's theory is not able to provide an
explanation for this cancellation.
\bexample{My cousin is not a bachelor. He is only five years old.
\name{seq1}} 
\bexample{$\rhd \! \! \! \!/$ ($\rhd$) My cousin is an adult.
\name{prep1}} 
In these conditions, it becomes clear why the
presuppositions are the last pieces of information that are added to
the conversational context.

	From our perspective, of maximal importance is the fact that
Gazdar formalizes presuppositions, in positive environments, as
logical implications~\cite[p.~140]{gazdar79}.  As we will show in
section~\ref{logical analysis}, this is a logical mistake.

\subsection{Mercer}
\label{mercer-critique}
An orthogonal approach is taken by
Mercer~\shortcite{mercer82,mercerphd,mercer88a,mercer88b,mercer90,mercer91}.
He abandons the
projection method for rules of inference in default logic. 
Mercer uses Gazdar's formalization of Grice's cooperative principle, one
of the basic pillars of all pragmatic approaches, and he provides a
proof-theoretic definition for presupposition in terms of default
inferences. Our main objection to this approach is Mercer's use of
natural disjunction as an exclusive disjunction, and the reduction of
natural implication to logical equivalence.  I present below our
counter-arguments to the disjunctive treatment. For conditionals, the
counter-arguments are similar.

Mercer~\shortcite{mercerphd} argues that when a speaker utters a
sentence in the form $A \vee B$, assuming the speaker respects Grice's
conversational principles, and considering Gazdar's formalization of
clausal implicatures, the corresponding default theory representing
this utterance will be:

\begin{equation} 
T = \left\{ K_{S}(A \vee B), P_{S} A, P_{S} \neg A,
P_{S} B, P_{S} \neg B, \alpha_{1}, \ldots, \alpha_{n}, \delta_{1},
\ldots ,\delta_{n} \right\} 
\end{equation} 
where $\alpha_{1}, \ldots, \alpha_{n}$ represent the appropriate
first-order statements, $\delta_{1}, \ldots, \delta_{n}$ represent the
appropriate default rules, and $K_{S}, P_{S}$ represent the common
modal operators for necessity and possibility respectively. Using the
$S_{4}$ modal logic system, Mercer argues that theory $T$ yields two
cases:

\begin{equation}
T_{1} = \left\{ A \wedge \neg B, \alpha_{1}, \ldots, \alpha_{n},
\delta_{1}, \ldots ,\delta_{n} \right\}
\end{equation}
and
\begin{equation}
T_{2} = \left\{ \neg A \wedge B, \alpha_{1}, \ldots, \alpha_{n},
\delta_{1}, \ldots ,\delta_{n} \right\}
\end{equation}
since $ A \wedge \neg B$ and $\neg A \wedge B$ {\em completely
determine } the truth value of both $A$ and $B$ and since the
statements $P_{S}(A \wedge \neg B)$ and $P_{S}(\neg A \wedge B)$ can be
derived from the given theory $T$.

However, the following proof gives us $P_{S}(A \wedge B)$ as a candidate
for the case analysis as well:

\vspace{5mm}
\begin{center}
\begin{tabular}{lll}
1. & $K_{S}(A \vee B) \rightarrow P_{S}A \vee K_{S}B$  &
(provable  in $S_{4}$ modal \\
   &     &  logic~\cite[p.~123]{chellas80}) \\
2. & $K_{S}(A \vee B)$ & (translation of speaker's \\
   &  &  utterance) \\
3. & $A \vee B \leftrightarrow (\neg A \wedge B) \vee (A \wedge \neg
B) \vee (A \wedge B)$ & (PL) \\
4. & $K_{S}((A \vee B) \leftrightarrow (\neg A \wedge B) \vee (A
\wedge  \neg B) \vee (A \wedge B))$ &  (3 + RN) \\
5. &  $K_{S}(A \vee B) \leftrightarrow K_{S}((\neg A \wedge B) \vee (A
\wedge  \neg B) \vee (A \wedge B))$ &  (4 + RE) \\
6. &  $K_{S}((\neg A \wedge B) \vee (A \wedge  \neg B) \vee (A \wedge
B))$ &  (2 + 5  + MP) \\
7. &  $P_{S}(\neg A \wedge B) \vee K_{S}((A \wedge  \neg B) \vee (A \wedge
B))$ &  (1 + 6 + MP) \\
8. &  $P_{S}(\neg A \wedge B) \vee P_{S}(A \wedge  \neg B) \vee K_{S}(A
\wedge B)$ & (1 + 7 + MP) \\
9. &  $K_{S}(A \wedge B) \rightarrow P_{S}(A \wedge B)$ &  (T + PL) \\
10. &  $P_{S}(\neg A \wedge B) \vee P_{S}(A \wedge  \neg B) \vee P_{S}(A
\wedge B)$ &  (8 + 9 + MP) \\
\end{tabular}
\end{center}
\vspace{5mm}

\noindent As shown, under a classical interpretation, the above theory
yields three cases. Mercer does not include in his theory the scalar
implicatures given by Gazdar~\shortcite{gazdar79}, which eventually
eliminates the third case $P_{S}(A \wedge B)$. Therefore, his
treatment of natural disjunction as an exclusive or cannot be
explained on formal grounds. Mercer (personal communication) argued
that he intended his ``proof by cases'' to be interpreted in a
non-traditional way, in which ``the cases are taken from a conjunctive
statement, where the conjuncts are the disjuncts in the above proof''.
He argued that this non-standard notion is the one that must be used
in nonmonotonic reasoning. If we restrain ourselves to a classical
interpretation, the foundations of Mercer's theory collapse: his
definitions do not reflect any longer one's intuitions.

\subsection{Other approaches}

Horton's theory~\shortcite{horton87,horton88}  has as its departure
point the observation that many theories of presupposition make the
following unrealistic assumptions:
\begin{itemize}
\item If sentence S presupposes proposition P, then P is true.
\item If sentence S presupposes proposition P, then all agents involved
share the prior belief that P is true.
\end{itemize}

Most of the intuitions provided by Horton sound acceptable, but when
it comes to their formalization, insurmountable problems occur. The
core of the system is built on two operators: {\em add-presupposition}
and {\em retract-presupposition}, but they are never defined in terms
of beliefs.  Moreover, no hints are given concerning the conditions
under which these operators are triggered.

	A more recent approach to the projection problem is built
from the observation that the data challenging Gazdar's
solution introduced by Soames, Heine, and Fauconnier
concern primarily hypothetical contexts and secondarily contexts of
reported speech or thought. A theory based on these types of contexts is
developed~\cite{kay92} but no references are made for a possible
formalism in which such a theory can be embedded, nor for a
computational model of it.

	A different perspective is given by van der
Sandt~\shortcite{sandt92} and Zeevat~\shortcite{zeevat92} where
presuppositions are understood as anaphoric expressions that have
internal structure and semantic content. Because they have more
semantic content than other anaphors, presuppositions are able to
create an antecedent in the case that the discourse does not provide
one. Van der Sandt provides a computational method for presupposition
resolution in an enhanced discourse representation theory while Zeevat
gives a declarative account for it using update semantics, but none of
the methods is able to accommodate the cancellation of presupposition
that is determined by information added later to the discourse.

	There is only one attempt to embed the theories of
presupposition within a working computational model. It was done by
Weischedel~\shortcite{weischedel79} and it claims that presuppositions
are purely structural in nature, and can thus be generated during
parsing. The claim is supported by an extension of an augmented
transition network grammar with actions capable of generating
presuppositions. Being so strongly bound to the lexical and syntactic
structure for a particular utterance, the theory is not able to
explain presupposition cancellations that occur late in the analyzed
text. Actually none of these theories is able to account for the
cancellation of the presupposition generated in~\bsent{seq1}.

\section{A logical analysis of pragmatic inferences}
\label{logical analysis}

The meaning, the implications, and the intention behind words are just
one part of our day-to-day experience that makes the study of language
difficult. We have already presented the main theories developed for
handling implicatures and presuppositions. We review some of the
examples from a logical perspective.

Consider the following sentence, which exhibits an entailment relation:
\bexample{All turtles are green. \name{turtles}}
This is the kind of statement that is representable
in first-order logic. In accordance with a Tarskian semantics, the
translation $(\forall x)(turtle(x) \rightarrow green(x))$ will
allow one to infer from $turtle(\mbox{Wolfgang})$ that
$green(\mbox{Wolfgang})$. 
By contraposition, anything that is not green cannot
be a turtle: 
\bexample{If something is not green, then it is not a turtle.}
As we will show, many pragmatic inferences are not subject
to this behavior, i.e., they do not obey the contrapositive rule,
so they are not representable in first-order logic using material
implication.   

It has been already shown~\cite[p.~90]{gazdar79} that a definition
of presuppositions in terms of entailment is a logical mistake. To fix
the problem, one would be tempted to  treat presuppositions as
defeasible information (see \ref{mercer-critique} for such an approach). 
Assume that presuppositions are defeasible in positive
and negative environments as well. If this is the case, the behavior
of the related presupposition~\bsent{pres-mary} should be the same for
each of the following utterances: 
\bexample{John regrets that Mary came to the party. \name{infel-pres}}
\bexample{John does not regret that Mary came to the party. \name{fel-pres}}
\bexample{$\rhd$ Mary came to the party. \name{pres-mary}}
This is not the case. Negating the presupposition is infelicitous for
the first utterance, but is acceptable for the second one:
\bexample{ $\star$ John regrets that Mary came to the party, but she did not
come. \name{infel3}} 
\bexample{ John does not regret that Mary came to the party, because she
did not come. \name{mary-yes-no}}
\bexample{$\rhd \! \! \! \! /$ Mary came to the party.
\name{not-pres-mary}}  
It is true that the surface form of a positive utterance is similar to
its negation, but the inferences derived from these utterances have
totally different properties, so they should be studied separately.

The same dichotomy occurs when one studies implicatures.  Sometimes
implicatures are felicitous to defeat, as in~\bsent{fel-implic1}
or~\bsent{fel-implic2}, but sometimes their cancellation is
infelicitous as in~\bsent{mary5}.  
\bexample{ Ed has three exams this term. Actually he has four.
\name{fel-implic1}} 
\bexample{The soup is warm, in fact hot. \name{fel-implic2}}
\bexample{$\star$ Mary is unfaithful, but I don't believe this. \name{mary5}}

We exploit a fundamental difference between semantic and pragmatic
information. On one hand, we consider that semantic information is
information that pertains to our knowledge of the world. It may come in
two flavours: indefeasible, as given in sentence~\bsent{turtles}, or
defeasible, as given in~\bsent{birds fly}.
\bexample{Typically, birds fly. \name{birds fly}}
As a simplification, for the purpose of this thesis, we will deal only
with indefeasible semantic knowledge, but the formalism and program we
propose can be generalized to handle semantic defeasible information
as well.

On the other hand, pragmatic information is information concerning our
knowledge of language use. Presuppositions and implicatures are not
explicitly uttered; therefore, {\em all} of them are defeasible. The
examples we have given with respect to implicatures and
presuppositions inspires us to consider more than one level of
defeasibility for pragmatic inferences. The refinement we propose
accounts for the different strength or commitment that seems to
differentiate certain types of pragmatic inferences.  When an
implicature or a presupposition is infelicitous to defeat, we say that
that implicature or presupposition is {\em infelicitously
defeasible}. When an implicature or presupposition is felicitous to
defeat, we say that that pragmatic inference is {\em felicitously
defeasible}.  The following table shows the defeasibility properties
of implicatures and presuppositions in positive and negative
environments:

\vspace{5mm}
\begin{center}
\begin{tabular}{|c|c|c|} \hline \hline
{\em Property} & {\em Implicature} & {\em Presupposition} \\ \hline 
Felicitously defeasible in &  yes/no    &  no       \\
positive sentences & & \\ \hline
Felicitously defeasible in & yes/no & yes \\
negative sentences & & \\ \hline
\end{tabular}
\end{center}
\vspace{5mm}

\noindent The difference between felicitously and infelicitously
defeasible implicatures and presuppositions, which we have emphasized
so far, determines us to attempt to build a taxonomy of pragmatic
inferences that has a finer granularity. We still subscribe to the
Gricean view that {\em all} conversational implicatures are
defeasible, but we distinguish between the ones that are felicitous to
cancel and those that are not. As shown, the taxonomy we propose does
not overlap the previous one that divides pragmatic inferences into
presuppositions and implicatures because both these classes can be
felicitously or infelicitously defeated.

In summary, we consider that the knowledge of an agent can be divided
as follows:
\begin{itemize}
\item Semantic knowledge 
\begin{itemize}
\item Indefeasible 
\item Defeasible
\end{itemize}
\item Pragmatic knowledge 
\begin{itemize}
\item Felicitously defeasible
\item Infelicitously defeasible
\end{itemize}
\end{itemize} 

Our ultimate goal is to develop a computational method for determining
pragmatic inferences and for signalling infelicities; this requires a
classification of the pragmatic inferences in accordance not only with
their source but also with their properties. A potential source is for
example a factive such as the verb {\em regret} that normally
presupposes its complement.  After one identifies the environment in
which a pragmatic inference is triggered, she should map it into a
specific representation which should reflect the properties that the
inference is expected to have.

If we formalize the infelicitously defeasible inferences as
entailments, an inconsistency will occur. But there is no
inconsistency when one utters~\bsent{mary5}.
The fact that {\em I believe that Mary is unfaithful}
is not uttered, it is only {\em implied}.  Another reason that
prevents one of formalizing infelicitously defeasible inferences as
entailments is their violation of the contrapositive rule. For
example, one would be tempted to say that~\bsent{mary12} is a logical
implication of~\bsent{mary11} or that~\bsent{cs2} is an implication
of~\bsent{cs1}. 
\bexample{John regrets that Mary came to the party. \name{mary11}}
\bexample{$\rhd$ Mary came to the party. \name{mary12}}
\bexample{I failed CS100 again. \name{cs1}}
\bexample{$\rhd$ I failed CS100 before. \name{cs2}}
This is not the case, because \bsent{mary13} does not imply in most of
the cases either \bsent{mary14}, or \bsent{mary15}; and  \bsent{cs3}
does not imply  either \bsent{cs4}, or \bsent{cs5}. The pairs
\bsent{mary14}, \bsent{mary15}, and \bsent{cs4}, \bsent{cs5} stand for
the two kinds of natural language negation: internal and external.
\bexample{Mary did not come to the party. \name{mary13}}
\bexample{John does not regret that Mary came to the party.
\name{mary14}}
\bexample{It is not the case that John regrets that Mary came to the
party. \name{mary15}}
\bexample{I did not fail CS100 before. \name{cs3}}
\bexample{I did not fail CS100 again. \name{cs4}}
\bexample{It is not the case that I did not fail CS100 again.
\name{cs5}}

If we consider again sentences that are infelicitous to utter such
as~\bsent{infel1},~\bsent{infel2},~\bsent{infel3}, or~\bsent{mary5},
we notice that the notion of {\em pragmatic infelicity} can be
associated with the cancellation of some Gricean inferences.  For
example, it is {\em infelicitous} to cancel an implicature derived
using the maxim of quality~\bsent{infel1},~\bsent{mary5}, the maxim of
manner~\bsent{infel2} or a presupposition triggered in a positive
environment~\bsent{infel3}.

The taxonomy we propose does not contradict Gazdar's
opinion~\shortcite[p.~40]{gazdar79}:
\begin{quote}
The implicature must not be a truth condition of the sentence involved.
\end{quote}

\section{A pragmatic analysis of default logics}

We are interested in a mathematical formalism that is capable of
capturing implicatures, presuppositions, and infelicities. Because
pragmatic inferences are defeasible, it seems that default logics are
appropriate tools for doing this.  However, we have already emphasized
that sometimes the cancellation of these pragmatic inferences is
felicitous, while sometimes it is infelicitous. This dichotomy implies
that if one wants to distinguish between these two types of behavior,
she has to take into consideration more than one level of defeasible
information. Moreover, to signal an infelicitous cancellation, one
should have available at the same time not only the presupposed or
implicated information but its negation as well.

The default theory proposed by Reiter~\shortcite{reiter80} is a formal
object consisting of a pair $(D,W)$ where $D$ is a set of defaults and
$W$ a set of closed well-formed formulas. The defaults have the form 
\[  \frac{\alpha : \beta}{w} \]
where $\alpha$ is the {\em prerequisite} and $w$ is the {\em
consequent}. Intuitively, if $\alpha$ holds and if it is consistent to
believe $\beta$, one would be entitled to derive $w$. For any default
theory, one can construct a set of extensions, i.e., a set of
acceptable beliefs that may hold, about the incompletely specified
world $W$. In a simplified account, extensions are defined in
terms of a fixed-point operator that maps an arbitrary set of formulas
to the smallest deductively closed set $E$ that contains $W$ and
satisfies the condition: for any default $ \frac{\alpha :
\beta}{\gamma} \in D$, if $\alpha \in E$ and $\neg \beta
\not\in E$ then $\gamma \in E$.  

Mercer (section ~\ref{mercer-critique}) tries to use this theory for
formalizing presuppositions. Even if we assume that his approach has
no problems, we are still unable to differentiate between felicitously
defeasible and infelicitously defeasible inferences in such a
framework.  Reiter's default logic is built on consistency basis; that
means that all the information that belongs to a given theory has the
same strength.  Therefore, there is no obvious encoding that can
determine the difference between felicitously and infelicitously
defeasible information.  We cannot accommodate the two types of
defeasibility even if we consider extensions of Reiter's logic such as
those developed by Brewka~\shortcite{brewka94} or
Delgrande~\shortcite{delgrande94}.  Their main interest is to
formalize the notion of preference over extensions and priority over
defaults, but their work is still inadequate for our purposes. Even if
we assign a higher priority to a default that formalizes an
infelicitously defeasible inference over a default that formalizes a
felicitously defeasible inference, the extension we get keeps no trace
of the defaults that were used for deriving the information within the
extension. In the process of analyzing a sequence of utterances, we
can observe how extensions change their content but we cannot specify
at a given time what information is presupposed, what is implicated,
and we cannot say if an infelicity has occured.

%Exactly the same problems are exhibited by formalizations using
%circumscriptive theories~\cite{mccarthy80,mccarthy85,lifschitz85}. The
%{\em extensions} are obtained in this case using a preference relation
%over a set of models but again, we cannot specify at a given time what
%information is a presupposition, what is a implicature, and we cannot
%say if an infelicity has occured. 

Nait Abdallah's~\shortcite{areski88,areski91a} ionic logic is build on
the notion of {\em default ions}. A default ion, $(a,b)_{\star}$ is a
syntactic object having a well-defined semantic behavior that is used
to refine a partially defined theory; it should be read: if $a$ is
consistent with the current logical scope then infer $b$. Ions are
used to construct partial models for theories.  Ionic logic differs
from Reiter's default logic in that instead of having a fixed-point
extension construction, it exhibits Scott's notion of continuity in
which the final solution is obtained as a limit of a continuous
process that generates better and better approximations of the
solution.

At the semantic level, ionic logic gives objects with two components:
a {\em kernel} part that cannot be revised and a {\em belt} part that
can be revised and changed. The belt part is supported by a set of
justifications.  Formally, a default interpretation is a triple
$\langle i_{0}, J, i_{1} \rangle$ such that:
\begin{itemize}
 \item $i_{0}$ (the kernel valuation) and $i_{1}$ (the belt
valuation) are partial mappings from the set of propositional symbols
to the interpretative structure;
\item  $J$ is the justification set that supports the belt valuation.
\end{itemize}
The partial models are constructed using a generalization of the
semantic tableau technique.

In ionic logic, all defeasible information is paired by
a nonempty set of justifications. Therefore, one is able to signal
when some defeasible information has been cancelled. Unfortunately,
ionic logic deals with only one kind of defeasible information. Hence
there is no possibility to accommodate within this theory the notion
of felicitously and infelicitously defeasible inferences.

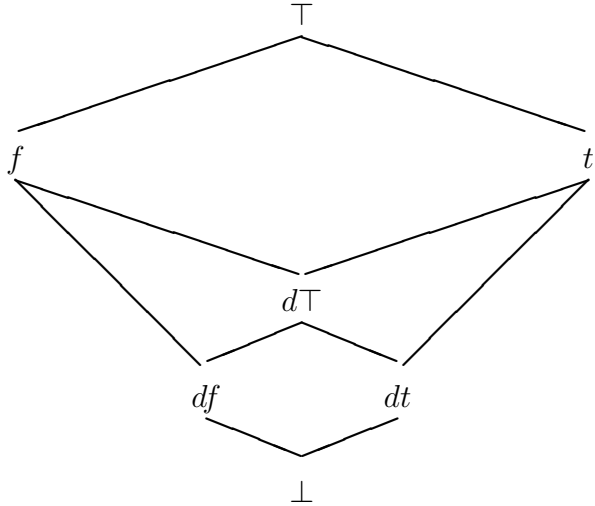
\begin{figure}[htbp]
\centering
\setlength{\unitlength}{0.012500in}%
\begingroup\makeatletter
% extract first six characters in \fmtname
\def\x#1#2#3#4#5#6#7\relax{\def\x{#1#2#3#4#5#6}}%
\expandafter\x\fmtname xxxxxx\relax \def\y{splain}%
\ifx\x\y   % LaTeX or SliTeX?
\gdef\SetFigFont#1#2#3{%
  \ifnum #1<17\tiny\else \ifnum #1<20\small\else
  \ifnum #1<24\normalsize\else \ifnum #1<29\large\else
  \ifnum #1<34\Large\else \ifnum #1<41\LARGE\else
     \huge\fi\fi\fi\fi\fi\fi
  \csname #3\endcsname}%
\else
\gdef\SetFigFont#1#2#3{\begingroup
  \count@#1\relax \ifnum 25<\count@\count@25\fi
  \def\x{\endgroup\@setsize\SetFigFont{#2pt}}%
  \expandafter\x
    \csname \romannumeral\the\count@ pt\expandafter\endcsname
    \csname @\romannumeral\the\count@ pt\endcsname
  \csname #3\endcsname}%
\fi
\endgroup
\begin{picture}(258,217)(31,600)
\thicklines
\put(160,795){\line( 3,-1){118.500}}
\put( 40,735){\line( 3,-1){118.500}}
\put(280,735){\line(-3,-1){118.500}}
\put( 40,735){\line( 1,-1){ 77.500}}
\put(280,735){\line(-1,-1){ 77.500}}
\put(160,675){\line(-5,-2){ 39.655}}
\put(160,675){\line( 5,-2){ 39.655}}
\put(160,795){\line(-3,-1){118.500}}
\put(120,635){\line( 5,-2){ 39.655}}
\put(200,640){\makebox(0,0)[b]{\smash{\SetFigFont{12}{14.4}{rm}$dt$}}}
\put(200,635){\line(-5,-2){ 39.655}}
\put(160,800){\makebox(0,0)[b]{\smash{\SetFigFont{12}{14.4}{rm}$\top$}}}
\put( 40,740){\makebox(0,0)[b]{\smash{\SetFigFont{12}{14.4}{rm}$f$}}}
\put(280,740){\makebox(0,0)[b]{\smash{\SetFigFont{12}{14.4}{rm}$t$}}}
\put(160,600){\makebox(0,0)[b]{\smash{\SetFigFont{12}{14.4}{rm}$\bot$}}}
\put(160,680){\makebox(0,0)[b]{\smash{\SetFigFont{12}{14.4}{rm}$d\top$}}}
\put(120,640){\makebox(0,0)[b]{\smash{\SetFigFont{12}{14.4}{rm}$df$}}}
\end{picture}
\caption{A bilattice for default reasoning}
\label{kifer-lattice}
\end{figure}

Kifer~\cite{kifer92a,kifer92b} constructs his annotated logic in
order to allow the reasoning in the presence of inconsistency. In
classical logic, anything follows from an inconsistent theory;
therefore, it is impossible to distinguish between $\{ flies(tweety),
\neg flies(tweety) \}$ and $\{ flies(tweety), \neg flies(tweety),
married(john, \linebreak mary) \}$ even if intuition tells us that the
second theory contains more information.  Underlying annotated logic
is a belief upper semi lattice with a unique upper bound for every pair
of elements. A similar lattice for dealing with defaults is considered
in Ginsberg~\shortcite{ginsberg88} and it looks like the one in
figure~\ref{kifer-lattice}. 

The syntax of annotated logic differs from that of first-order logic
in having the atomic formulas appended by an annotation that is drawn
from the underlying belief semi lattice. Thus, well-formed formulas are
$(\exists x)q(x,z):f$ or $flies(tweety):dt$. 
The semantics of annotated logic defines that an atomic formula
$p(t_1,\ldots,t_n):s$ is satisfied by a valuation $v$ if and only if
$s \leq I_P(p)(v(t_1), \ldots, v(t_n))$ where $I_P$ associates to each
predicate symbol $p$ a function $I_P(p) : Domain \rightarrow
Semilattice$. 

It is obvious now that such a definition is able to deal with
inconsistencies using an appropriate underlying semi lattice, but it
still does not help us in differentiating between felicitously and
infelicitously defeasible inferences. As soon as some stronger
information is derived, in order to satisfy the new set, the $I_P(p)$
function should pick up a stronger value from the underlying
bilattice. The new theory will be satisfied, but we will not be able to
notice that some weaker information has been cancelled. The logic is
not designed to deal with inconsistencies occuring between different
levels of strength. A technical drawback is the fact that there are
two types of negations and implications in annotated logic and there is
no criterion to choose between one use or another.

 	% relevant literature review
\chapter{Stratified logic}
\label{stratified-logic}

A careful analysis of pragmatic inferences has shown that a simple
dichotomy between defeasible and indefeasible inferences is not enough
if one wants to deal with infelicities. A refined taxonomy built in
terms of felicitously and infelicitously defeasible inferences has
been proposed instead, but no candidate has been found in the default
literature that fulfills our requirements and no obvious extension
having the desired properties has been foreseen for any
candidate. Either the formalisms studied handle only one type of
defeasibility, or they cannot signal inconsistencies that occur
between information that has been inferred with different
strengths. Useful features have been identified however in
Kifer's~\cite{kifer92b,kifer92a} and Nait
Abdallah's~\shortcite{areski88,areski91a} work. Our purpose is to find
a way to put together the idea that logical formulas may have
different strengths (Kifer), with the semantic tableau method that
seems to be more attractive from a computational perspective (Nait
Abdallah) than a fixed-point operator. The result is a new logical
formalism called {\em stratified logic}.

\section{Informal intuitions}

\begin{figure}
\setlength{\unitlength}{0.012500in}%
\begingroup\makeatletter
% extract first six characters in \fmtname
\def\x#1#2#3#4#5#6#7\relax{\def\x{#1#2#3#4#5#6}}%
\expandafter\x\fmtname xxxxxx\relax \def\y{splain}%
\ifx\x\y   % LaTeX or SliTeX?
\gdef\SetFigFont#1#2#3{%
  \ifnum #1<17\tiny\else \ifnum #1<20\small\else
  \ifnum #1<24\normalsize\else \ifnum #1<29\large\else
  \ifnum #1<34\Large\else \ifnum #1<41\LARGE\else
     \huge\fi\fi\fi\fi\fi\fi
  \csname #3\endcsname}%
\else
\gdef\SetFigFont#1#2#3{\begingroup
  \count@#1\relax \ifnum 25<\count@\count@25\fi
  \def\x{\endgroup\@setsize\SetFigFont{#2pt}}%
  \expandafter\x
    \csname \romannumeral\the\count@ pt\expandafter\endcsname
    \csname @\romannumeral\the\count@ pt\endcsname
  \csname #3\endcsname}%
\fi
\endgroup
\begin{picture}(280,182)(80,615)
\thicklines
\put( 95,715){\vector( 0, 1){ 50}}
\put( 95,635){\vector( 0, 1){ 50}}
\put(255,715){\vector( 0, 1){ 55}}
\put(255,635){\vector( 0, 1){ 50}}
\put( 95,635){\vector( 3, 1){145.500}}
\put(255,635){\vector(-3, 1){145.500}}
\put( 95,715){\vector( 3, 1){145.500}}
\put(255,715){\vector(-3, 1){145.500}}
\put( 80,775){\makebox(0,0)[lb]{\smash{\SetFigFont{12}{14.4}{rm}$\topd$}}}
\put( 80,695){\makebox(0,0)[lb]{\smash{\SetFigFont{12}{14.4}{rm}$\topi$}}}
\put( 80,615){\makebox(0,0)[lb]{\smash{\SetFigFont{12}{14.4}{rm}$\topu$}}}
\put(240,775){\makebox(0,0)[lb]{\smash{\SetFigFont{12}{14.4}{rm}$\botd$}}}
\put(240,695){\makebox(0,0)[lb]{\smash{\SetFigFont{12}{14.4}{rm}$\boti$}}}
\put(240,615){\makebox(0,0)[lb]{\smash{\SetFigFont{12}{14.4}{rm}$\botu$}}}
\put(360,620){\makebox(0,0)[lb]{\smash{\SetFigFont{12}{14.4}{rm}{\bf U}ndefeasible Layer}}}
\put(360,700){\makebox(0,0)[lb]{\smash{\SetFigFont{12}{14.4}{rm}{\bf I}nfelicitously Defeasible Layer}}}
\put(360,780){\makebox(0,0)[lb]{\smash{\SetFigFont{12}{14.4}{rm}Felicitously {\bf D}efeasible Layer}}}
\end{picture}
\caption{The lattice that underlies stratified logic}
\label{my-lattice}
\end{figure}
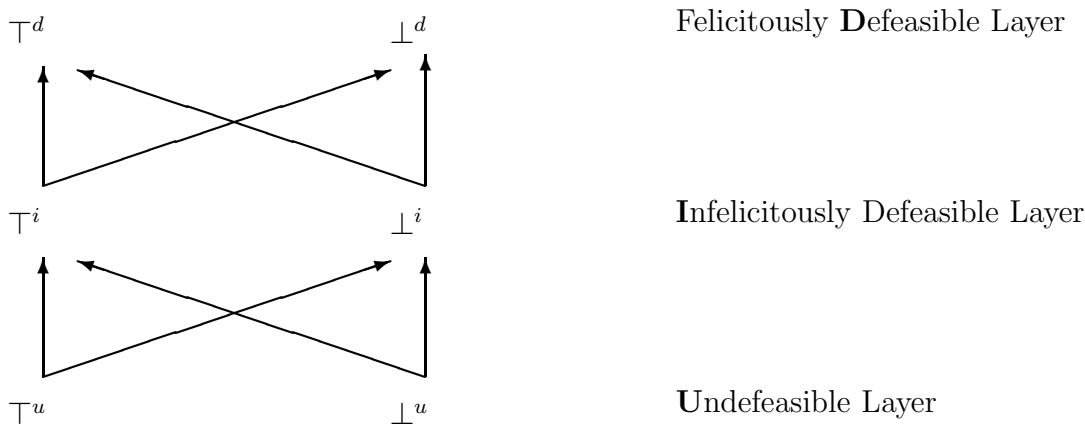

The solution presented here is new. Instead of treating the notion of
defeasibility on justification-based grounds, we conjecture that
defeasible information should have a different status not only
semantically, but also syntactically. The lattice in
figure~\ref{my-lattice} underlies the semantics of stratified logic.
The lattice depicts the three levels of strength that seem to account
for the inferences that pertain to natural language semantics and
pragmatics: indefeasible information belongs to the $u$ layer,
infelicitously defeasible information belongs to the $i$ layer, and
felicitously defeasible information belongs to the $d$ layer. Each
layer is partitioned according to its polarity in truth,
$\topu,\topi,\topd$, and falsity, $\botu,\boti,\botd$. The lattice
shows a partial order that is defined over the different levels of
truth. For example, something that is indefeasibly false, $\botu$, is
stronger than something that is infelicitously defeasible true,
$\topi$, or felicitously defeasible false, $\botd$. Formally, we say
that the $u$ level is stronger than the $i$ level, which is stronger
than the $d$ level: $u < i < d$.
At the syntactic level we allow formulas to be labelled according to
the same underlying lattice. This will give us formulas such as
$a^d, b^u$, or $\neg c^i$  for the propositional case and such as
$bird^u(tweety)$ or $\neg flies^d(tweety)$ for the first-order case.

We summarize the features that make stratified logic appropriate for
handling pragmatic inferences and signalling infelicities:
\begin{itemize}
\item {\em Stratified logic has more than one level of truth.} This
allows one to distinguish among the indefeasible information that
characterizes the semantic inferences and the defeasible information
that characterizes the pragmatic ones. Because there is more than one
level of defeasibility, both infelicitously and felicitously
defeasible inferences can be formalized.
\item {\em In stratified logic formulas are labelled,} or are given a
specific strength. Therefore, one will no longer need justifications
for determining when a defeasible inference is acceptable. If some
defeasible inference is overlapped by some stronger information we can
easily account for this. On one hand we can update our sematic model
so that it reflects the fact that some new stronger information has
been inferred. On the other hand, there is no need for withdrawing the
old weak information that is anyway overlapped by the stronger
information, so our framework is in this sense monotonic. Because weak
and strong information coexist at the same time, it is easy {\em to
signal} when such an inconsistency between different levels of truth
has occurred.
\item {\em Stratified logic is computationally attractive.} Details
about this issue are given in chapter~\ref{implementation issues}.
\end{itemize}

The rest of the chapter contains a formal description of propositional
and first-order stratified logic; therefore, for some people, it is
inherently boring.  Most of the definitions, lemmas, and theorems in
this chapter are used only to facilitate the proofs that show that
stratified tableaux are sound and complete. A reader who is not so
enthusiastic about formal proofs can follow the rest of the thesis if
she has a solid understanding of the way satisfaction and
model-ordering is defined in our framework and if she is familiar with
the tableau method.

\section{The syntax and the semantics of stratified propositional logic}

\subsection{Syntax}

For any set $P$ of propositional letters, the stratified well-formed
formulas {\em (s-wff)}  are recursively defined as follows:
\begin{enumerate}
\item The constants $\topu, \botu, \topi, \boti, \topd,$ and $\botd$
are s-wffs.
\item Each propositional letter labelled with one of the symbols $\{
u,i,d \}$ is a {\em s-wff}.
\item The set of formulas is closed under $\neg, \rightarrow, \wedge,
\vee, \leftrightarrow$.
\end{enumerate}

Examples of {\em s-wffs} are: $b^u \rightarrow f^d$, $b^u \wedge a^i \vee (a^d \rightarrow b^u)$.

The following are not {\em s-wffs}: $(a^i \rightarrow b^u)^i$, $a
\rightarrow  \rightarrow b$.

For the purpose of this thesis, as a simplification, we will omit the
superscript {\em u}: whenever a formula $\alpha$ lacks a superscript,
by default it will be read as $\alpha^u$.

\subsection{$\cal S$-propositional semantics}

The satisfiability relation is extended to the three levels we have
introduced. Hence, we will have {\em u-satisfiability}, $\uentails$,
{\em i-satisfiability}, $\ientails$, and {\em d-satisfiability},
$\dentails$. An {\em extended valuation} or an {\em extended truth
assignment} is a mapping from the set of propositional formulae into
the partially ordered set $\{ \topu, \botu, \topi, \boti, \topd, \botd
\}$ that constitutes the lattice in figure~\ref{my-lattice}. We use
the notion of {\em x-satisfiability} as a compact term for {\em u, i}
or {\em d-satisfiability}. The strength of stratified logic relies in
the way satisfaction is defined. An extended truth assignment will
{\em x-satisfy} a set of formulas, if it uses only truth values weaker
than or equal to $x$. This means that if one wants to {\em i-satisfy} a
set of formulas, she will be allowed to choose for the truth assignment
only values from the following set: $\{\topi, \boti, \topd, \botd \}$.
If she wants to {\em u-satisfy} the same set of formulas, she is
allowed to choose any value that is given in the underlying lattice. 

For the satisfaction notion, we use the same simplification as the one
we use for formulas, i.e., whenever a satisfiability relation,
$\entails$, lacks its superscript, it is read as $\uentails$.

\begin{definition} \label{xentailsyes}
For any atomic formula $a$ and any extended-valuation $\sigma$,\\ the
x-satisfiability relations are defined as follows:

\begin{itemize}
\item $\sigma \entails a$ iff $a^\sigma = \top$
\item $\sigma \entails a^i$ iff $a^\sigma \in \{ \top, \bot, \topi \}$
\item $\sigma \entails a^d$ iff $a^\sigma \in \{ \top, \bot, \topi,
\boti, \topd \}$ 

\item $\sigma \ientails a$ iff $a^\sigma = \topi $
\item $\sigma \ientails a^i$ iff $a^\sigma = \top^i $
\item $\sigma \ientails a^d$ iff $a^\sigma \in \{  \topi, \boti, \topd \}$
 
\item $\sigma \dentails a$ iff $a^\sigma  = \topd$
\item $\sigma \dentails a^i$ iff $a^\sigma = \topd$
\item $\sigma \dentails a^d$ iff $a^\sigma = \topd$
\end{itemize}

For any negation of an atomic formula $\neg a$ and any
extended-valuation $\sigma$, \\ the x-satisfiability relations are
defined as follows:

\begin{itemize}
\item $\sigma \entails \neg a$ iff $a^\sigma = \bot$
\item $\sigma \entails \neg a^i$ iff $a^\sigma \in \{ \top, \bot, \boti \}$
\item $\sigma \entails \neg a^d$ iff $a^\sigma \in \{ \top, \bot, \topi,
\boti, \botd \}$ 

\item $\sigma \ientails \neg  a$ iff $a^\sigma = \boti $ 
\item $\sigma \ientails \neg a^i$ iff $a^\sigma = \boti $
\item $\sigma \ientails \neg a^d$ iff $a^\sigma \in \{  \topi, \boti, \botd \}$
 
\item $\sigma \dentails \neg a$ iff $a^\sigma  = \botd $
\item $\sigma \dentails \neg a^i$ iff $a^\sigma = \botd $
\item $\sigma \dentails \neg a^d$ iff $a^\sigma = \botd$
\end{itemize}
\end{definition}

\begin{definition} \label{satisfaction-definition}
An extended truth valuation $\sigma$ x-satisfies a formula $\alpha$
iff 
\begin{itemize}
\item $\alpha$ is an atomic formula or the negation of an atomic
formula and  $\sigma \xentails \alpha$ as specified 
in definition~\ref{xentailsyes};
\item $\sigma \xentails \alpha_1$ where $\alpha = \neg \neg \alpha_1$;
\item $\sigma \xentails \neg \alpha_1$ or $\sigma \xentails \alpha_2$, where
$\alpha = \alpha_1 \rightarrow \alpha_2$;
\item $\sigma \xentails \alpha_1$ and $\sigma \xentails \neg \alpha_2$, where
$\alpha = \neg (\alpha_1 \rightarrow \alpha_2$).
\end{itemize}
\end{definition}

Definition~\ref{satisfaction-definition} can be easily extended to
conjunction and disjunction as well, in the usual manner. For example,
$\sigma \xentails \alpha$ where $\alpha = \alpha_1 \wedge \alpha_2$
iff $\sigma \xentails \alpha_1$ and $\sigma \xentails \alpha_2$. To
ease our job, we will give our proofs only for negation and
implication, but one should keep in mind that the same methodologies
can be applied for other logical connectors as well.  However, it
is important to notice that our definitions imply neither that from
$\sigma \xentails \alpha$ one may conclude $\sigma
\xnotentails \neg \alpha$, nor that from  $\sigma \xnotentails \alpha$ one
may conclude  $\sigma \xentails \neg \alpha$. For example $\{ a/\top\}
\entails a^i$ and $\{a/\top\} \entails \neg a^i$

\begin{definition}
An extended truth valuation $\sigma$ x-satisfies a set of formulas
$\Phi$ iff $\sigma$ x-satisfies each formula $\varphi \in \Phi$ 
\end{definition}

We are interested in interpreting a given theory with different
degrees of {\em optimism}. This is exactly what the different levels
of satisfaction provide us. The reader can notice that the way
satisfaction is defined for the three levels of truth differs only
with respect to the set of values that the corresponding truth
assignment is allowed to range over.  The $u$-level is the one that
puts no constraints with respect to the values that a truth assignment
can take from the underlying lattice. Therefore, there are more
possibilities to {\em u-satisfy} a formula than to {\em d-satisfy} it.
In this sense, the {\em u-satisfiability} relation is {\em more
optimistic} than the other two. Note that a given theory becomes
$u$-inconsistent only if it is inconsistent in the classical sense.
We cannot say the same thing about the other two levels of
satisfaction. For example the set $\{ a, a^i,\neg a^d\}$ is {\em
i-satisfiable} but is not {\em d-satisfiable}. The trichotomy among
the satisfiability relations gives one a very powerful tool for
analyzing a set of sentences. Depending on her purposes, one can
interpret the same theory from an optimistic or pessimistic
perspective. Thus, one is able to give the expected interpretation
when some information is overlapped by a stronger one, but she is also
able to signal that a cancellation has occurred. For example, the set
$\{a, \neg a^i,\neg a^d\}$ is {\em u-satisfiable} but is not {\em i-}
or {\em d-satisfiable}. Informally, one can still find a truth
assignment to satisfy the initial set, but she is also aware that some
infelicitously defeasible and felicitously defeasible information has
been cancelled.

For the purpose of this thesis, an {\em extended propositional truth
valuation} will be encoded as a list of pairs such that the first
member in the pair represents a propositional variable and the second
one an extended truth value, i.e., a member of the set $\{ \top, \bot,
\topi, \boti, \topd, \botd \}$. Here are some examples:

$\{ a/ \top \} \entails \{ a, \neg a^i, a^d \}$, but $\{ a, \neg a^i,
a^d \}$ is not {\em i-} or {\em d-satisfiable};

$\{ a/ \topi \} \ientails \{ a, a^i \}$;

$\{ a/ \topd \} \dentails \{ a \}$;

$\{ a/ \top \} \inotentails \{ a^i \}$.

\begin{definition}
A set of formulas $\Phi$  {x-entails} a formula $\alpha$ iff every
extended truth assignment that x-satisfies $\Phi$, x-satisfies
$\alpha$ as well.
\end{definition}

\begin{theorem} \label{satisfaction trans}
If $\sigma \xentails \varphi$, then $\sigma \yentails \varphi$, where
$y \leq x$. For example, if $\sigma \dentails \alpha$, then $\sigma
\uentails \alpha$.
\end{theorem}

{\em Proof: } For the atomic case, the theorem follows directly from
the definition. For compound formulas, use induction on the degree of
$\varphi$.

\begin{theorem} \label{not-sat-cor}
If $\sigma \xnotentails \varphi$, then $\sigma \ynotentails \varphi$,
where $x \leq y$. For example, if $\sigma \unotentails \alpha$, then
$\sigma \inotentails \alpha$.  
\end{theorem}

{\em Proof: } For the atomic case the theorem follows directly from
the definition. For compound formulas, use induction on degree of
$\varphi$.

\subsection{Stratified propositional tableaux}

Model construction in stratified logic is accomplished by a
generalization of the ``Beth tableau'' technique. The method was
introduced by Hintikka and Beth, and has its origins in the sequent
calculus of Gentzen. A formal definition is given by
Smullyan~\shortcite{smullyan70}. We give a propositional calculus
example to illustrate the method.

\begin{figure}[htbp]
\centering
\setlength{\unitlength}{0.0125in}%
\begin{picture}(310,230)(40,570)
\put( 40,610){\line( 1, 0){ 30}}
\put(120,610){\line( 1, 0){ 30}}
\put(200,610){\line( 1, 0){ 30}}
\put(240,570){\line( 1, 0){ 30}}
\put(320,570){\line( 1, 0){ 30}}
\put(160,695){\line(-2,-1){ 54}}
\put(160,695){\line( 4,-1){ 95.294}}
\put(100,655){\line(-4,-3){ 34.400}}
\put(100,655){\line( 5,-3){ 40.441}}
\put(260,655){\line(-4,-3){ 34.400}}
\put(260,655){\line( 5,-3){ 40.441}}
\put(300,610){\line(-5,-3){ 34.559}}
\put(300,610){\line( 2,-1){ 40}}
\put(120,790){\makebox(0,0)[lb]{\raisebox{0pt}[0pt][0pt]{\tenrm (1)  $\neg[ p \wedge ( q \vee r)] \rightarrow [(p \wedge q) \vee (p \wedge r)]$}}}
\put(120,775){\makebox(0,0)[lb]{\raisebox{0pt}[0pt][0pt]{\tenrm (2) $[p \wedge (q \vee r)]$}}}
\put(120,760){\makebox(0,0)[lb]{\raisebox{0pt}[0pt][0pt]{\tenrm (3) $\neg [(p \wedge q) \vee (p \wedge r)]$}}}
\put(120,745){\makebox(0,0)[lb]{\raisebox{0pt}[0pt][0pt]{\tenrm (4) $p$}}}
\put(120,715){\makebox(0,0)[lb]{\raisebox{0pt}[0pt][0pt]{\tenrm (6) $\neg (p \wedge q)$}}}
\put(120,700){\makebox(0,0)[lb]{\raisebox{0pt}[0pt][0pt]{\tenrm (7) $\neg (p \wedge r)$}}}
\put(120,730){\makebox(0,0)[lb]{\raisebox{0pt}[0pt][0pt]{\tenrm (5) $(q \vee r)$}}}
\put( 80,660){\makebox(0,0)[lb]{\raisebox{0pt}[0pt][0pt]{\tenrm (8) $q$}}}
\put( 40,620){\makebox(0,0)[lb]{\raisebox{0pt}[0pt][0pt]{\tenrm (10) $\neg p$}}}
\put(120,620){\makebox(0,0)[lb]{\raisebox{0pt}[0pt][0pt]{\tenrm (11) $\neg q$}}}
\put(240,660){\makebox(0,0)[lb]{\raisebox{0pt}[0pt][0pt]{\tenrm (9) $r$}}}
\put(200,620){\makebox(0,0)[lb]{\raisebox{0pt}[0pt][0pt]{\tenrm (12) $\neg p$}}}
\put(280,620){\makebox(0,0)[lb]{\raisebox{0pt}[0pt][0pt]{\tenrm (13) $\neg q$}}}
\put(240,580){\makebox(0,0)[lb]{\raisebox{0pt}[0pt][0pt]{\tenrm (14) $\neg p$}}}
\put(320,580){\makebox(0,0)[lb]{\raisebox{0pt}[0pt][0pt]{\tenrm (15) $\neg r$}}}
\end{picture}
\caption{The Beth Tableau Technique --- An Example}
\label{tableau:example}
\end{figure}
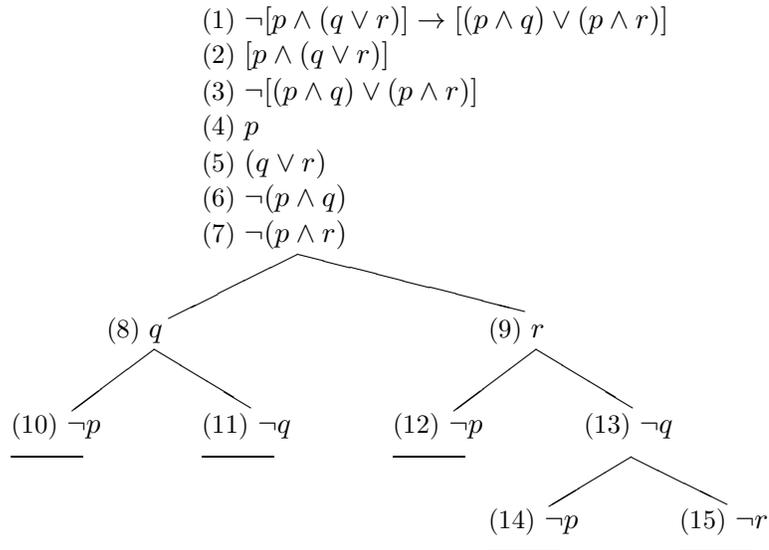

Suppose we wish to show that the sentence $[p \wedge (q \vee r)]
\rightarrow [( p \wedge q) \vee (p \wedge r)]$ is valid. The 
tableau in Figure~\ref{tableau:example} does this.
The tableau is constructed as follows. We see if we can derive a
contradiction from the assumption that the given formula is false. So
our first line consists of the negation of this formula. A formula of
the form $x \rightarrow y$ can be false only if $x$ is true and $y$ is
false. Thus (in the language of the tableau), $x$ and $\neg y$ are
{\em direct consequences} of the formula $\neg (x \rightarrow y)$. So
we write lines (2) and (3) as direct consequences of line (1).  Let us
consider now line (2). Any formula of the form $x \wedge y$ can be
true if only both $x$ and $y$ are true. So we derive lines (4) and (5)
as direct consequences of line (2). Line (3) has the form $\neg (x
\vee y)$. A disjunction to be false needs to have both disjuncts
false. So lines (6) and (7) are added as direct consequences of line
(3). Let us now turn to line (5) which has the form $(x \vee y)$. We
cannot draw any direct conclusion about the truth value of $x$, nor
about the truth value of $y$; all we can infer is that either $x$ or
$y$ is true.  So the tableau {\em branches} into two columns. So line
(5) branches into two possibilities: line (8) and line (9). In the
same manner, line (6) gives two possibilities : $\neg p$ and $\neg
q$. Hence, the tableau now branches to four possibilities: (10), (11),
(12), and (13). Looking now at the leftmost branch we shall see that
(10) is a direct contradiction of (4), so we close this branch to
signify that it leads to a contradiction. Similarly (11) contradicts
(8), and (12) contradicts (4). So these branches are inconsistent too.
Going back now to formula (7) --- it is false when either $p$ or $r$
are false. Branches (10), (11) and (12) are already inconsistent so
there is no point to add these new alternatives. The only consistent
branch left is branch (13). Adding the new branches (14) and (15)
leads us to inconsistency in both cases as (14) contradicts (4) and
(15) contradicts (9). Thus all branches lead us to a contradiction, so
line (1) is untenable. Thus $[p \wedge (q \vee r)] \rightarrow [(p
\vee q) \wedge (p \vee r)]$ can never be false in any interpretation,
so it is a tautology.

The tableau method can be used as a theorem prover. We know that a set
of formulas $\Phi \entails \alpha$ iff $\{ \Phi, \neg \alpha \}$ is
inconsistent. Therefore, to prove that $\alpha$ follows from $\Phi$,
it is enough to construct a closed tableau having as root the formulas
$\Phi \cup \neg \alpha$.

In what follows, we will use the terminology related to classical
propositional tableaux given by~\cite{bell86}. We generalize the
classical  propositional tableaux, such that if a given {\em stratified
propositional tableau} $T$ for a given theory $\Phi_0$ has been
obtained, we are allowed to extend it into a new one $T'$ by any of
the following three rules. In each case $T'$ will have all the nodes
of $T$, plus one or two new nodes. 
\begin{itemize}
\item {\bf Rule $\neg \neg$}: If among the formulas of a branch of $T$
terminating at node $\Phi$ there is a formula $\neg\neg\alpha$, add a
new node $\{\alpha\}$ as successor to $\Phi$.

\item {\bf Rule $\rightarrow$}: If among the formulas of a branch of $T$
terminating at node $\Phi$ there is a formula $\alpha \rightarrow
\beta$, add two new  nodes $\{\neg \alpha\}$ and $\{\beta\}$ as
successors to $\Phi$. 

\item {\bf Rule $\neg \rightarrow$}: If among the formulas of a branch of $T$
terminating at node $\Phi$ there is a formula $\neg(\alpha
\rightarrow \beta)$, add a new  node $\{\alpha, \neg \beta\}$  as
successor to $\Phi$.  
\end{itemize}

A branch of a tableau is {\em x-closed} if there is a atomic formula
$\alpha$ such that both $\alpha^y$ and $\neg \alpha^t$ are formulas of
that branch, and $y,t \leq x$. In other words, a branch is {\em
x-closed} if the set of formulas that belong to that branch is not
{\em x-satisfiable}.  The formulas $\alpha^y$ and $\neg \alpha^t$ are
said to be {\em used} for closing the branch. A propositional tableau
for $\Phi$ is called an {\em x-confutation} of $\Phi$ if all its
branches are {\em x-closed}. To {\em x-confute} $\Phi$ is to construct
an {\em x-confutation} of $\Phi$.

\begin{figure}[hbt]
\centering
\input{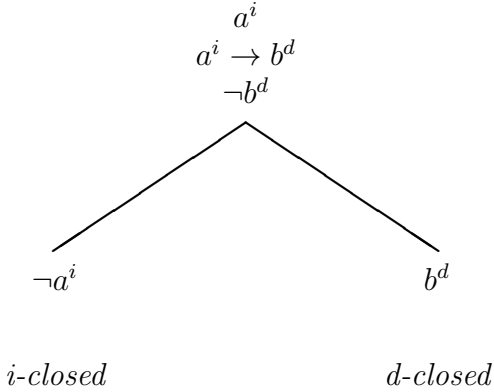}
\caption{A {\em d-closed} stratified tableau}
\label{strat-tabl-1}
\end{figure}

For example, the tableau given in figure~\ref{strat-tabl-1} has two
branches.  The left one is {\em i-closed} and the right one is {\em
d-closed}.  Thus, the whole tableau is {\em d-closed}. This follows
from corollary~\ref{cor-closed}.

\begin{corollary}
\label{cor-closed}
If a branch in a tableau is x-closed, that branch is also y-closed for
any $y \geq x$.
\end{corollary}

{\em Proof:} This is a direct consequence of
theorem~\ref{not-sat-cor}.

\begin{theorem} 
{\em Soundness of stratified propositional tableaux.} If a set of
stratified propositional formulas $\Phi$ have a finite x-closed
tableau, then $\Phi$ is x-unsatisfiable.
\end{theorem}

One can reformulate the soundness theorem as follows: If an extended
truth valuation $\sigma$ {\em x-satisfies} all the formulas of a given
branch in a tableau, and if that branch is extended into a new branch
(or extended and split into two new branches) by one of the rules,
then $\sigma$ also {\em x-satisfies} the new branch (or at least one
of the two branches).

{\em Proof:} Lemma~\ref{lemma-soundness} is  a stronger form of the
contrapositive of the theorem. Using it, the soundness follows immediately.

\begin{lemma} \label{lemma-soundness}
If $T$ is a tableau for $\Phi$ and $\sigma$ x-satisfies $\Phi$, then for
some branch of $T$, $\sigma$ x-satisfies all formulas on that branch
(hence, the branch is open).
\end{lemma} 

{\em Proof:} The proof is by induction on the height of the tableau
$T$ ($\Phi$ and $\sigma$ are fixed).

{\em Base case:} \newline
Assume that $T$ consists of the root alone. The result is trivial.

{\em Induction step:} 
\begin{itemize}

\item Assume that $\sigma$ {\em x-satisfies} all formulas on a branch of
height $i$ that contains $\neg \neg \alpha$. In accordance with the
tableau rule for negation, that branch can be extended with $\alpha$
into a branch of height $i+1$.
Definition~\ref{satisfaction-definition} assures one that this branch
is also {\em x-satisfiable}.

\item Assume that $\sigma$ {\em x-satisfies} all formulas on a branch of
height $i$ that contains $\alpha_1 \rightarrow \alpha_2$. In
accordance with the tableau rule for implication, this branch can be
extended into two branches of height $i+1$, one containing $\neg
\alpha_1$, the other containing $\alpha_2$.
Definition~\ref{satisfaction-definition} assures one that at least one
of these branches is also {\em x-satisfiable}.

\item Assume that $\sigma$ {\em x-satisfies} all formulas on a branch of
height $i$ that contains $\neg (\alpha_1 \rightarrow \alpha_2)$. In
accordance with the corresponding tableau rule, this branch can be
extended with two new formulas: $\alpha_1$ and  $\neg \alpha_2$.
Definition~\ref{satisfaction-definition} assures one that 
the extended branch is also {\em x-satisfiable}.
\end{itemize}

By induction, the lemma follows.

\begin{theorem}
{\em Completeness of stratified propositional tableaux.} If $\Phi$ is
x-unsatisfiable then $\Phi$ has a finite x-closed tableau.
\end{theorem}

{\em Proof: } To prove this theorem we need some extra definitions:

\begin{definition} \label{usedup1}
Let $\varphi$ be a formula on a branch $B$ of a tableau $T$. We say
$\varphi$ is {\em used up} on $B$ iff one of the following holds:
\begin{enumerate}
\item $\varphi$ is $\neg \neg \alpha$ and $\alpha$ is on branch $B$;
\item $\varphi$ is $\neg(\alpha \rightarrow \beta)$ and $\alpha, \neg
\beta$ are on $B$;
\item $\varphi$ is $\alpha \rightarrow \beta$ and either $\alpha$ or
$\beta$ is on $B$;
\item $\varphi$ is $P^x$ or $\neg P^y$ for some atoms $P$, where $x,y \in
\{u,i,d\}$. 
\end{enumerate}

The branch $B$ is {\em used up} iff all formulas on $B$ are {\em
used up}.

The tableau $T$ is {\em used up} or {\em exhausted} iff all its
branches are {\em used up}.  
\end{definition}

Completeness follows from lemmas~\ref{lemma-comp1} and~\ref{lemma-comp2}.

\begin{lemma} \label{lemma-comp1}
Every finite\footnote{A proof similar to the classical one can be
given for a countable $\Phi$ as well.} set of formulas $\Phi$ has a
tableau which is used up. 
\end{lemma}

{\em Proof: }
Consider an algorithm that starts with $\Phi$ as the single node of
the tree and at each step extends one of the branches that are not
used up in accordance with the tableau rules, i.e., applies these rules
for a formula $\alpha$ that is not used up.  Every formula generated
on $T$ is either  of the form $\beta$ or $\neg \beta$ where
$\beta$ is a sub-formula of $\alpha$. Thus, each branch has at most
$2k$ formulas where $k$ is the number of sub-formulas in $\Phi$. That
means the algorithm terminates. Its result is an exhausted tableau.

\begin{lemma} \label{lemma-comp2} 
If $T$ is a tableau for $\Phi$ with
a branch $B$ that is used up and is not x-closed, then the set of
formulas on $B$ is x-satisfiable. Therefore, $\Phi$ is x-satisfiable.
\end{lemma}

{\em Proof: } Each formula on $B$ is used up, so for each atomic
formula $P$, not both $P^y$ and $\neg P^z$ occur on $B$ where $y,z
\leq x$.  We define $P^\sigma = \top^m$ iff $P^y$ is a label on the
branch, $m = max(y,x)$ and for any $z$ such that $P^z$ or $\neg P^z$
occur on the branch $m \leq z$. We define $P^\sigma = \bot^m$ iff
$\neg P^y$ is a label on the branch, $m = max(y,x)$ and for any $z$
such that $P^z$ or $\neg P^z$ occur on the branch $m \leq z$.
Intuitively, we pick up the strongest labelled atomic formula, $P^t$
or $\neg P^t$ and we assign to $P$ the maximum value between $t$ and
$x$ where $x$ is the level for which we want to find an extended
valuation. Assume that $x \leq t$. In this case, in accordance with
definition~\ref{satisfaction-definition}, the truth assignment will
satisfy $P^t$ and any other weaker formula. If $x > t$, because there
is no other formula of the form $\neg P^z$ with $t \leq z \leq x$,
then the truth assignment will satisfy $P^t$ and any other formula
weaker than $x$.

The claim is that $\sigma \xentails \Phi$.
We prove this by structural induction on $\Phi$. 

{\em Base case:} \newline
If $\Phi$ is an atom or a negated atom, the result is trivial.

{\em Induction step:} 
\begin{itemize}
\item Assume that $\Phi = \neg \neg \alpha$. Then $\alpha$ occurs on
$B$. Following the definitions, any extended truth assignment that
{\em x-satisfies} $\alpha$ also {\em x-satisfies} $\neg \neg \alpha$.

\item Assume that $\Phi = \neg(\alpha \rightarrow \beta)$. Then both
$\alpha$ and $\neg \beta$ occur on branch $B$, because it is used up.
By induction hypothesis, the extended truth valuation {\em
x-satisfies} both $\alpha$ and $\neg \beta$. In accordance
with~\ref{satisfaction-definition}, the extended truth assignment {\em
x-satisfies} $\neg(\alpha \rightarrow \beta)$.

\item Assume that $\Phi = \alpha \rightarrow \beta$. Then either $\neg
\alpha$ or $\beta$ occurs on branch $B$, because it is used up.  By
induction hypothesis, the extended truth valuation {\em x-satisfies}
either $\neg \alpha$ or $\beta$. In accordance
with~\ref{satisfaction-definition}, the extended truth assignment {\em
x-satisfies} $\alpha \rightarrow \beta$.
\end{itemize}

\section{Stratified models and model ordering}

\begin{figure}[hbt]
\centering
\setlength{\unitlength}{0.012500in}%
\begingroup\makeatletter
% extract first six characters in \fmtname
\def\x#1#2#3#4#5#6#7\relax{\def\x{#1#2#3#4#5#6}}%
\expandafter\x\fmtname xxxxxx\relax \def\y{splain}%
\ifx\x\y   % LaTeX or SliTeX?
\gdef\SetFigFont#1#2#3{%
  \ifnum #1<17\tiny\else \ifnum #1<20\small\else
  \ifnum #1<24\normalsize\else \ifnum #1<29\large\else
  \ifnum #1<34\Large\else \ifnum #1<41\LARGE\else
     \huge\fi\fi\fi\fi\fi\fi
  \csname #3\endcsname}%
\else
\gdef\SetFigFont#1#2#3{\begingroup
  \count@#1\relax \ifnum 25<\count@\count@25\fi
  \def\x{\endgroup\@setsize\SetFigFont{#2pt}}%
  \expandafter\x
    \csname \romannumeral\the\count@ pt\expandafter\endcsname
    \csname @\romannumeral\the\count@ pt\endcsname
  \csname #3\endcsname}%
\fi
\endgroup
\begin{picture}(181,307)(9,275)
\thicklines
\put(120,480){\line(-2,-1){ 60}}
\put(120,480){\line( 2,-1){ 60}}
\put( 60,420){\line(-6,-5){ 40.820}}
\put( 60,420){\line( 6,-5){ 40.820}}
\put(100,360){\vector( 0,-1){ 25}}
\put( 20,360){\vector( 0,-1){ 65}}
\put(120,565){\makebox(0,0)[b]{\smash{\SetFigFont{12}{14.4}{rm}$a \vee b$}}}
\put(120,545){\makebox(0,0)[b]{\smash{\SetFigFont{12}{14.4}{rm}$\neg(a \rightarrow b)$}}}
\put(120,525){\makebox(0,0)[b]{\smash{\SetFigFont{12}{14.4}{rm}$a \vee c$}}}
\put(120,505){\makebox(0,0)[b]{\smash{\SetFigFont{12}{14.4}{rm}$a$}}}
\put(120,485){\makebox(0,0)[b]{\smash{\SetFigFont{12}{14.4}{rm}$\neg b$}}}
\put( 60,430){\makebox(0,0)[b]{\smash{\SetFigFont{12}{14.4}{rm}$a$}}}
\put(180,430){\makebox(0,0)[b]{\smash{\SetFigFont{12}{14.4}{rm}$b$}}}
\put( 20,365){\makebox(0,0)[b]{\smash{\SetFigFont{12}{14.4}{rm}$a$}}}
\put(100,365){\makebox(0,0)[b]{\smash{\SetFigFont{12}{14.4}{rm}$c$}}}
\put(180,390){\makebox(0,0)[b]{\smash{\SetFigFont{12}{14.4}{rm}X}}}
\put(105,320){\makebox(0,0)[lb]{\smash{\SetFigFont{12}{14.4}{rm}$m_1 = \{ a, \neg b, c \}$}}}
\put( 20,275){\makebox(0,0)[lb]{\smash{\SetFigFont{12}{14.4}{rm}$m_0 = \{a, \neg b \}$}}}
\end{picture}
\caption{Semantic tableaux as means for building models}
\label{tableau-models}
\end{figure}
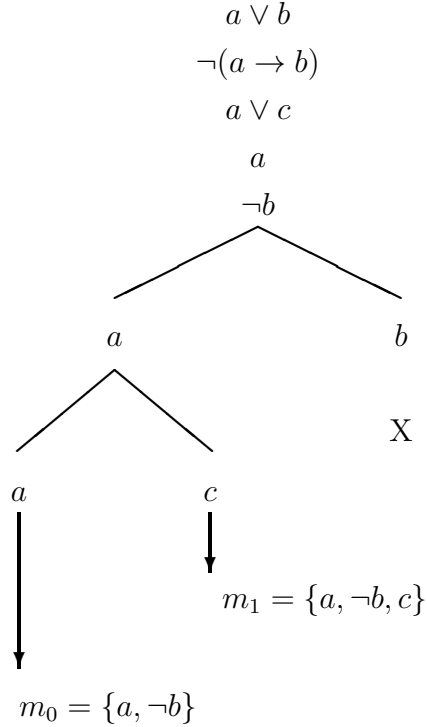

As seen, propositional tableaux can be used as theorem provers, but
there is another way one can look at them. Consider the following
classical propositional theory: $\{ a \vee b, \neg (a \rightarrow b),
a \vee c \}$.  If one constructs the tableau for this theory (see
figure~\ref{tableau-models}), she will obtain two open branches.
Collecting the atomic formulas on each branch yields two model schemata
for the given theory. For example, the branch that yields model schema
$m_0$ contains atomic formulas $a,a,\neg b$, and $a$. Removing the
duplicates, we obtain the model schema $m_0 = \{a,\neg b\}$. The branch
ending in X contains both $b$ and $\neg b$; therefore, it is closed,
i.e., there is no way to construct a model for the formulas on that
branch. 

Observe that collecting the atomic formulas on each branch does not
provide a model for the theory, but rather a model schema, because it
may be the case that some of the atomic formulas that occur in that
theory do not belong to the branch. In our example, model $m_0$
contains no reference to the atomic formula $c$. This means that the
choice of the truth assignment for $c$ has no impact on the
satisfiability relation: both $\{a/\top,b/\bot,c/\top\}$ and
$\{a/\top,b/\bot,c/\bot\}$ will satisfy the initial theory. 
Thus, the model schemata provide a compact representation for all the
possible models of a theory.

The same idea will be applied to stratified semantic tableaux.
Consider a classical example from the default reasoning literature,
where the system knows that {\em Tweety is a bird} and {\em Birds
typically fly} and would like to conclude that {\em Tweety flies}.  As
known, typicality is not representable in first-order logic, because
a formula such as 
\[ (\forall x)(bird(x) \rightarrow flies(x))\] 
yields incorrect results for penguins or ostriches for example, and a formula
such as  
\[(\forall x)(bird(x) \wedge \neg penguin(x) \wedge \neg
ostrich(x) \wedge \ldots \rightarrow flies(x))\] presupposes that one
has to prove first that Tweety is not a penguin or an ostrich in order
to prove that Tweety flies (see
McCarthy~\shortcite{mccarthy80,mccarthy86} for a detailed discussion).

\subsection{Why does Tweety fly?}

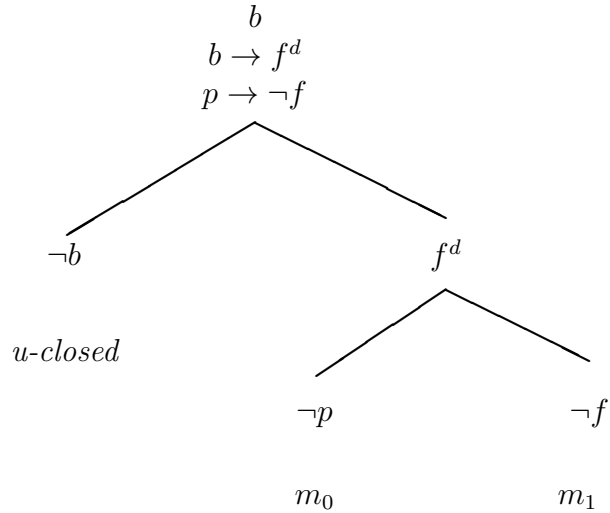
\begin{figure}[hbt]
\centering
\setlength{\unitlength}{0.012500in}%
\begingroup\makeatletter
% extract first six characters in \fmtname
\def\x#1#2#3#4#5#6#7\relax{\def\x{#1#2#3#4#5#6}}%
\expandafter\x\fmtname xxxxxx\relax \def\y{splain}%
\ifx\x\y   % LaTeX or SliTeX?
\gdef\SetFigFont#1#2#3{%
  \ifnum #1<17\tiny\else \ifnum #1<20\small\else
  \ifnum #1<24\normalsize\else \ifnum #1<29\large\else
  \ifnum #1<34\Large\else \ifnum #1<41\LARGE\else
     \huge\fi\fi\fi\fi\fi\fi
  \csname #3\endcsname}%
\else
\gdef\SetFigFont#1#2#3{\begingroup
  \count@#1\relax \ifnum 25<\count@\count@25\fi
  \def\x{\endgroup\@setsize\SetFigFont{#2pt}}%
  \expandafter\x
    \csname \romannumeral\the\count@ pt\expandafter\endcsname
    \csname @\romannumeral\the\count@ pt\endcsname
  \csname #3\endcsname}%
\fi
\endgroup
\begin{picture}(267,211)(212,560)
\thicklines
\put(320,720){\line( 2,-1){ 80}}
\put(400,650){\line(-3,-2){ 54.231}}
\put(400,650){\line( 2,-1){ 60}}
\put(320,760){\makebox(0,0)[b]{\smash{\SetFigFont{12}{14.4}{rm}$b$}}}
\put(320,744){\makebox(0,0)[b]{\smash{\SetFigFont{12}{14.4}{rm}$b \rightarrow 
f^d$}}}
\put(320,728){\makebox(0,0)[b]{\smash{\SetFigFont{12}{14.4}{rm}$p \rightarrow
\neg f$}}}
\put(320,720){\line(-5,-3){ 78.677}}
\put(240,660){\makebox(0,0)[b]{\smash{\SetFigFont{12}{14.4}{rm}$\neg b$}}}
\put(455,560){\makebox(0,0)[b]{\smash{\SetFigFont{12}{14.4}{tt}$m_1$}}}
\put(400,660){\makebox(0,0)[b]{\smash{\SetFigFont{12}{14.4}{rm}$f^d$}}}
\put(345,595){\makebox(0,0)[b]{\smash{\SetFigFont{12}{14.4}{rm}$\neg p$}}}
\put(460,595){\makebox(0,0)[b]{\smash{\SetFigFont{12}{14.4}{rm}$\neg f$}}}
\put(240,620){\makebox(0,0)[b]{\smash{\SetFigFont{12}{14.4}{tt}{\em u-closed}}}}
\put(345,560){\makebox(0,0)[b]{\smash{\SetFigFont{12}{14.4}{tt}$m_0$}}}
\end{picture}
\caption{A propositional stratified tableau for theory {\em Tweety is a bird. Typically birds fly. Penguins do not fly.}}
\label{tweety-tableau-1}
\end{figure}

Consider the following example: {\em Tweety is a bird. Typically,
birds fly. Penguins do not fly. Does Tweety fly?}
An appropriate formalization in stratified propositional logic and its
corresponding semantic tableau (figure~\ref{tweety-tableau-1}) follows:

\vspace{5mm}
\begin{center} \begin{tabular}{ll}
$b$ & Tweety is a bird. \\
$b \rightarrow f^d$ & Typically birds fly (this is
defeasible information). \\
$p \rightarrow \neg f$ & Penguins do not fly. \\
\end{tabular} \end{center}
\vspace{5mm}

\noindent The semantic tableau has two open branches, which give two model
skeletons or schemata: 

\vspace{5mm} \noindent
{\small
\begin{center} \begin{tabular}{|l|l|l|l|} \hline \hline
{\em Schema \#} & {\em Indefeasible} & {\em Infelicitously defeasible} & {\em
Felicitously defeasible} \\ \hline 
$m_0$ & $b$      & &       \\
      & $\neg p$ & &       \\
      &          & & $f^d$ \\ \hline
$m_1$ & $b$      & &       \\
      & $\neg f$ & & $f^d$ \\ \hline
\end{tabular} \end{center}
}
\vspace{5mm}

\noindent In model $m_0$, one can conjecture the fact that Tweety
flies, but she cannot do this in model $m_1$. If we want an extended
truth assignment to satisfy the skeleton $m_1$, we are constrained to
assign $f$ the truth value $\bot$. This is not the case with model
skeleton $m_0$ which can be satisfied in a more ``optimistic'' way,
assigning for example $\topd$ to $f$. In this sense, from an
optimistic perspective, $m_0$ is the preferred model for this theory.
A formal definition for the model ordering relation is given
in~\ref{prop-optimistic-def}.

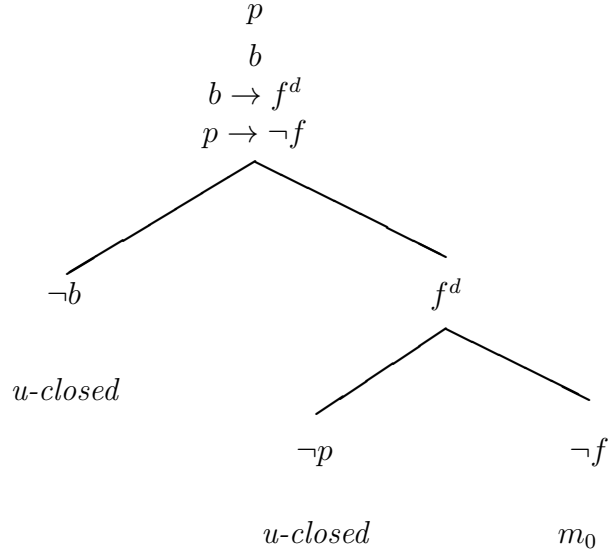
\begin{figure}[hbt]
\centering
\setlength{\unitlength}{0.012500in}%
\begingroup\makeatletter
% extract first six characters in \fmtname
\def\x#1#2#3#4#5#6#7\relax{\def\x{#1#2#3#4#5#6}}%
\expandafter\x\fmtname xxxxxx\relax \def\y{splain}%
\ifx\x\y   % LaTeX or SliTeX?
\gdef\SetFigFont#1#2#3{%
  \ifnum #1<17\tiny\else \ifnum #1<20\small\else
  \ifnum #1<24\normalsize\else \ifnum #1<29\large\else
  \ifnum #1<34\Large\else \ifnum #1<41\LARGE\else
     \huge\fi\fi\fi\fi\fi\fi
  \csname #3\endcsname}%
\else
\gdef\SetFigFont#1#2#3{\begingroup
  \count@#1\relax \ifnum 25<\count@\count@25\fi
  \def\x{\endgroup\@setsize\SetFigFont{#2pt}}%
  \expandafter\x
    \csname \romannumeral\the\count@ pt\expandafter\endcsname
    \csname @\romannumeral\the\count@ pt\endcsname
  \csname #3\endcsname}%
\fi
\endgroup
\begin{picture}(267,232)(212,560)
\thicklines
\put(320,720){\line( 2,-1){ 80}}
\put(400,650){\line(-3,-2){ 54.231}}
\put(400,650){\line( 2,-1){ 60}}
\put(320,760){\makebox(0,0)[b]{\smash{\SetFigFont{12}{14.4}{rm}$b$}}}
\put(320,744){\makebox(0,0)[b]{\smash{\SetFigFont{12}{14.4}{rm}$b \rightarrow f^d$}}}
\put(240,660){\makebox(0,0)[b]{\smash{\SetFigFont{12}{14.4}{rm}$\neg b$}}}
\put(320,720){\line(-5,-3){ 78.677}}
\put(400,660){\makebox(0,0)[b]{\smash{\SetFigFont{12}{14.4}{rm}$f^d$}}}
\put(345,560){\makebox(0,0)[b]{\smash{\SetFigFont{12}{14.4}{tt}{\em u-closed}}}}
\put(345,595){\makebox(0,0)[b]{\smash{\SetFigFont{12}{14.4}{rm}$\neg p$}}}
\put(460,595){\makebox(0,0)[b]{\smash{\SetFigFont{12}{14.4}{rm}$\neg f$}}}
\put(240,620){\makebox(0,0)[b]{\smash{\SetFigFont{12}{14.4}{tt}{\em u-closed}}}}
\put(455,560){\makebox(0,0)[b]{\smash{\SetFigFont{12}{14.4}{tt}$m_0$}}}
\put(320,728){\makebox(0,0)[b]{\smash{\SetFigFont{12}{14.4}{rm}$p \rightarrow \neg f$}}}
\put(320,780){\makebox(0,0)[b]{\smash{\SetFigFont{12}{14.4}{tt}$p$}}}
\end{picture}
\caption{A propositional stratified tableau for theory: {\em Tweety is
a bird and a penguin. Typically birds fly. Penguins do not fly.}} 
\label{tweety-tableau-2}
\end{figure}

If we learn that Tweety is a penguin, the branch that gave the model
$m_0$ will be {\em u-closed}, so that the only model for the theory can be
constructed assigning $\bot$ to $f$. Hence, Tweety will no longer be
thought to fly (figure~\ref{tweety-tableau-2}).

\vspace{5mm} \noindent
{\small
\begin{center} \begin{tabular}{|l|l|l|l|} \hline \hline
{\em Schema \#} & {\em Indefeasible} & {\em Infelicitously defeasible} & {\em
Felicitously defeasible} \\ \hline 
$m_0$ & $b$      & &       \\
      & $p$      & &       \\
      & $\neg f$ & & $f^d$ \\ \hline
\end{tabular} \end{center}
}
\vspace{5mm}

\section{Model ordering}

We are interested in giving a formal account for the notion of
preferred or {\em optimistic} model. We accomplish this by formalizing
the following two intuitions. 
\begin{enumerate}
\item Assume that one utters: 
\bexample{If the weather is good we can go to the beach tomorrow. We
can swim, lie in the sun, and build sand castles. We can have a lot of
fun.}  
There are two ways one can make the utterance true. The first one is
to assign falsity to the antecedent, while the second one is to assume
that the antecedent holds, therefore, the consequent holds too. If the
antecedent is false we cannot say too much about the corresponding
interpretation. However, if the antecedent is true, there are many
other things that can be said within the corresponding interpretation.
In other words, the second interpretation is more informative than the
first one, so an {\em optimistic} agent will prefer it.

\item Assume that one knows that {\em Tweety is a bird} and that {\em
Typically birds fly}. It may be the case that a rational agent finds
only later that Tweety is a penguin, but as long as she doesn't have
this information she is still inclined to believe that {\em Tweety
flies}. In other words, an {\em optimistic} agent will prefer to trigger
as many inferences as possible even though some of them may be
retracted in the future.
\end{enumerate}
   
Let us analyze the first example again, where we know nothing about
Tweety being a penguin. As seen, the theory yields two model schemata
for this case. There are two different ways one can analyze and
compare these models. On one hand, notice that model $m_0$ is more
informative than model $m_1$, i.e., it contains information about $p$,
while $m_1$ does not.  On the other hand notice that model $m_0$ puts
no constraints on Tweety's flying ($f$). It assumes that Tweety flies,
and this information is labelled as defeasible. The second model
contains also the felicitously defeasible information that Tweety
flies, but it also contains the indefeasible information that Tweety
does not fly. This indefeasible information overlaps the defeasible
inference and prevents a rational agent from finding an extented truth
assignment that {\em d-satisfies} the initial theory: in other words,
the second model is less optimistic.
Formal definitions for the model ordering follow.

\begin{definition}
An atomic formula $a$ is said to be {\em affirmative}\footnote{If
$t=x$, no judgment can be made with respect to the polarity of the
formula that is discussed. Finding a reasonable solution for this
problem is the key issue in choosing among interactive defaults that
have the same strength.} in a model schema $m$, if $a^x$ is a member
of $m$ and for any $a^t$ or $\neg a^t$ occurring in the model schema,
$t > x$.
\end{definition}

\begin{definition}
An atomic formula $a$ is said to be {\em negative} in a model
schema $m$, if $\neg a^x$ is a member of $m$ and for any
$a^t$ or $\neg a^t$ occurring in the model schema, $t > x$.
\end{definition}

\begin{definition}
An atomic formula is {\em self-cancellable} in a model schema $m$, if
both $a^x$ and $\neg a^y$ occur in $m$, where $x,y \in \{ u,i,d \}$.
\end{definition}

\begin{definition}
A model schema $m_1$ is more {\em informative} than a model schema
$m_2$ iff for any positive or atomic formula $a^x$ in $m_2$, there is a
positive or negative atomic formula $a^y$ or $\neg a^y$ in $m_1$,
where $x,y \in \{ u,i,d \}$.
\end{definition}

\begin{definition}
An atomic formula $a^x$ is {\em weaker} than $a^y$ iff $x$ and $y$ are
members of the set $\{u,i,d\}$ and $x \geq y$ ($u < i < d$).
\end{definition}

\begin{definition}
A model schema $m_1$ is {\em weaker} than a model schema $m_2$ if
for every non-self-cancellable atomic formula $a^x$ in $m_2$ there is a
corresponding $a^y$ formula in $m_1$ that is weaker and has the same
polarity (they are both positive or negative).
\end{definition}

\begin{definition} \label{prop-optimistic-def}
A model schema $m_1$ is more {\em optimistic} than a model $m_2$ if it is more
informative and weaker.
\end{definition}

\begin{definition}
The set of {\em optimistic model schemata} for an exhausted tableau
are given by the set of most optimistic and weakest models in the
partial order defined by the above two relations.
\end{definition}

Intuitively, this corresponds to having a truth-valuation with as much
defeasible information as possible. For example, for the atomic
formula $a^d$, $\{a/\topd\}$ is an optimistic valuation, while
$\{a/\top\}$ and $\{a/\bot\}$ are not.

If we reconsider now the example involving Tweety, $m_0$ is more
optimistic than $m_1$ and this corresponds to our intuitions.

\section{First-order stratified logic}

We now discuss the first-order extension of stratified logic. The
reader is referred to~\cite{bell86} for a comprehensive study
of classical first-order logic. 

\subsection{Syntax}

In order to interpret stratified terms and formulas, it is necessary to
fix a stratified structure or stratified interpretation  ${\cal SL}$
consisting of the following elements: 
\begin{itemize}
\item A non-empty class $\cal D$ called the {\em universe of
discourse}, or the {\em domain}.
\item A mapping that assigns to each function symbol $f$  of ${\cal
SL}$ an operation $f^{\cal D}$ on $\cal D$ such that if $f$ is an $n$-ary
function symbol, $f^{\cal D}$ is an $n$-ary operation on $\cal D$. The
functions of arity $0$ give the constants of the language. Let $F$ be
the set of all these mappings.

\item A mapping that assigns to each extra-logical predicate symbol
$P$ of ${\cal L}$ a relation $P^{\cal D}$ on $\cal D$ such that if $P$
is an $n$-ary predicate symbol, $P^{\cal D}$ is an $n$-ary relation on
$\cal D$.  We are interested in extending the stratification found in
the propositional case to predicates; hence we consider that relations
$P$ can be assigned a strength (indefeasible, infelicitously
defeasible, and felicitously defeasible relations) and a polarity
(positive and negative relations). Thus, the set of relations $R$ will
be given by the union $R^u \cup \overline{R^u} \cup R^i \cup
\overline{R^i} \cup R^d \cup \overline{R^d}$ where $R^u$ stands for
positive indefeasible relations, $\overline{R^u}$ for negative
indefeasible relations, $R^i$ for positive infelicitously defeasible
relations, $\overline{R^i}$ for negative infelicitously defeasible
relations, $R^d$ for positive felicitously defeasible relations, and
$\overline{R^d}$ for negative felicitously defeasible relations.  
\end{itemize}

It is important to notice that the stratification is embedded into the
predicate level; hence,  predicates are stronger or weaker. There is
no strength ordering at the function or constant level. Another
difference with respect to first-order logic is that the relations
found in a stratified interpretation may have negative polarity. This
is a consequence of the way we have defined the satisfiability relation.

The terms are recursively defined as follows:
\begin{enumerate}
\item Any constant or variable is a term.
\item Any expression $f(t_1,t_2,...,t_n)$ is a term iff $f \in F$ and
each $t_i$ is a term.
\end{enumerate}

The formulae of stratified first-order logic are recursively defined
as follows:
\begin{enumerate}
\item Each member of the set $\{ \topu, \botu, \topi, \boti, \topd,
\botd \}$ is a formula.
\item Each expression $p^x(t_1,t_2,...,t_n)$ where $p \in R$, $x
\in \{u,i,d \}$, and $t_i$ are terms is an atomic formula.
\item Each combination of atomic formulae using the logical
connectives $\neg, \rightarrow, \vee, \wedge$, and $\leftrightarrow$
is a formula.
\item If $f$ is a formula and $x$ is a variable then $(\forall x)f$
and $(\exists x)f$ are formulas. 
\end{enumerate}

For example, $(\forall x,y,z)(regret(x,come(y,z)) \rightarrow
come^i(y,z))$ or $(\forall x)(\neg bachelor(x) \rightarrow
adult^d(x))$ are well-formed formulas.  The first one expresses that
factive {\em regret} in a positive environment implies its complement.
The implication is infelicitously defeasible in this case, because one
cannot felicitously utter {\em John regrets that Mary came to the
party, but she did not come}. The latter expresses that a {\em
bachelor} referent in a negative environment implies that that
referent is also an adult. The implication is felicitously
defeasible, because one can felicitously utter {\em John is not a
bachelor; he is five years old}.

\subsection{Semantics of stratified first-order logic}

\begin{definition}
An \ul{${\cal SL}$ valuation} $\sigma$ is a stratified
structure ${\cal SL}$ together with an assignment of the value
$x^\sigma \in \cal D$ to each variable $x$.
\end{definition}

\begin{definition} \label{term-definition}
Given an ${\cal SL}$ valuation $\sigma$ with universe $\cal D$, we define
for each term $t$, the value of $t$ under $\sigma$ ($t^\sigma$) in such
a way that $t^\sigma \in \cal D$:
\begin{itemize}
\item If $x$ is a variable, then $x^\sigma$ is already defined.
\item If $f$ is an n-ary function symbol in $F$ and
$t_1,t_2,\ldots,t_n$ are terms, then \\ $(f(t_1,t_2,\ldots,t_n))^\sigma =
f^\sigma(t_1^\sigma,t_2^\sigma,\ldots,t_n^\sigma)$
\end{itemize}
\end{definition}

\begin{definition} \label{xentailsfol}
Assume $\sigma$ is an  ${\cal SL}$ valuation such that $t_i^\sigma =
d_i \in {\cal D}$ and assume that ${\cal SL}$  maps $n$-ary predicates
$p$ to relations $R \subset {\cal D} \times \ldots \times {\cal D}$.  
For any atomic formula $p^x(t_1,t_2,\ldots,t_n)$, and any stratified
valuation $\sigma$, where $x \in \{u,i,d\}$ and
$t_i$ are terms, the x-satisfiability relations are
defined as follows:

\begin{itemize}
\item $\sigma \entails p(t_1,t_2,\ldots,t_n)$ iff
$\langle d_1, d_2, \ldots, d_n \rangle \in R^u$ 
\item $\sigma \entails p^i(t_1,t_2,\ldots,t_n)$ iff
$\langle d_1, d_2, \ldots, d_n \rangle \in R^u \cup \overline{R^u}
\cup R^i$ 
\item $\sigma \entails p^d(t_1,t_2,\ldots,t_n)$ iff
$\langle d_1, d_2, \ldots, d_n \rangle \in R^u \cup \overline{R^u}
\cup R^i \cup \overline{R^i} \cup R^d$
\item $\sigma \ientails p(t_1,t_2,\ldots,t_n)$ iff
$\langle d_1, d_2, \ldots, d_n \rangle \in R^i$ 
\item $\sigma \ientails p^i(t_1,t_2,\ldots,t_n)$ iff
$\langle d_1, d_2, \ldots, d_n \rangle \in R^i$ 
\item $\sigma \ientails p^d(t_1,t_2,\ldots,t_n)$ iff
$\langle d_1, d_2, \ldots, d_n \rangle \in R^i \cup \overline{R^i}
\cup R^d$ 
\item $\sigma \dentails p(t_1,t_2,\ldots,t_n)$ iff
$\langle d_1, d_2, \ldots, d_n \rangle \in R^d$ 
\item $\sigma \dentails p^i(t_1,t_2,\ldots,t_n)$ iff
$\langle d_1, d_2, \ldots, d_n \rangle \in R^d$ 
\item $\sigma \dentails p^d(t_1,t_2,\ldots,t_n)$ iff
$\langle d_1, d_2, \ldots, d_n \rangle \in R^d $
\end{itemize}

For any negation of an atomic formula $\neg p^x(t_1,t_2,\ldots,t_n)$,
and any stratified valuation $\sigma$, where $x
\in \{u,i,d\}$ and $t_i$ are terms, the x-satisfiability relations are
defined as follows: 

\begin{itemize}
\item $\sigma \entails \neg p(t_1,t_2,\ldots,t_n)$ iff
$\langle d_1, d_2, \ldots, d_n \rangle \in \overline{R^u}$
\item $\sigma \entails \neg  p^i(t_1,t_2,\ldots,t_n)$ iff
$\langle d_1, d_2, \ldots, d_n \rangle \in R^u \cup \overline{R^u}
\cup \overline{R^i}$ 
\item $\sigma \entails \neg  p^d(t_1,t_2,\ldots,t_n)$ iff
$\langle d_1, d_2, \ldots, d_n \rangle \in R^u \cup \overline{R^u}
\cup R^i \cup \overline{R^i} \cup \overline{R^d}$
\item $\sigma \ientails \neg  p(t_1,t_2,\ldots,t_n)$ iff
$\langle d_1, d_2, \ldots, d_n \rangle \in \overline{R^i}$ 
\item $\sigma \ientails \neg  p^i(t_1,t_2,\ldots,t_n)$ iff
$\langle d_1, d_2, \ldots, d_n \rangle \in \overline{R^i}$ 
\item $\sigma \ientails \neg  p^d(t_1,t_2,\ldots,t_n)$ iff
$\langle d_1, d_2, \ldots, d_n \rangle \in R^i \cup \overline{R^i}
\cup \overline{R^d}$ 
\item $\sigma \dentails \neg  p(t_1,t_2,\ldots,t_n)$ iff
$\langle d_1, d_2, \ldots, d_n \rangle \in \overline{R^d}$ 
\item $\sigma \dentails \neg  p^i(t_1,t_2,\ldots,t_n)$ iff
$\langle d_1, d_2, \ldots, d_n \rangle \in \overline{R^d}$ 
\item $\sigma \dentails \neg  p^d(t_1,t_2,\ldots,t_n)$ iff
$\langle d_1, d_2, \ldots, d_n \rangle \in \overline{R^d}$
\end{itemize}
\end{definition}

For example, $\sigma \entails p^d(t_1,t_2,\ldots,t_n)$ iff
$\langle d_1,d_2,\ldots,d_n \rangle \in R^u \cup \overline{R^u} \cup R^i \cup
\overline{R^i} \cup R^d$ should be read as: ${\cal SL}$ valuation
$\sigma$ {\em u-satisfies} the atomic formula
$p^d(t_1,t_2,\ldots,t_n)$ if and only if the mapping that is defined
for the predicate symbol $p$ assigns to $p$ one and only one relation
from the union $R^u \cup \overline{R^u} \cup R^i \cup \overline{R^i}
\cup R^d$.

For the purpose of this thesis, the predicate mappings will be encoded
as lists of predicates of a given strength, optionally preceded by
negation. If for example $\langle 1 \rangle \in p^u$, $\langle 2
\rangle \in p^u$, and $\langle 1 \rangle \in \overline{q^i}$,  we will
represent it as $\{ p(1),p(2),\neg q^i(1) \}$.

Let $\sigma$ be a stratified valuation with universe $\cal D$ and let
$u \in \cal D$. We define $\sigma(x/u)$ to be a stratified valuation
which agrees with $\sigma$ on every variable other than $x$ as well as
on every extra-logical symbol, while $x^{\sigma(x/u)} = u$.  Thus, the
stratified structure underlying $\sigma(x/u)$ is the same as that
underlying $\sigma$.

\begin{definition} \label{formula-definition}
A stratified valuation ${\sigma}$ x-satisfies a formula $\alpha$ iff
\begin{itemize}
\item $\alpha$ is an atomic formula or the negation of an atomic
formula and $\sigma \xentails \alpha$, as specified
in definition~\ref{xentailsfol};
\item $\sigma \xentails \alpha_1$ where $\alpha = \neg \neg  \alpha_1$; 
\item $\sigma \xentails \neg \alpha_1$ or $\sigma \xentails 
\alpha_2$ where $\alpha = \alpha_1 \rightarrow \alpha_2$; 
\item $\sigma \xentails \alpha_1$ and $\sigma \xentails \neg
\alpha_2$ where $\alpha = \neg (\alpha_1 \rightarrow \alpha_2)$; 
\item $\sigma(v/t) \xentails \alpha_1$ for all $t \in
\cal D$, where $\alpha = \forall v \alpha_1$ and $\sigma(v/t)$ is the
same as $\sigma$ except $v^{\sigma(v/t)} = t$;
\item $\sigma(v/t) \xentails \alpha_1$ for at least one $t \in
\cal D$, where $\alpha = \exists v \alpha_1$ and $\sigma(v/t)$ is the
same as $\sigma$ except $v^{\sigma(v/t)} = t$;
\item $\sigma(v/t) \xentails \neg \alpha_1$ for at least one $t \in
\cal D$, where $\alpha = \neg \forall v \alpha_1$ and $\sigma(v/t)$ is the
same as $\sigma$ except $v^{\sigma(v/t)} = t$;
\item $\sigma(v/t) \xentails \neg \alpha_1$ for all $t \in
\cal D$, where $\alpha = \neg \exists v \alpha_1$ and $\sigma(v/t)$ is the
same as $\sigma$ except $v^{\sigma(v/t)} = t$.
\end{itemize}
\end{definition}
As in the propositional case, the other logical connectors are not
taken into consideration in our proofs, because they are expressible
in terms of negation and implication.

\begin{definition}
For any stratified formula $\alpha$ and any term $t$, $\alpha(x/t)$ is the
formula that results from replacing all free occurrences of $x$ in
$\alpha$ by the term $t$.
\end{definition}

\begin{theorem}
{\em Substitution theorem for terms.} For any term $s$,
$(s(x/t))^\sigma = s^{\sigma(x/t^\sigma)}$. 
\end{theorem}

{\em Proof: } The theorem follows immediately using induction on $s$
and definition~\ref{term-definition}.

\begin{definition}
A term $t$ is free for $x$ in $\alpha$ iff no free occurrence of $x$ in
$\alpha$ is within a sub-formula $\forall y \beta$, where $y$ occurs in
$t$. 
\end{definition}

\begin{theorem} \label{equiv-terms}
Let $t$ be a term, and let $\sigma$ and $\tau$ be stratified
valuations which agree on all variables and function symbols occurring
in $t$. Then $t^\sigma = t^\tau$.
\end{theorem}

The proof is straightforward if one uses induction on the degree of
the term $t$.

\begin{theorem} \label{equiv-valuations}
Let $\sigma$ and $\tau$ be stratified valuations which have the same
universe $\cal D$ and which agree on every free variable of $\alpha$ and on
every extra-logical symbol. Then $\sigma \xentails \alpha$ iff $\tau
\xentails \alpha$.
\end{theorem}

The proof is by induction on the degree of $\alpha$.

\begin{theorem} \label{subst-theorem}
{\em Substitution theorem for formulas.} For any
stratified valuation $\sigma$, formula $\alpha$ and term $t$ free of
$v$ in $\alpha$, $\sigma \xentails \alpha(v/t)$ iff
$\sigma(v/t^\sigma) \xentails \alpha$. 
\end{theorem}

{\em Proof: } We use induction on the degree of $\alpha$. The only
case that may raise problems is $\alpha = \forall y \beta$:

Suppose that $v$ is not free in $\alpha$. Then $\alpha(v/t) = \alpha$. 
By theorem~\ref{equiv-valuations}, $\sigma \xentails \alpha$ iff
$\sigma(v/t^\sigma) \xentails \alpha$. Thus $\sigma \xentails
\alpha(v/t)$ iff $\sigma \xentails \alpha$ iff  $\sigma(v/t^\sigma)
\xentails \alpha$.

Now suppose that $v$ is free in $\alpha$ and $t$ is free for $v$ in
$\beta$ and $y$ does not occur in $t$. Then we have
\begin{equation}
 \sigma \xentails \alpha(v/t) \mbox{ iff } \sigma \xentails
(\forall y [\beta(v/t)]) \label{eq1} 
\end{equation}
Using definition~\ref{formula-definition} we have that 
\begin{equation}
  \sigma \xentails (\forall y [\beta(v/t)]) \mbox{  iff  }
\sigma(y/u) \xentails \beta(v/t) \mbox{ for all $u$ in $\cal D$}
\label{eq2} 
\end{equation}
Since the degree of $\beta$ is smaller than the degree of $\alpha$,
using the induction hypothesis, we get: 
\begin{equation}
\sigma(y/u) \xentails \beta(v/t) \mbox{ iff }
\sigma(y/u)(v/t') \xentails \beta \mbox{ where } t' = t^{\sigma(y/u)}
\label{eq3}
\end{equation}
We know that $y$ does not occur in $t$. Hence by
theorem~\ref{equiv-terms} 
\begin{equation} t' = t^{\sigma(y/u)} = t^\sigma \end{equation}
At the same time, $v$ and $y$ are different (otherwise $v$ could not
be free in $\alpha$); hence
\begin{equation} 
\sigma(y/u)(v/t^\sigma) = \sigma(v/t^\sigma)(y/u) 
\end{equation}
Hence we can rewrite~\ref{eq3} as 
\begin{equation} 
\sigma(y/u) \xentails \beta(v/t) \mbox{ iff } \sigma(v/t^\sigma)(y/u)
\xentails \beta  
\end{equation} 
Definition~\ref{formula-definition} yields:
\begin{equation}
\sigma(v/t^\sigma)(y/u) \xentails \beta \mbox{ for all } u \in {\cal D},
\mbox{ iff } \sigma(v/t^\sigma) \xentails \forall y \beta \label{eq4} 
\end{equation}
Combining~\ref{eq1},~\ref{eq2}, and~\ref{eq4} we get the required result.

\subsection{Stratified first-order tableaux}

We now generalize the semantic tableau technique to stratified logic.
Preserving Smul-lyan's notation for conjunctive, disjunctive, universal,
and existential types of formulas, the rules for constructing a
stratified tableau are the following:

\begin{itemize}

\item {\em Conjunctive type}: If $\alpha$ appears in the tableau, both
$\alpha_{1}$ and $\alpha_{2}$ are added:

\vspace{5mm} 
\begin{center} \begin{tabular}{|c|c c|}
\hline

% head of the table
$\alpha$        & $\alpha_{1}$  & $\alpha_{2}$  \\ \hline

$p \wedge q$ & $p$ & $q$ \\
$\neg(p \vee q)$   & $\neg p$ & $\neg q$ \\
$\neg (p \rightarrow q)$ & $p$ & $\neg q$ \\ 
$\neg \neg p$      & $p$ & \\ \hline
\end{tabular} \end{center}
\vspace{5mm}

\item {\em Disjunctive type}: If $\beta$ appears in the tableau, it is
split into two branches, one headed by $\beta_{1}$ , the other by
$\beta_{2}$: 
 
\vspace{5mm} 
\begin{center} \begin{tabular}{|c|c|c|}
\hline

% head of the table
$\beta$         & $\beta_{1}$   & $\beta_{2}$  \\ \hline

$\neg (p \wedge q)$ & $\neg p$ & $\neg q$ \\
$p \vee q$   & $p$ & $q$ \\
$p \rightarrow q$ & $\neg p$ & $q$ \\ \hline
\end{tabular} \end{center}
\vspace{5mm}

\item{\em Universal type}: An universal formula may be instantiated with an 
arbitrary constant. In systematically applying it, one chooses all of the
constants that have already appeared in the tableau.
\label{univ-quant-rule}

\vspace{5mm}
\begin{center} \begin{tabular}{|c|c|}
\hline

% head of the table
$\gamma$        & $\gamma(t)$  \\ \hline

$ \forall x \alpha$ & $ \alpha(x/t)$ \\
$\neg \exists x \alpha$ & $\neg \alpha(x/t)$ \\ \hline
\end{tabular} \end{center}
\vspace{5mm}

\item{\em Existential type}: The constant introduced by these rules must not 
have occurred previously in the tableau. It is essentially a skolemization 
operation, and it is only applied once to each existential quantifier.
 
\vspace{5mm} 
\begin{center} \begin{tabular}{|c|c|}
\hline

% head of the table
$\delta$        & $\delta(c)$  \\ \hline

$\neg \forall x \alpha$ & $\neg \alpha(x/c)$ \\
$ \exists x \alpha$ & $\alpha(x/c)$ \\ \hline
\end{tabular} \end{center}
\vspace{5mm}

\end{itemize}

A branch of a tableau is {\em x-closed} if there is an atomic formula
$p(t_1,t_2,\ldots,t_n)$ such that both $p^y(t_1,t_2,\ldots,t_n)$ and
$\neg p^t(t_1,t_2,\ldots,t_n)$ are formulas on that branch, and $y,t
\leq x$. 

\begin{lemma} \label{forall-lemma}
A stratified valuation $\sigma$ x-satisfies $\alpha(v/t)$ for any $t$
in the universe $\cal D$ if it x-satisfies $\forall v \alpha$.
\end{lemma}

{\em Proof: } Let $\sigma$ be any valuation such that $\sigma
\xentails \forall v \alpha$. By definition~\ref{formula-definition}, 
$\sigma(v/u) \xentails \alpha$ for all $u$ in $\cal D$. We take $u =
t^\sigma$. That means, $\sigma(v/t^\sigma) \xentails \alpha$. Using
the substitution theorem~\ref{subst-theorem}, we obtain:
\begin{equation}
 \sigma(v/t^\sigma) \xentails \alpha \mbox{ iff } \sigma \xentails
\alpha(v/t) 
\end{equation}
Hence $\sigma \xentails \alpha(v/t)$.

\begin{lemma} \label{soundness-lemma}
Suppose $\sigma$ is a ${\cal SL}$ stratified valuation which x-satisfies
a set of stratified formulas $\Phi$. Let $T$ be a finite tableau for
$\Phi$. Then there is a branch $B$ of the tableau $T$ and an extension
$\sigma'$ of $\sigma$ to ${\cal SL'}$ such that $\sigma'$ x-satisfies
all formulas on $B$. Therefore, $B$ is open.
\end{lemma}

{\em Proof: } We use induction on $T$.

{\em Base case: } \\
If  $T$ is a single node, $\sigma'$ is any extension of $\sigma$.

{\em Induction step: } \\ 
The arguments for $\neg \neg$, $\rightarrow$, and $\neg \rightarrow$
are the same those that were given for the propositional case.

{\em Case 1 ($\forall$):} \\
Assume that $\sigma$ {\em x-satisfies} all formulas on a branch of
height $i$ that contains $\forall v \alpha$. Thus, in particular, it
{\em x-satisfies} $\forall v \alpha$. By lemma~\ref{forall-lemma},
$\sigma$ {\em x-satisfies} $\alpha(v/t)$ for any term $t$ in the
universe. Thus $\sigma' = \sigma$ already {\em x-satisfies} all
formulas on the new branch.

{\em Case 2 ($\neg \forall$):} \\
Assume that $\sigma$ {\em x-satisfies} all formulas on a branch of height
$i$ that contains $\neg \forall v \alpha$. Thus, in particular, it
{\em x-satisfies} $\neg \forall v \alpha$. Assume the branch is expanded 
according to the corresponding rule, such that it contains now
$\neg \alpha(v/c)$.  We must modify $\sigma$ to $\sigma'$ so that
$\sigma'$ {\em x-satisfies} all formulas on $B$, including the new one. 
Stratified valuation $\sigma$ {\em x-satisfies} $\neg \forall v \alpha$,
hence, by definition~\ref{formula-definition}, $\sigma(v/u) \xentails
\neg \alpha$ for some $u \in \cal D$. Let $\sigma' = \sigma(c/u)$. By
substitution theorem~\ref{subst-theorem}, $\sigma(v/u) \xentails \neg
\alpha \mbox{ iff } \sigma' \xentails \neg \alpha(v/c)$. Note that $c$
did not occur on $B$ earlier, so $\sigma'$ continues to satisfy all
formulas on branch $B$.

Cases $\exists$ and $\neg \exists$ can be treated in the same manner.

\begin{theorem}
{\em Soundness of stratified first-order tableaux.} If a set of
stratified first-order formulas $\Phi$ has a finite x-closed tableau,
then $\Phi$ is x-unsatisfiable. 
\end{theorem}

{\em Proof: } The result follows as a contrapositive of
lemma~\ref{soundness-lemma}. 

\vspace{5mm} 

To prove completeness we need a definition and some lemmas:

\begin{definition} \label{hintikka-set}
Let $\cal L$ be a language with at least one constant symbol and
$\Psi$ be a set of $\cal L$-sentences. We say that $\Psi$ is an
{\em extended Hintikka set with respect to level $x$}, provided for
all $\varphi \in \Psi$ 
\begin{enumerate}
\item if $\varphi$ is an atom of the form $\alpha^y$  then any
negation $\neg \alpha^t$ with $y,t \leq x$ is not in $\Psi$;
\item if $\varphi$ is a negated atomic formula of the form $\neg
\alpha^y$, then any atomic formula  $\alpha^t$ with $y,t \leq x$ is
not in $\Psi$; 
\item if $\varphi$ has the form $\neg \neg \alpha$ then $\alpha \in
\Psi$;
\item if $\varphi$ has the form $\alpha \rightarrow \beta$ then
either $\neg \alpha \in \Psi$ or $\beta \in \Psi$;
\item if $\varphi$ has the form $\neg(\alpha \rightarrow \beta)$ then
either $\alpha \in \Psi$ and $\neg \beta \in \Psi$;
\item if $\varphi$ has the form $\forall x \alpha$ then $\alpha(x/t)
\in \Psi$ for every ground term $t$;
\item if $\varphi$ has the form $\neg \forall x \alpha$ then $\neg
\alpha(x/t) \in \Psi$ for some ground terms $t$;
\item if $\varphi$ has the form $\exists x \alpha$ then $\alpha(x/t)
\in \Psi$ for some ground term $t$;
\item if $\varphi$ has the form $\neg \exists x \alpha$ then $\neg
\alpha(x/t) \in \Psi$ for all ground term $t$;
\end{enumerate}
\end{definition}

\begin{lemma}
Every extended Hintikka set with respect to level $x$ is
x-satisfiable. \label{hintikka-set-satisfaction} 
\end{lemma}

{\em Proof: } We construct an ${\cal SL}$ structure as follows: 
\begin{enumerate}
\item let $\cal D$ be the set of all ground terms in the extended Hintikka
set.  
\item let $(f(t_1,t_2,\ldots,t_n))^{\cal SL} = f^{\cal SL}(t_1^{\cal
SL},t_2^{\cal SL} \ldots t_n^{\cal SL})$ for each n-ary
function symbol $f$ in $\cal L$
\item $\langle t_1, \ldots, t_n \rangle \in {P^x}^{\cal SL} \mbox{ iff }
P^x(t_1, \ldots, t_n) \in \Psi$.
\item $\langle t_1, \ldots, t_n \rangle \in \overline{P^x}^{\cal SL} \mbox{
iff } \neg P^x(t_1, \ldots, t_n) \in \Psi$.  
\end{enumerate}

One can prove using induction on the degree of  $\varphi$ that any formula
$\varphi \in \Psi$ is {\em x-satisfiable} under ${\cal SL}$.

We extend now the notion of {\em used up} formula to the first-order
case. The definitions for {\em used up branch} and {\em used up
tableau} are similar with the ones found in the propositional case.

\begin{definition}
Let $\varphi$ be a first-order formula on a branch $B$ of a tableau $T$. We say
$\varphi$ is {\em used up} on $B$ iff one of the following holds:
\begin{enumerate}
\item $\varphi$ is $\neg \neg \alpha$ and $\alpha$ is on branch $B$;
\item $\varphi$ is $\neg(\alpha \rightarrow \beta)$ and $\alpha, \neg
\beta$ are on $B$;
\item $\varphi$ is $\alpha \rightarrow \beta$ and either $\alpha$ or
$\beta$ is on $B$;
\item $\varphi$ is $(\forall x)\alpha$ and $\alpha(x/t)$ is on $B$ for
all constant symbols $t$ occuring on formulas that belong to $B$.
\item $\varphi$ is $(\exists x)\alpha$ and $\alpha(x/t)$ is on $B$ for
some constant symbol $t$ occuring on formulas that belong to $B$.
\item $\varphi$ is $P^x$ or $\neg P^y$ for some atoms $P$, where $x,y \in
\{u,i,d\}$. 
\end{enumerate}
\end{definition}

\begin{lemma} 
The set of sentences of any used up x-open branch is an extended
Hintikka set with respect to level x. \label{hintikka-set-construction} 
\end{lemma}

\begin{lemma}
Every x-unsatisfiable countable or finite set of sentences $\Phi$ has a
possibly infinite tableau in which every branch is used up.
\label{tableau-construction}
\end{lemma}

{\em Proof: } The proofs of the last two lemmas follow exactly the
same steps as the ones for classical first-order logic found
in~\cite{bell86}. 

One can notice that the last three definitions do not prohibit the
existence of infinite branches. However, in our domain, we will use
function symbols mainly to formalize sentences. As it is shown in
chapter 6, it is counter intuitive to apply the universal
instantiation rule over sentences as well. Therefore, out tableaux
will be finite.  

\begin{theorem}
{\em Completeness of stratified first-order tableaux.} If $\Phi$ is an
x-unsatisfiable countable or finite set of sentences, then $\Phi$ has a
finite x-closed tableau.
\end{theorem}

{\em Proof: } The theorem is a consequence of
Lemma~\ref{hintikka-set-satisfaction},
\ref{hintikka-set-construction}, and \ref{tableau-construction}.

\section{Model ordering in first-order stratified logic}

We generalize the notions concerning model ordering found in
propositional stratified logic to first-order stratified logic.

\begin{definition}
A {\em model schema} for an exhausted stratified tableau is
represented by the set of all atomic formulas and negated atomic
formulas that belong to an open branch.
\end{definition}

\begin{definition}
An atomic formula $P(t_1, \ldots, t_n)$ is said to be {\em
affirmative} in a model schema $m$, if $P^x(t_1, \ldots, t_n)$ is a
member of $m$ and for any $P^y(t_1, \ldots, t_n)$ or $\neg P^y(t_1,
\ldots, t_n)$ occurring in $m$, $y > x$ $(u < i < d)$.  
\end{definition}

\begin{definition}
An atomic formula $P(t_1, \ldots, t_n)$ is said to be {\em
negative} in a model schema $m$, if $\neg P^x(t_1, \ldots, t_n)$ is a
member of $m$ and for any $P^y(t_1, \ldots, t_n)$ or $\neg P^y(t_1,
\ldots, t_n)$ occurring in $m$, $y > x$.  
\end{definition}

\begin{definition}
An atomic formula is {\em self-cancellable} in a model schema $m$, if
both $P^x(t_1, \ldots, t_n)$ and $\neg P^y(t_1, \ldots, t_n)$ occur in
$m$, where $x,y \in \{ u,i,d \}$. 
\end{definition}

\begin{definition}
A model schema $m_1$ is more {\em informative} than a model schema
$m_2$ iff for any positive or negative atomic formula $\varphi$ in
$m_2$, there is a positive or negative atomic formula $\varphi$ in $m_1$. 
\end{definition}

\begin{definition}
\label{weaker-formula}
An atomic formula $P^x(t_1, \ldots, t_n)$ is {\em weaker} than
$P^y(t_1, \ldots, t_n)$ iff $x$ and $y$ are members of the set
$\{u,i,d\}$ and $x \geq y$. 
\end{definition}

\begin{definition}
\label{weaker-model}
A model schema $m_1$ is {\em weaker} than a model schema $m_2$ if for
every non-self-cancellable atomic formula $\varphi^x$ in $m_2$ there is a
corresponding $\varphi^y$ formula in $m_1$ that is weaker and has the
same polarity.
\end{definition}

\begin{definition} \label{optimistic-models}
A model schema $m_1$ is more {\em optimistic} than a model schema
$m_2$ if it is more informative and weaker. We write this as $m_1 < m_2$.
\end{definition}

Consider a theory described in stratified first-order logic that
appropriately formalizes the classical default logic example
involving Tweety:
\[ 
\begin{array}{l}
bird(tweety) \\
(\forall x)(bird(x) \rightarrow flies^d(x)) \\
(\forall x)(penguin(x) \rightarrow \neg flies(x)) \\
\end{array}
\] 
If one applies the stratified tableau method, she will obtain two model
schemata that {\em u-satisfy} the theory:

\vspace{5mm} \noindent
{\small
\begin{center} \begin{tabular}{|l|l|l|l|} \hline \hline
{\em Schema \#} & {\em Indefeasible} & {\em Infelicitously defeasible} & {\em
Felicitously defeasible} \\ \hline 
$m_0$ & $bird(tweety)$      & &       \\
      & $\neg penguin(tweety)$  & &       \\
      &          & & $flies^d(tweety)$ \\ \hline
$m_1$ & $bird(tweety) $      & &       \\
      & $\neg flies(tweety)$ & & $flies^d(tweety)$ \\ \hline
\end{tabular} \end{center}
}
\vspace{5mm}

\noindent Model schema $m_0$ is more optimistic than model schema $m_1$. It
corresponds to an ${\cal SL}$ structure defined over a universe
${\cal D}$ that contains only one object, $tweety^{\cal D}$, and no
function symbols. The relations defined on the universe are
$\langle tweety^{\cal D} \rangle \in bird^u$, $\langle tweety^{\cal D}
\rangle \in \overline{penguin^u}$, and $\langle tweety^{\cal D}
\rangle \in flies^d$. 

For the sake of compactness and clarity we
represent model schemata as unions of relations partitioned according
to their strength: \[ m_0 = \{ bird(tweety),\neg penguin(tweety) \}
\cup \O^i \cup \{ flies^d(tweety) \} \]

It remains to be seen how this approach to nonmonotonic reasoning can
handle the sensitive problems found in the literature.
 	% stratified logic
\chapter{Infelicity, presupposition, and lexical pragmatics}

The lack of an appropriate framework capable of handling both
felicitously and infelicitously defeasible information was the main
reason that we could not formalize pragmatic inferences and signalling
infelicities (see chapter~\ref{relevant-research}).
Chapter~\ref{stratified-logic} was dedicated to the development of
such a framework. Here, we argue that a bare translation of natural
language utterances into stratified logic is still insufficient if one
wants to account for the subtleties that are inherent to pragmatic
inferences. The subtle feature that makes pragmatic inferences
different from the semantic ones is the fact that, as Grice
said~\shortcite{grice75}, they are carried by the saying of what is
said, rather than by what is said.  It is true that felicitously and
infelicitously defeasible inferences can be associated with certain
syntactic or lexical constructs, but they are triggered by {\em the
saying} of those constructs. We devote this chapter to an incremental
understanding of the notion of {\em lexical pragmatics}.

\section{Felicitously versus infelicitously defeasible inferences}

We have shown that the presuppositions carried by~\bsent{infel-pres},
{\em John regrets that Mary came to the party}, and~\bsent{fel-pres},
{\em John does not regret that Mary came to the party} do not have
similar properties: the presupposition carried by
sentence~\bsent{infel-pres} cannot be felicitously defeated, but that
related to sentence~\bsent{fel-pres} can be.  As we have already
stated, it is a logical mistake to formalize the pragmatic inference
in~\bsent{infel-pres} as a logical implication because the
contrapositive does not hold in all readings: {\em Mary did not come
to the party} does not imply that {\em It is not the case that John
regrets that Mary came to the party}.  We now have stratified logic to
capture these inferences. We delineate between felicitously and
infelicitously defeasible inferences by assigning them different
strengths. We still treat presuppositions as defeasible information,
but those in positive environments are formalized as infelicitously
defeasible inferences, while those in negative environments as
felicitously defeasible inferences. Formally, we write:

\begin{equation}
\begin{array}{l}
(\forall x,y,z)(regrets(x,\$come(y,z)) \rightarrow come^i(y,z)) \\
(\forall x,y,z)(\neg regrets(x, \$come(y,z)) \rightarrow come^d(y,z)) \\
\end{array}
\end{equation}
Obviously, in order to have well-formed formulas, proposition
$come(x,y)$ should be a function in the antecedent and a predicate in
the consequent of each implication. To avoid any ambiguity, we prefix
function symbols with the $\$$ sign.

Consider again utterance~\ref{fel-pres}. Its logical translation and
the requisite pragmatic knowledge follows:
\begin{equation}
\left\{
\begin{array}{l}
\neg regrets(john,\$come(mary,party)) \\
(\forall x,y,z)(regrets(x,\$come(y,z)) \rightarrow come^i(y,z)) \\
(\forall x,y,z)(\neg regrets(x, \$come(y,z)) \rightarrow come^d(y,z)) \\
\end{array}
\right.
\end{equation}
This theory yields the stratified semantic tableau given in
figure~\ref{regrets1}. 

\begin{figure}[htbp]
\centering
\label{regrets1}
\setlength{\unitlength}{0.012500in}%
\begingroup\makeatletter\ifx\SetFigFont\undefined
% extract first six characters in \fmtname
\def\x#1#2#3#4#5#6#7\relax{\def\x{#1#2#3#4#5#6}}%
\expandafter\x\fmtname xxxxxx\relax \def\y{splain}%
\ifx\x\y   % LaTeX or SliTeX?
\gdef\SetFigFont#1#2#3{%
  \ifnum #1<17\tiny\else \ifnum #1<20\small\else
  \ifnum #1<24\normalsize\else \ifnum #1<29\large\else
  \ifnum #1<34\Large\else \ifnum #1<41\LARGE\else
     \huge\fi\fi\fi\fi\fi\fi
  \csname #3\endcsname}%
\else
\gdef\SetFigFont#1#2#3{\begingroup
  \count@#1\relax \ifnum 25<\count@\count@25\fi
  \def\x{\endgroup\@setsize\SetFigFont{#2pt}}%
  \expandafter\x
    \csname \romannumeral\the\count@ pt\expandafter\endcsname
    \csname @\romannumeral\the\count@ pt\endcsname
  \csname #3\endcsname}%
\fi
\fi\endgroup
\begin{picture}(465,377)(2,450)
\thinlines
\put(220,730){\line( 0,-1){ 20}}
\put(220,640){\line(-4,-1){ 80}}
\put(220,640){\line( 5,-1){100}}
\put(320,580){\line(-2,-1){122}}
\put(320,580){\line( 4,-3){ 80}}
\put(220,810){\makebox(0,0)[b]{\smash{\SetFigFont{12}{14.4}{rm}{\small $\neg regrets(john,\$come(mary,party))$}}}}
\put(220,775){\makebox(0,0)[b]{\smash{\SetFigFont{12}{14.4}{rm}{\small $(\forall x,y,z)(\neg regrets(x,\$come(y,z)) \rightarrow come^d(y,z))$}}}}
\put(220,735){\makebox(0,0)[b]{\smash{\SetFigFont{12}{14.4}{rm}{\small $(\forall x,y,z)(regrets(x,\$come(y,z)) \rightarrow come^i(y,z))$}}}}
\put(220,690){\makebox(0,0)[b]{\smash{\SetFigFont{12}{14.4}{rm}{\small $\neg regrets(john,\$come(mary,party)) \rightarrow come^d(mary,party))$}}}}
\put(220,655){\makebox(0,0)[b]{\smash{\SetFigFont{12}{14.4}{rm}{\small $regrets(john,\$come(mary,party)) \rightarrow come^i(mary,party))$}}}}
\put(320,590){\makebox(0,0)[b]{\smash{\SetFigFont{12}{14.4}{rm}{\small $come^d(mary,party)$}}}}
\put(125,595){\makebox(0,0)[b]{\smash{\SetFigFont{12}{14.4}{rm}{\small $ regrets(john,\$come(mary,party))$}}}}
\put(200,495){\makebox(0,0)[b]{\smash{\SetFigFont{12}{14.4}{rm}{\small $\neg regrets(john,\$come(mary,party))$}}}}
\put(120,555){\makebox(0,0)[b]{\smash{\SetFigFont{12}{14.4}{rm}{\small {\em u-closed}}}}}
\put(200,450){\makebox(0,0)[b]{\smash{\SetFigFont{12}{14.4}{rm}{\small $m_0$}}}}
\put(400,495){\makebox(0,0)[b]{\smash{\SetFigFont{12}{14.4}{rm}{\small $come^i(mary,party)$}}}}
\put(395,455){\makebox(0,0)[b]{\smash{\SetFigFont{12}{14.4}{rm}{\small $m_1$}}}}
\end{picture}
\caption{Stratified tableau for {\em John does not regret that Mary
came to the party}} 
\end{figure}
The tableau yields  two model schemata; in both of them, it is
defeasibly inferred that {\em Mary came to the party}. The tableau in
figure~\ref{regrets1} is exhausted if one considers a sorted language
to have been used and no instantiations over sentences have been made. 

\vspace{5mm}
{\small \noindent
\begin{center} \begin{tabular}{|l|l|l|l|} \hline \hline
{\em Schema \#} & {\em Indefeasible} & {\em Infelicitously}  & {\em
Felicitously} \\ 
 & & \hspace*{2mm} {\em defeasible} & \hspace*{2mm} {\em defeasible} \\ \hline 
$m_0$ & $\neg regrets(john,\$come(mary,party)$      & &       \\
      &  &  & $come^d(mary,party)$ \\ \hline
$m_1$ & $\neg regrets(john,\$come(mary,party)$      & &       \\
      &  & $come^i(mary,party)$  & $come^d(mary,party)$ \\ \hline
\end{tabular} \end{center}
}
\vspace{5mm}

\noindent As expected, the model-ordering relation
$<$~(definition \ref{optimistic-models}) establishes $m_1$ as the
optimistic model for the theory because it contains as much
information as $m_0$ and an extra default inference has been applied
in it.

Reconsider now the utterance~\bsent{mary-yes-no}: {\em John does not
regret that Mary came to the party because she didn't come.}  An
appropriate formalization in stratified logic yields again two model
schemata:

\begin{equation}
\label{t1}
\left\{
\begin{array}{l}
\neg regret(john,\$come(mary,party)) \wedge \neg come(mary,party) \\
(\forall x,y,z)(regrets(x,\$come(y,z)) \rightarrow come^i(y,z)) \\
(\forall x,y,z)(\neg regrets(x, \$come(y,z)) \rightarrow come^d(y,z)) \\
\end{array}
\right.
\end{equation}

\vspace{5mm}
{\small \noindent
\begin{center} \begin{tabular}{|l|l|l|l|} \hline \hline
{\em Schema \#} & {\em Indefeasible} & {\em Infelicitously}  & {\em
Felicitously} \\ 
 & & \hspace*{2mm} {\em defeasible} & \hspace*{2mm} {\em defeasible}
\\ \hline 
$m_0$ & $\neg regrets(john,\$come(mary,party))$      & &       \\
      & $\neg come(mary,party)$   &  & $come^d(mary,party)$ \\ \hline
$m_1$ & $\neg regrets(john,\$come(mary,party))$      & &       \\
      & $\neg come(mary,party)$  & $come^i(mary,party)$  &
$come^d(mary,party)$ \\ \hline 
\end{tabular} \end{center}
}
\vspace{5mm}

\noindent For each of the two schemata, an ${\cal SL}$ structure can
be found that {\em u-satisfies} the initial theory. However there is a
fundamental difference between the two schemata: the first one is {\em
i-satisfiable} while the second one is not. For example
\begin{equation} \{ \neg regrets^i(john,\$come(mary,party)) \} \cup \{
\neg come^i(mary,party) \} \cup \O^d \ientails m_0 \end{equation} but there
is no ${\cal SL}$ structure such that
\begin{equation} {\cal SL} \ientails m_1. \end{equation}

Rational agents tend to notice if something goes wrong. A model schema
such as $m_1$ is a good example for this: any agent who notices that
an infelicitously defeasible inference is cancelled, as it happens in
$m_1$, will treat that model as infelicitous. We will make this idea
precise in definition~\ref{infelicity-def}.  This intuition gives us
a good reason to discard the infelicitous model schemata, if
possible. In this case, we are left with model $m_0$ where some
felicitously defeasible information is cancelled. This corresponds
entirely to our expectations: the initial utterance is felicitously
described by only one model in which a presupposition has been
cancelled.  Formally, $m_0$ is {\em u-satisfiable} but not {\em
d-satisfiable}. We make the following conjecture:

\begin{conjecture}
An utterance is infelicitous if there is no stratified structure that
i-satisfies its logical translation. 
\end{conjecture}
In other words, if all the model schemata that characterize an
utterance are not {\em i-satisfiable}, then that utterance is
infelicitous. Obviously, this is not the case for theory~\ref{t1}.
But reconsider utterance~\bsent{infel3}: {\em John regrets that Mary
came to the party but she didn't come.}  An appropriate formalization
in stratified logic yields only one model schema that is {\em
u-satisfiable}, but not {\em i-satisfiable}:
\begin{equation}
\label{t2}
\left\{
\begin{array}{l}
regret(john,\$come(mary,party)) \wedge \neg come(mary,party) \\
(\forall x,y,z)(regrets(x,\$come(y,z)) \rightarrow come^i(y,z)) \\
(\forall x,y,z)(\neg regrets(x, \$come(y,z)) \rightarrow come^d(y,z)) \\
\end{array}
\right.
\end{equation}

\vspace{5mm} \noindent
{\small
\begin{center} \begin{tabular}{|l|l|l|l|} \hline \hline
{\em Schema \#} & {\em Indefeasible} & {\em Infelicitously}  & {\em
Felicitously} \\ 
 & & \hspace*{2mm} {\em defeasible} & \hspace*{2mm} {\em defeasible}
\\ \hline 
$m_0$ & $regrets(john,\$come(mary,party))$      & &       \\
      & $\neg come(mary,party)$  & $come^i(mary,party)$  &
$come^d(mary,party)$  \\ \hline
\end{tabular} \end{center}
}
\vspace{5mm}

\noindent The theory we have obtained in stratified logic is still {\em
u-consistent}. For example, 
\begin{equation}
\{ regret(john,\$come(mary,party)), \neg come(mary,party) \}
\hspace{1mm} \cup \hspace{1mm} \O^i \hspace{1mm} \cup \hspace{1mm}
\O^d \entails \ref{t2}
\end{equation} 
 We can add new utterances and interpret them, but the
infelicity is signalled. 

If one analyzes the above examples, she will notice that
presuppositions can be associated with the felicitously defeasible
information that belongs to a stratified model schema:

\begin{conjecture}
The presuppositions and the implicatures  of an utterance are given by
the uncancelled defeasible information that belongs to an optimistic
model schema of a theory described in terms of stratified logic.  
\end{conjecture}

\section{From lexical semantics to lexical pragmatics}

\subsection{Equating lexical semantics to lexical pragmatics --- the
first attempt}

Consider the word {\em bachelor}. A semantic definition will specify
that a bachelor is an unmarried male adult. However, a statement such
as 
\bexample{My cousin is not a bachelor. \name{not-bac}}
is {\em usually} uttered with respect to objects that qualify as being 
bachelors, i.e.,  male adults. We say {\em usually} because is
acceptable to repair a misuse of the word {\em bachelor},
\bsent{not-bac1} and \bsent{not-bac2}, or to reinforce it, \bsent{not-bac3}:
\bexample{My cousin is not a bachelor, he is five years old. \name{not-bac1}}
\bexample{My cousin is not a bachelor, she is a female. \name{not-bac2}}
\bexample{My cousin is not a bachelor, he is married. \name{not-bac3}}
It seems that {\em bachelor} is a word associated with male adults,
and its salient property is being or not being married.  Being
unmarried is a salient property associated with bachelors, but not
being a bachelor does not automatically imply that the entity having
this property is married.  Examples~\bsent{not-bac1} and~\bsent{not-bac2} show
this. We use stratified logic to provide a tentative definition for
the semantics and pragmatics of the word {\em bachelor}.
\begin{equation}
B_S = 
\left\{
\begin{array}{l}
(\forall x)(bachelor(x) \rightarrow \neg married(x) \wedge male(x)
\wedge adult(x)) \\
(\forall x)(\neg bachelor(x) \rightarrow married^i(x)) \\
(\forall x)(\neg bachelor(x) \rightarrow adult^d(x)) \\
(\forall x)(\neg bachelor(x) \rightarrow male^d(x)) 
\end{array} 
\right.
\end{equation}
At the first sight, this definition seems correct. One may argue that
usually the term {\em bachelor} is used only for persons, so a correct
formalization would be: 
\begin{equation}
 (\forall x)((person(x) \wedge \neg bachelor(x)) \rightarrow
married^i(x)),
\end{equation}
but for the moment we keep it simple in order to emphasize our point.
If one knows that $x$ is a bachelor, she may infer that he is an
unmarried male adult. If $x$ is not a bachelor, it is very likely that
he is married --- more likely than being a child or a female. In other
words, being {\em married} is the salient feature of a non-bachelor;
to reflect this, we assign it a stronger level in stratified
logic. Let us see how this definition works. Reconsider
utterance~\bsent{not-bac}. A complete formalization of the utterance
and the relevant commonsense knowledge is given by theory:
\begin{equation}
B_S \hspace{1mm} \cup \hspace{1mm} \neg bachelor(cousin) 
\label{bachelor1sp}
\end{equation}
If one applies the stratified semantic tableau, she will obtain one
optimistic model schema for the theory:

\vspace{5mm} \noindent
{\small
\begin{center} \begin{tabular}{|l|l|l|l|} \hline \hline
{\em Schema \#} & {\em Indefeasible} & {\em Infelicitously defeasible} & {\em
Felicitously defeasible} \\ \hline 
$m_0$ & $\neg bachelor(cousin)$ &  &       \\
      &  & $married^i(cousin)$ &        \\
      &  &   & $male^d(cousin)$ \\ 
      &  &   & $adult^d(cousin)$ \\ \hline
\end{tabular} \end{center}
}
\vspace{5mm}

\subsection{Equating lexical semantics to lexical pragmatics --- the
last attempt}

The model we have obtained for theory~\ref{bachelor1sp} reflects one's
intuitions: my cousin is not a bachelor because most likely he is a
married male adult.  However, it is unreasonable to assume that a
real-world domain formalizes only facts about bachelors and
cousins. Assume that the initial world contains also the fact that
there is a linguist called $Chris$ in the environment. For the same
utterance, a complete formalization is given by theory:
\begin{equation}
B_S \hspace{1mm} \cup \hspace{1mm} \{ \neg bachelor(cousin),
linguist(Chris) \}
\label{bachelor2sp}
\end{equation} 
The theory provides two felicitous optimistic model schemata:

\vspace{5mm} \noindent
{\small
\begin{center} \begin{tabular}{|l|l|l|l|} \hline \hline
{\em Schema \#} & {\em Indefeasible} & {\em Infelicitously defeasible} & {\em
Felicitously defeasible} \\ \hline 
$m_0$ & $\neg bachelor(cousin)$ &  &  \\
      &  & $married^i(cousin)$ & \\
      &  &   & $male^d(cousin)$ \\
      &  &   & $adult^d(cousin)$ \\
      &  $linguist(Chris)$  &   &  \\
      &  $\neg bachelor(Chris)$ &  & \\
      &  & $married^i(Chris)$ & \\
      &  &   & $male^d(Chris)$ \\
      &  &   & $adult^d(Chris)$ \\  \hline
$m_1$ & $\neg bachelor(cousin)$ &  &  \\
      &  & $married^i(cousin)$ & \\
      &  &   & $male^d(cousin)$ \\
      &  &   & $adult^d(cousin)$ \\
      &  $linguist(Chris)$  &   &  \\
      &  $bachelor(Chris)$ &  & \\
      &  $\neg married(Chris)$ &  & \\
      & $male(Chris)$ &   & $male^d(Chris)$ \\
      & $adult(Chris)$ &   & $adult^d(Chris)$ \\  \hline
\end{tabular} \end{center}
}
\vspace{5mm}

\noindent Up to a point, the results seem correct. In both model
schemata, the cousin is not a bachelor and therefore, he may be a
married male adult, and Chris is a linguist. As expected there are two
ways one can extend the initial interpretation: in one of them Chris
is not a bachelor, in the other Chris is a bachelor. But obviously,
something goes wrong because both these extensions carry the extra
defeasible information that Chris is a male adult. In other words, the
defeasible information {\em Chris is a male adult} occurs in {\em all}
felicitous optimistic model schemata of the theory. Hence, one would
be inclined to assert that {\em Chris is a male adult} is a
presupposition for the initial utterance as well. This blatantly
contradicts our intuitions.  The problem we have just emphasized
pertains to what is called ``ungrounded inference'' in the literature
of nonmonotonic reasoning.  The solution we propose in the next
section takes advantage of the fact that presuppositions are triggered
by utterances. Therefore, the use of a presuppositional environment is
the element that determines a default conclusion to be inferred or
not.

\subsection{Lexical semantics versus lexical pragmatics --- drawing
the boundaries}

The problem with theory~\ref{bachelor2sp} is that the universal
instantiation rule (see page~\pageref{univ-quant-rule}) quantifies
over all the objects in a domain. So any object, independent or not of
the fact that it qualifies as being a bachelor, will trigger the
associated inferences.  To circumvent this problem one should pay
attention to the heart of the issue: defeasible inferences are
triggered by pragmatic maxims, so they should be fired only when a
statement containing the corresponding syntactic or lexical construct
is used in an utterance.  Therefore, it seems appropriate to formalize
pragmatic maxims as rules having the following form:
\begin{equation} 
\mbox{\underline{if} {\em uttered(something)}
\underline{then} {\em infer(something-else)}. }
\label{pragmatic-inference} 
\end{equation}
We solve the problem in three steps:

\begin{enumerate}
\item We delineate clearly what is uttered from what is not by
introducing a meta-logical construct, $uttered(x)$, that takes
one argument: the logical translation of an utterance. The metalogical
symbol $uttered$ is subject to the same logical rules as well-formed
formulas. For example, if $uttered(a \wedge b)$, then it is
appropriate to conclude $uttered(a)$ and $uttered(b)$. Hence, the
utterance of~\bsent{not-bac} will be formalized as: 
\begin{equation} uttered(\neg bachelor(cousin)) \end{equation}
\item We formalize pragmatic inferences as rules having the form given
in~\ref{pragmatic-inference}.
\item We allow the pragmatic rules to apply only to utterances. To
accomplish this, we introduce a new quantifier $\forallu$. The
semantics of $\forallu$ enhances definition~\ref{formula-definition}
as follows: 
\begin{definition} \label{pragmatic-satisfaction}
A stratified valuation $\sigma$ x-satisfies a formula $\alpha$ with
respect to an utterance $u = \alpha_1$ if and only if
\begin{itemize}
\item $\sigma \xentails \alpha$ where $\alpha$ is as the one specified in
definition~\ref{formula-definition}. 
\item $\sigma(\vec{v}/ \vec{t}) \xentails \alpha_1 \rightarrow
\alpha_2$ for all $\vec{t}$ such that $uttered(\alpha_1(\vec{t}))$ and
$\alpha = (\forallu \vec{v})(\alpha_1 \rightarrow \alpha_2)$ 
\end{itemize}
\end{definition}
\end{enumerate}

The second part of definition~\ref{pragmatic-satisfaction} says that a
pragmatic inference $(\forallu \vec{v})(\alpha_1 \rightarrow
\alpha_2)$ is instantiated only for those objects $\vec{t}$ that
belong to an utterance having the form $\alpha_1(\vec{t})$. Hence,
only if the antecedent of the pragmatic rule has been uttered can that
rule be applied.

Using the new quantifier, the semantic and pragmatic connotation 
for the word {\em bachelor} is: 
\begin{equation}
\label{bachelor}
B_{SP} = \left\{
\begin{array}{l}
(\forall x)(bachelor(x) \rightarrow \neg married(x) \wedge male(x)
\wedge adult(x)) \\
(\forallu x)(\neg bachelor(x) \rightarrow married^i(x)) \\
(\forallu x)(\neg bachelor(x) \rightarrow adult^d(x)) \\
(\forallu x)(\neg bachelor(x) \rightarrow male^d(x)) 
\end{array} 
\right.
\end{equation}

According to definition~\ref{pragmatic-satisfaction}, the rules
$(\forallu x)(\neg bachelor(x) \rightarrow is\_like^{i,d}(x))$ are
trigerred only if there is an object $someone$ in the universe of
discourse and an utterance containing $\neg bachelor(someone)$ has
been issued. The above definition provides now the expected results
even if the universe of discourse contains other
objects. Theory~\ref{Chris2} provides two optimistic model schemata
that correctly reflect one's intuitions. 

\begin{equation}
\label{Chris2}
B_{SP} \hspace{1mm} \cup \hspace{1mm} \{ linguist(Chris), uttered(\neg
bachelor(cousin)) \}
\end{equation}

\vspace{5mm} \noindent
{\small
\begin{center} \begin{tabular}{|l|l|l|l|} \hline \hline
{\em Schema \#} & {\em Indefeasible} & {\em Infelicitously defeasible} & {\em
Felicitously defeasible} \\ \hline 
$m_0$ & $\neg bachelor(cousin)$ &  &  \\
      &  & $married^i(cousin)$ & \\
      &  &   & $male^d(cousin)$ \\
      &  &   & $adult^d(cousin)$ \\
      &  $linguist(Chris)$  &   &  \\
      &  $\neg bachelor(Chris)$ &  & \\ \hline
$m_1$ & $\neg bachelor(cousin)$ &  &  \\
      &  & $married^i(cousin)$ & \\
      &  &   & $male^d(cousin)$ \\
      &  &   & $adult^d(cousin)$ \\
      &  $linguist(Chris)$  &   &  \\
      &  $\neg married(Chris)$ &  & \\
      & $male(Chris)$ &   &  \\
      & $adult(Chris)$ &   &  \\  \hline
\end{tabular} \end{center}
}
\vspace{5mm}

\noindent In the first schema, the cousin is not a bachelor, but
according to the pragmatic maxims one is inclined to believe that he
is a married male adult; Chris is a linguist and not a bachelor. In
the second model, the cousin is not a bachelor, but according to the
pragmatic maxims one is inclined to believe that he is a married male
adult; Chris is an unmarried male adult and he is a linguist too. Both
schemata contain the felicitously defeasible information that the
cousin is a male adult, and this corresponds to our intuition: they
are presuppositions of the utterance.

As seen, the use of a certain presuppositional environment --- in this
case the word {\em bachelor} --- warrants the application of the
appropriate default rule. One may argue that one would like to apply
the same default for inferences strongly support that a given object
is a bachelor, even though this property has not been explicitely
uttered. Defining the boundaries beyond which such a property should
apply is not a trivial issue. Therefore, we will ignore it for the
rest of this thesis.

The meta-logical symbol $uttered$ and the quantifier $\forallu$ are
syntactic sugar used to keep a theory as clean as possible. It means
that all properties concerning stratified logic are true for this
extension as well. To clarify this, note that we can formalize the
same problem by introducing a new predicate $uttered$ and treating a
given utterance as a conjunction between its logical translation and
its uttered units. Thus, the utterance of~\bsent{not-bac} would
be formalized as 
\begin{equation} 
\neg bachelor(cousin) \wedge uttered(not(bachelor(cousin))) 
\end{equation}
In the second conjunct, bachelor is a function. Instead of $\forallu$
one can write the pragmatic rules as
\begin{equation} (\forall x)(uttered(not(bachelor(x))) \rightarrow adult^d(x)) \end{equation}

We formalize now our intuitions:

\begin{definition} \label{infelicity-def}
 Let $\Phi$ be a theory described in terms of stratified first-order
logic that appropriately formalizes the semantics of lexical items
and the pragmatics of lexical and syntactic constructs. The
semantics of lexical terms is formalized using the quantifier
$\forall$, while the pragmatic maxims are captured using $\forallu$.
Let $uttered(u)$ be the logical translation of a given utterance or
set of utterances. We say that utterance $u$ is \ul{\em
infelicitous} if and only if there is no stratified structure ${\cal
SL}$ that {i-satisfies} $\Phi \cup uttered(u)$.
\end{definition}

\begin{definition} \label{presupposition-def}
Let $\Phi$ be a theory described in terms of stratified first-order
logic that appropriately formalizes the semantics of lexical items and
the pragmatics of lexical and syntactic constructs. The semantics of
lexical terms is formalized using the quantifier $\forall$, while the
pragmatic maxims are captured using $\forallu$.  Let $uttered(u)$ be
the logical translation of a given utterance or set of utterances. We
say that utterance $u$ \ul{\em presupposes} or \ul{\em
implicates} $p$ if and only if $p^d$ was derived using pragmatic
maxims in at least one felicitous optimistic model of the theory $\Phi
\cup uttered(u)$, and if $p^d$ is not cancelled by any stronger
information ($\neg p, \neg p^i, \neg p^d$) in any felicitous
optimistic model schema of the theory. Symmetrically, utterance $u$
\ul{\em presupposes} or \ul{\em implicates} $\neg p$ if
and only if $\neg p^d$ was derived using pragmatic maxims in at least
one felicitous optimistic model of the theory $\Phi \cup uttered(u)$,
and if $\neg p^d$ is not cancelled by any stronger information ($p,
p^i, p^d$) in any felicitous optimistic model schema of the theory. In
both cases, $\Phi \cup uttered(u)$ is u-consistent.
\end{definition}

\section{Lexical pragmatics at work}

\subsection{Presupposition cancellation}

Definitions~\ref{infelicity-def} and~\ref{presupposition-def} provide
the expected results for the examples studied so far. They explain why
{\em Mary came to the party} is a presupposition of the utterance {\em
John does not regret that Mary came to the party} and why {\em My
cousin is a male adult} is a presupposition of the utterance {\em My
cousin is not a bachelor}. They also explain why {\em John regrets
that Mary came to the party but she did not come} is an infelicitous
utterance.  We show now, how the same definitions explain the
presupposition cancellation phenomenon. Reconsider
utterance~\bsent{not-bac1}.  If one evaluates its logical translation
against a knowledge base that formalizes both the semantics and the
pragmatics of the word {\em bachelor} as given in~\ref{bachelor} , she will
obtain one felicitous optimistic model schema.
\begin{equation}
B_{SP} \hspace{1mm} \cup \hspace{1mm} uttered(\neg bachelor(cousin)
\wedge \neg adult(x)) 
\end{equation}

\vspace{5mm} \noindent
{\small
\begin{center} \begin{tabular}{|l|l|l|l|} \hline \hline
{\em Schema \#} & {\em Indefeasible} & {\em Infelicitously defeasible} & {\em
Felicitously defeasible} \\ \hline 
$m_0$ & $\neg bachelor(cousin)$ &  &       \\
      &  & $married^i(cousin)$ &        \\
      &  &   & $male^d(cousin)$ \\ 
      & $\neg adult(cousin)$  &   & $adult^d(cousin)$ \\ \hline
\end{tabular} \end{center}
}
\vspace{5mm}

As expected, the felicitously defeasible information,
$adult^d(cousin)$, is overwritten by some stronger information,
$\neg adult(cousin)$. Therefore, according to
definition~\ref{presupposition-def}, only {\em My cousin is a male}
will be assigned the status of presupposition for the utterance.
The same type of presupposition cancellation occurred
in~\bsent{mary-yes-no}.

\subsection{Implicatures, infelicities, and cancelability}

\subsubsection{Implicatures trigerred by the maxim of quality}

We have seen that by uttering $u$, a speaker conversationally
implicates (according to the maxim of quality) that she believes $u$
and has adequate evidence for it. If we consider only one agent, in a
second-order stratified model, this can be formalized as:
\begin{equation} 
(\forallu x)(uttered(x) \rightarrow believe(x)) 
\end{equation}
To stay within first-order logic, we instantiate the above formula for
our domain of interest. Assume that~\bsent{believe1} is uttered.
\bexample{The book is on the table. \name{believe1}} 
An appropriate translation that reflects the Gricean implicature
follows:
\begin{equation}
\left\{
\begin{array}{l}
uttered(on(book,table)) \\
(\forallu x,y)(on(x,y) \rightarrow believe^i(\$on(x,y))) \\
\end{array} 
\right.
\end{equation}
If one utters~\bsent{believe2}, theory~\ref{believe2eq} will have one
model schema that is infelicitous.
\bexample{The book is on the table, but I do not believe it. \name{believe2}}
\begin{equation}
\label{believe2eq}
\left\{
\begin{array}{l}
uttered(on(book,table) \wedge \neg believe(\$on(book,table))) \\
(\forallu x,y)(on(x,y) \rightarrow believe^i(\$on(x,y))) \\
\end{array} 
\right.
\end{equation}

\vspace{5mm} \noindent
{\small
\begin{center} \begin{tabular}{|l|l|l|l|} \hline \hline
{\em Schema \#} & {\em Indefeasible} & {\em Infelicitously defeasible} & {\em
Felicitously defeasible} \\ \hline 
$m_0$ & $on(book,table)$ &  &       \\
      & $\neg believe(\$on(book,table))$ & $believe^i(\$on(book,table))$
& \\ \hline
\end{tabular} \end{center}
}
\vspace{5mm}

\subsubsection{Scalar Implicatures}

Reconsider~\bsent{scalar-ex}, {\em John says that some of the boys
went to the theatre}, and its implicatures~\bsent{scalar-ex-implic},
{\em Not many/most/all of the boys went to the theatre}.  An
appropriate formalization is given in~\ref{scal-implic1}, where the
second formula captures the felicitously defeasible scalar
implicatures and the third formula reflects the relevant semantic
information for {\em all}.
\begin{equation}
\label{scal-implic1}
\left\{
\begin{array}{l}
uttered(went(\$some(boys), theatre)) \\
went(\$some(boys),theatre) \rightarrow (\neg
went^d(\$many(boys),theatre) \wedge \\
\hspace*{5mm} \neg went^d(\$most(boys),theatre) \wedge \neg
went^d(\$all(boys),theatre)) \\ 
went(\$all(boys),theatre) \rightarrow (went(\$most(boys),theatre) \wedge \\
\hspace*{5mm}  went(\$many(boys),theatre) \wedge went(\$some(boys),theatre)) \\
\end{array}
\right.
\end{equation}
The theory provides one felicitous optimistic model that reflects
the expected pragmatic inferences.

\vspace{5mm} \noindent
{\small
\begin{center} \begin{tabular}{|l|l|l|l|} \hline \hline
{\em Schema \#} & {\em Indefeasible} & {\em Infelicitously}  & {\em
Felicitously} \\ 
 & & \hspace*{2mm} {\em defeasible} & \hspace*{2mm} {\em defeasible}
\\ \hline 
$m_0$ 	& $went(\$some(boys),theatre)$ & &  \\
	&  & & $\neg went^d(\$most(boys),theatre)$ \\
	&  & & $\neg went^d(\$many(boys),theatre)$ \\
	& $\neg went(\$all(boys),theatre)$  & & $\neg
went^d(\$all(boys),theatre)$  \\ \hline
\end{tabular} \end{center}
}
\vspace{5mm}

Assume now, that after a moment of thought, the same person utters:
\bexample{ Some of the boys went to the theatre. In fact all of them
went to the theatre.} 
Adding the extra utterance to the initial theory~\ref{scal-implic1},
$uttered(went(\$all(boys), theatre))$, one would obtain one felicitous
optimistic model schema in which the conventional implicatures have
been cancelled:

\vspace{5mm} \noindent
{\small
\begin{center} \begin{tabular}{|l|l|l|l|} \hline \hline
{\em Schema \#} & {\em Indefeasible} & {\em Infelicitously}  & {\em
Felicitously} \\ 
 & & \hspace*{2mm} {\em defeasible} & \hspace*{2mm} {\em defeasible}
\\ \hline 
$m_0$ 	& $went(\$some(boys),theatre)$ & &  \\
	& $went(\$most(boys),theatre)$  & & $\neg
went^d(\$most(boys),theatre)$ \\ 
	& $went(\$many(boys),theatre)$ & & $\neg
went^d(\$many(boys),theatre)$ \\ 
	& $went(\$all(boys),theatre)$  & & $\neg
went^d(\$all(boys),theatre)$  \\ \hline
\end{tabular} \end{center}
}
\vspace{5mm}

\subsubsection{Natural language disjunction}

It has been argued that there is a strong inclination to interpret the
natural disjunction as an exclusive or. Here is our formal
explanation: If one considers the linguistic scale $\langle and, or
\rangle$, the intuition can be explained as being an instantiation of
the {\em scalar implicature} phenomenon.  One can formalize
utterance~\bsent{excl-or} and its scalar implicature as we have done
in~\ref{excl-or-eq}.
\bexample{Mary or Susan is in France. \name{excl-or}}
\begin{equation}
\label{excl-or-eq}
\left\{
\begin{array}{l}
uttered(is\_in(Mary,France) \vee is\_in(Susan,France)) \\
(is\_in(Mary,France) \vee is\_in(Susan,France)) \rightarrow \\
\hspace*{5mm} \neg (is\_in^d(Mary,France) \wedge is\_in^d(Susan,France)) \\
\end{array}
\right.
\end{equation}
The stratified semantic tableau provides two felicitous optimistic
models:

\vspace{5mm} \noindent
{\small
\begin{center} \begin{tabular}{|l|l|l|l|} \hline \hline
{\em Schema \#} & {\em Indefeasible} & {\em Infelicitously defeasible} & {\em
Felicitously defeasible} \\ \hline 
$m_0$ 	& $is\_in(Mary,France)$ & & \\
	& & & $\neg is\_in^d(Susan,France)$ \\ \hline
$m_1$ 	& $is\_in(Susan,France)$ & & \\
	& & & $\neg is\_in^d(Mary,France)$ \\ \hline
\end{tabular} \end{center}
}
\vspace{5mm}

\noindent The two schemata account for our inclination to treat the natural
disjunction as an exclusive or. Notice that they are obtained on the
basis of the Gricean Maxims. Scalar implicatures are felicitously
defeasible: there is only one way in which one can {\em u-satisfy}
utterance~\bsent{excl-or2}.
\bexample{ Mary will wash the car or do the dishes. Actually, she will
do both of them. \name{excl-or2}} 
This is done by building an ${\cal SL}$ structure such as:
\begin{equation}
\{will\_wash(mary,car),will\_do(mary,dishes) \} \hspace*{1mm} \cup
\hspace*{1mm} \O^i \hspace*{1mm} \cup  \hspace*{1mm} \O^d.
\end{equation}

  	% from lexical semantics to lexical pragmatics
\chapter{Solving the projection problem}

The Achille's heel for any pragmatic theory is its vulnerability to the
projection problem. A solution for the projection problem should
explain which of the pragmatic inferences are inherited by a structure
that uses the constructs that bear them, and why.
Gazdar~\shortcite{gazdar79} makes a clear delineation between the
projection of implicatures and presuppositions. Other
researchers~\cite{karttunen79,soames79,soames82} focus only on the
projection of presuppositions. We are interested in a unified
explanation for both phenomena. We believe that there is no
fundamental difference between presuppositions and implicatures: they
are both triggered by pragmatic rules and they are both felicitously
or infelicitously defeasible. Moreover, we believe that a solution for
the projection problem should not differ from a solution for
individual utterances: it would be unnatural to consider that
conversants apply different rules when they utter simple utterances
than when they utter complex ones.

The solution we propose here adds nothing to the solution developed so
far for simple utterances. As a matter of fact, the direct
presupposition cancellation phenomenon is treated as a projection
problem by many theories of presuppositions. We have already shown how
direct cancellation, such as the one found
in~\bsent{mary-yes-no},~\bsent{not-bac1}, or~\bsent{not-bac2}, is
handled by our formalism. We show now that the same
definition~\ref{presupposition-def} can explain how presuppositions
are inherited in more complex cases as well.

\section{Disjunctive utterances}

Consider the following utterances and some of their associated
presuppositions:   
\bexample{Chris is not a bachelor or he regrets that Mary came to
the party. \name{or1-not}} 
\bexample{Chris is a bachelor or a spinster. \name{or1-yes}}
\bexample{$\rhd$ Chris is a (male) adult. \name{or1-pres}}
\bexample{Either John is away or his wife is away. \name{or2-not}}
\bexample{Either John has no wife or his wife is away. \name{or2-yes}} 
\bexample{$\rhd$ (John has a wife). \name{or2-pres}}
As we have already seen, {\em Chris is not a bachelor} presupposes
that {\em Chris is a male adult}; {\em Chris regrets that Mary came to
the party} presupposes that {\em Mary came to the party}. There is no
contradiction between these two presuppositions, so one would expect a
conversant to infer both of them if she hears an utterance such
as~\bsent{or1-not}. However, when one examines
utterance~\bsent{or1-yes}, she observes immediately that there is a
contradiction between the presuppositions carried by the individual
components. Being a bachelor presupposes that {\em Chris is a male},
while being a spinster presupposes that {\em Chris is a
female}. Normally, we would expect a conversant to notice this
contradiction and to drop each of these elementary presuppositions
when she interprets \bsent{or1-yes}.  The same analysis can be done
for examples \bsent{or2-not} and \bsent{or2-yes}. The first utterance
inherits the elementary presupposition \bsent{or2-pres}, while the
second one does not.

\subsection{{\em Or} --- non-cancellation}

We now study in detail how stratified logic and  the model-ordering
relation capture one's intuitions.

An appropriate formalization for utterance~\bsent{or1-not} and the
necessary semantic and pragmatic knowledge is given in~\ref{or1-not-eq}.
\begin{equation}
\label{or1-not-eq}
\left\{
\begin{array}{l}
uttered(\neg bachelor(Chris) \vee regret(Chris, \$come(Mary,party))) \\
(\neg bachelor(Chris) \vee regret(Chris, \$come(Mary,party)))
\rightarrow  \\
\hspace*{5mm} \neg  (\neg bachelor^d(Chris) \wedge regret^d(Chris,
\$come(Mary,party))) \\
\neg male(Mary) \\
(\forall x)(bachelor(x) \rightarrow male(x) \wedge adult(x) \wedge
\neg married(x)) \\
(\forallu x) (\neg bachelor(x) \rightarrow married^i(x)) \\
(\forallu x) (\neg bachelor(x) \rightarrow adult^d(x)) \\
(\forallu x) (\neg bachelor(x) \rightarrow male^d(x)) \\
(\forallu x,y,z)(\neg regret(x, \$come(y,z)) \rightarrow come^d(y,z))
\\
(\forallu x,y,z)(regret(x, \$come(y,z)) \rightarrow come^i(y,z))
\\
\end{array}
\right.
\end{equation}
Besides the translation of the utterance, the initial theory contains
a formalization of the felicitously defeasible implicature that natural
disjunction is used as an exclusive {\em or}, the lexical semantics
for the word {\em bachelor}, and the lexical pragmatics for {\em
bachelor} and {\em regret}. 

\begin{figure} [hbt] 
\epsfxsize=8cm        % Define length of figure 
                      % With \epsfysize you can define width of figure
\leavevmode
\centering
\epsffile{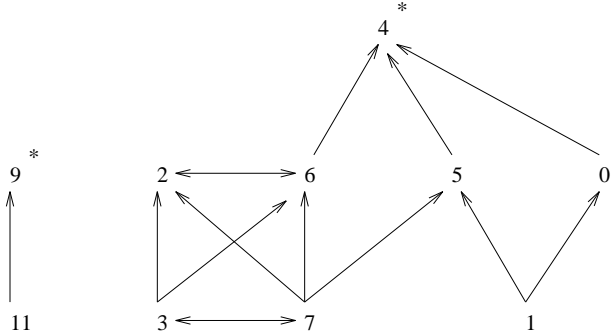}  % Works for both .ps or .eps
\caption{The felicitous models and their ordering for utterance {\em
Chris is not a bachelor or he regrets that Mary came to the party.}
The models labelled with an asterisk sign (4 and 9) are the most optimistic
ones.}      % Title of the figure (optional)
\label{or1-not-fig}
\end{figure}

The stratified semantic tableau generates 12 model schemata. Among
them, model $m_8$ and $m_{10}$ are infelicitous. The ordering relation
defined in~\ref{optimistic-models} generates over the space of model
schemata, the lattice given in figure~\ref{or1-not-fig}.  Here follow
the model schemata and their ordering.

\vspace{5mm} \noindent
{\small
\begin{center} \begin{tabular}{|l|l|l|l|} \hline \hline
{\em Schema \#} & {\em Indefeasible} & {\em Infelicitously defeasible} & {\em
Felicitously defeasible} \\ \hline 
$m_0$ & $regret(Chris,$ & & \\
      & \hspace*{3mm} $\$come(Mary,party))$      & &       \\
      & $\neg male(Mary)$  & & \\
      & $\neg bachelor(Mary)$ & & \\
      & $\neg married(Chris)$ & & \\
      & $adult(Chris)$ & & \\
      & $male(Chris)$ & & \\
      &  & $come^i(Mary,party)$ & $come^d(Mary,party)$ \\ \hline
$m_1$ & $regret(Chris,$ & & \\
      & \hspace*{3mm} $\$come(Mary,party))$      & &       \\
      & $\neg male(Mary)$  & & \\
      & $\neg bachelor(Mary)$ & & \\
      & $\neg married(Chris)$ & & \\
      & $adult(Chris)$ & & \\
      & $male(Chris)$ & & \\
      &  & $come^i(Mary,party)$ &  \\ \hline
$m_2$ & $regret(Chris,$ & & \\
      &  \hspace*{3mm} $\$come(Mary,party))$  & &       \\
      & $\neg male(Mary)$  & & \\
      & $\neg bachelor(Mary)$ & & \\
      & $\neg bachelor(Chris)$ & & \\
      &  & $come^i(Mary,party)$ & $come^d(Mary,party)$ \\ \hline
$m_3$ & $regret(Chris,$ & & \\ 
      &  \hspace*{3mm} $\$come(Mary,party))$  & &       \\
      & $\neg male(Mary)$  & & \\
      & $\neg bachelor(Mary)$ & & \\
      & $\neg bachelor(Chris)$ & & \\
      &  & $come^i(Mary,party)$ &  \\ \hline
$m_4$ & $regret(Chris,$ & & \\
      &  \hspace*{3mm} $ \$come(Mary,party))$  & &    \\ 
      & & & $bachelor^d(Chris)$ \\
      & $\neg male(Mary)$  & & \\
      & $\neg bachelor(Mary)$ & & \\
      & $\neg married(Chris)$ & & \\
      & $adult(Chris)$ & & \\
      & $male(Chris)$ & & \\
      &  & $come^i(Mary,party)$ & $come^d(Mary,party)$ \\ \hline
\end{tabular} \end{center}
}
\vspace{5mm}

\vspace{5mm} \noindent
{\small
\begin{center} \begin{tabular}{|l|l|l|l|} \hline \hline
{\em Schema \#} & {\em Indefeasible} & {\em Infelicitously defeasible} & {\em
Felicitously defeasible} \\ \hline
$m_5$ & $regret(Chris,$ & & \\
      &  \hspace*{3mm} $\$come(Mary,party))$  & &    \\ 
      & & & $bachelor^d(Chris)$ \\
      & $\neg male(Mary)$  & & \\
      & $\neg bachelor(Mary)$ & & \\
      & $\neg married(Chris)$ & & \\
      & $adult(Chris)$ & & \\
      & $male(Chris)$ & & \\
      &  & $come^i(Mary,party)$ &  \\ \hline
$m_6$ & $regret(Chris,$ & & \\
      & \hspace*{3mm} $\$come(Mary,party))$  & &    \\ 
      & $\neg male(Mary)$  & & \\
      & $\neg bachelor(Mary)$ & & \\
      & $\neg bachelor(Chris)$ & & $bachelor^d(Chris)$ \\
      &  & $come^i(Mary,party)$ & $come^d(Mary,party)$ \\ \hline
$m_7$ & $regret(Chris,$ & & \\
      & \hspace*{3mm} $\$come(Mary,party))$  & &    \\ 
      & & & $bachelor^d(Chris)$ \\
      & $\neg male(Mary)$  & & \\
      & $\neg bachelor(Mary)$ & & \\
      & $\neg bachelor(Chris)$ & & $bachelor^d(Chris)$ \\
      &  & $come^i(Mary,party)$ &  \\ \hline
$m_8$ & & & $\neg regret^d(Chris,$ \\
      & & & \hspace*{3mm} $\$come(Mary,party))$   \\ 
      & $\neg bachelor(Chris)$  & &  \\
      & $\neg male(Mary)$  & & \\
      & $\neg bachelor(Mary)$ & & \\
$*$   & $\neg married(Chris)$ & $married^i(Chris)$ & \\
      & $adult(Chris)$ & & $adult^d(Chris)$ \\
      & $male(Chris)$ & & $male^d(Chris)$ \\ \hline
$m_9$ & & & $\neg regret^d(Chris,$ \\
      & & & \hspace*{3mm} $\$come(Mary,party))$   \\ 
      & $\neg bachelor(Chris)$  & &  \\
      & $\neg male(Mary)$  & & \\
      & $\neg bachelor(Mary)$ & & \\
      & & $married^i(Chris)$ & \\
      & & & $adult^d(Chris)$ \\
      & & & $male^d(Chris)$ \\ \hline
\end{tabular} \end{center}
}
\vspace{5mm}

\vspace{5mm} \noindent
{\small
\begin{center} \begin{tabular}{|l|l|l|l|} \hline \hline
{\em Schema \#} & {\em Indefeasible} & {\em Infelicitously defeasible} & {\em
Felicitously defeasible} \\ \hline
$m_{10}$ & $\neg bachelor(Chris)$  & &  \\
      & $\neg male(Mary)$  & & \\
      & $\neg bachelor(Mary)$ & & \\
$*$   & $\neg married(Chris)$ & $married^i(Chris)$ & \\
      & $adult(Chris)$ & & $adult^d(Chris)$ \\
      & $male(Chris)$ & & $male^d(Chris)$ \\ \hline
$m_{11}$ & $\neg bachelor(Chris)$  & & $bachelor^d(Chris)$  \\
      & $\neg male(Mary)$  & & \\
      & $\neg bachelor(Mary)$ & & \\
      & & $married^i(Chris)$ & \\
      & & & $adult^d(Chris)$ \\
      & & & $male^d(Chris)$ \\ \hline
\end{tabular} \end{center}
}
\vspace{5mm}

The model-ordering relation gives $m_4$ and $m_9$ as optimistic models
for the utterance. The presupposition candidates for the utterance
are: $bachelor^d(Chris)$, $come^d(Mary,party)$, $\neg regret^d(Chris,
\$come(Mary,party))$, $adult^d(Chris)$, and $male^d(Chris)$. 
Note that the optimistic models contain $\neg bachelor(Chris)$ and
$regret(Chris, \$come(Mary,party))$, so according to
definition~\ref{presupposition-def} only $come^d(Mary,party)$,
$adult^d(Chris)$, and $male^d(Chris)$ will be projected as
presuppositions for the complex utterance~\bsent{or1-not}.  Therefore,
according to our theory, {\em Mary came to the party; Chris is a
male;} and {\em Chris is an adult} are labelled as presuppositions of
utterance~\bsent{or1-not}.

\subsection{{\em Or} --- cancellation}

Consider now utterance~\bsent{or1-yes}. An appropriate translation in
stratified logic and the relevant semantic and pragmatic knowledge are
given in theory~\ref{or1-yes-eq}.  Besides the translation of the utterance,
the initial theory contains a formalization of the felicitously
defeasible implicature that natural disjunction is used as an
exclusive {\em or} and the lexical semantics and pragmatics for the
word {\em bachelor} and {\em spinster}.

\begin{equation}
\label{or1-yes-eq}
\left\{
\begin{array}{l}
uttered(bachelor(Chris) \vee spinster(Chris)) \\
(bachelor(Chris)  \vee spinster(Chris) \rightarrow  \\
\hspace*{5mm} \neg  (bachelor^d(Chris) \wedge spinster^d(Chris)) \\
(\forall x)(bachelor(x) \rightarrow male(x) \wedge adult(x) \wedge
\neg married(x)) \\
(\forall x)(spinster(x) \rightarrow female(x) \wedge adult(x) \wedge
\neg married(x)) \\
(\forall x)(male(x) \leftrightarrow \neg female(x)) \\
(\forallu x) (\neg bachelor(x) \rightarrow married^i(x)) \\
(\forallu x) (\neg bachelor(x) \rightarrow adult^d(x)) \\
(\forallu x) (\neg bachelor(x) \rightarrow male^d(x)) \\
(\forallu x) (\neg spinster(x) \rightarrow married^i(x)) \\
(\forallu x) (\neg spinster(x) \rightarrow adult^d(x)) \\
(\forallu x) (\neg spinster(x) \rightarrow female^d(x)) \\
\end{array}
\right.
\end{equation}

The corresponding stratified semantic tableau yields 16 models. Among
them, four are infelicitous ($m_0, m_4, m_8$ and $m_{12}$).

\vspace{5mm} \noindent
{\small 
\begin{center} \begin{tabular}{|l|l|l|l|} \hline \hline
{\em Schema \#} & {\em Indefeasible} & {\em Infelicitously defeasible} & {\em
Felicitously defeasible} \\ \hline
$m_0$	& $spinster(Chris)$ &  &  $\neg spinster^d(Chris)$ \\
	& $\neg bachelor(Chris)$ & & \\
	& $adult(Chris)$ & & \\
	& $female(Chris)$ & & \\
$*$	& $\neg married(Chris)$ & $married^i(Chris)$ & \\
	& $\neg male(Chris)$ & & \\ \hline
$m_1$	& $spinster(Chris)$ &  &  $\neg spinster^d(Chris)$ \\
	& $\neg bachelor(Chris)$ & & \\
	& $adult(Chris)$ & & \\
	& $female(Chris)$ & & $female^d(Chris)$  \\
	& $\neg married(Chris)$ & & \\
	& $\neg male(Chris)$ & & \\ \hline
$m_2$	& $spinster(Chris)$ &  &  $\neg spinster^d(Chris)$ \\
	& $\neg bachelor(Chris)$ & & \\
	& $adult(Chris)$ & & $adult^d(Chris)$ \\
	& $female(Chris)$ & &   \\
	& $\neg married(Chris)$ &  & \\
	& $\neg male(Chris)$ & & \\ \hline
\end{tabular} \end{center}
}
\vspace{5mm}

\vspace{5mm} \noindent
{\small 
\begin{center} \begin{tabular}{|l|l|l|l|} \hline \hline
{\em Schema \#} & {\em Indefeasible} & {\em Infelicitously defeasible} & {\em
Felicitously defeasible} \\ \hline
$m_3$	& $spinster(Chris)$ &  &  $\neg spinster^d(Chris)$ \\
	& $\neg bachelor(Chris)$ & & \\
	& $adult(Chris)$ & &  \\
	& $female(Chris)$ & &   \\
	& $\neg married(Chris)$ &  & \\
	& $\neg male(Chris)$ & & \\ \hline
$m_4$	& $spinster(Chris)$ &  &  \\
	& $\neg bachelor(Chris)$ & & $\neg bachelor^d(Chris)$ \\
	& $adult(Chris)$ & &  \\
	& $female(Chris)$ & &   \\
$*$	& $\neg married(Chris)$ & $married^i(Chris)$ & \\
	& $\neg male(Chris)$ & & \\ \hline
$m_5$	& $spinster(Chris)$ &  &  \\
	& $\neg bachelor(Chris)$ & & $\neg bachelor^d(Chris)$ \\
	& $adult(Chris)$ & &  \\
	& $female(Chris)$ & & $female^d(Chris)$  \\
	& $\neg married(Chris)$ &  & \\
	& $\neg male(Chris)$ & & \\ \hline
$m_6$	& $spinster(Chris)$ &  &  \\
	& $\neg bachelor(Chris)$ & & $\neg bachelor^d(Chris)$ \\
	& $adult(Chris)$ & & $adult^d(Chris)$ \\
	& $female(Chris)$ & &   \\
	& $\neg married(Chris)$ &  & \\
	& $\neg male(Chris)$ & & \\ \hline
$m_7$	& $spinster(Chris)$ &  &  \\
	& $\neg bachelor(Chris)$ & & $\neg bachelor^d(Chris)$ \\
	& $adult(Chris)$ & &  \\
	& $female(Chris)$ & &   \\
	& $\neg married(Chris)$ &  & \\
	& $\neg male(Chris)$ & & \\ \hline
$m_8$	& $bachelor(Chris)$ & & \\
$*$	& $\neg married(Chris)$ & $married^i(Chris)$ & \\
	& $adult(Chris)$ & & \\
	& $male(Chris)$ & & \\
	& $\neg spinster(Chris)$ & & $\neg spinster^d(Chris)$ \\
	& $\neg female(Chris)$ & & \\ \hline
$m_9$	& $bachelor(Chris)$ & & \\
	& $\neg married(Chris)$ &  & \\
	& $adult(Chris)$ & & \\
	& $male(Chris)$ & & $male^d(Chris)$ \\
	& $\neg spinster(Chris)$ & & $\neg spinster^d(Chris)$ \\
	& $\neg female(Chris)$ & & \\ \hline
\end{tabular} \end{center}
}
\vspace{5mm}

\vspace{5mm} \noindent
{\small
\begin{center} \begin{tabular}{|l|l|l|l|} \hline \hline
{\em Schema \#} & {\em Indefeasible} & {\em Infelicitously defeasible} & {\em
Felicitously defeasible} \\ \hline
$m_{10}$& $bachelor(Chris)$ & & \\
	& $\neg married(Chris)$ &  & \\
	& $adult(Chris)$ & & $adult^d(Chris)$ \\
	& $male(Chris)$ & &  \\
	& $\neg spinster(Chris)$ & & $\neg spinster^d(Chris)$ \\
	& $\neg female(Chris)$ & & \\ \hline
$m_{11}$& $bachelor(Chris)$ & & \\
	& $\neg married(Chris)$ &  & \\
	& $adult(Chris)$ & &  \\
	& $male(Chris)$ & &  \\
	& $\neg spinster(Chris)$ & & $\neg spinster^d(Chris)$ \\
	& $\neg female(Chris)$ & & \\ \hline
$m_{12}$& $bachelor(Chris)$ & & $\neg bachelor^d(Chris)$ \\
$*$	& $\neg married(Chris)$ & $married^i(Chris)$  & \\
	& $adult(Chris)$ & &  \\
	& $male(Chris)$ & &  \\
	& $\neg spinster(Chris)$ & &  \\
	& $\neg female(Chris)$ & & \\ \hline
$m_{13}$& $bachelor(Chris)$ & & $\neg bachelor^d(Chris)$ \\
	& $\neg married(Chris)$ & & \\
	& $adult(Chris)$ & &  \\
	& $male(Chris)$ & & $male^d(Chris)$ \\
	& $\neg spinster(Chris)$ & &  \\
	& $\neg female(Chris)$ & & \\ \hline
$m_{14}$& $bachelor(Chris)$ & & $\neg bachelor^d(Chris)$ \\
	& $\neg married(Chris)$ &  & \\
	& $adult(Chris)$ & & $adult^d(Chris)$ \\
	& $male(Chris)$ & &  \\
	& $\neg spinster(Chris)$ & &  \\
	& $\neg female(Chris)$ & & \\ \hline
$m_{15}$& $bachelor(Chris)$ & & $\neg bachelor^d(Chris)$ \\
	& $\neg married(Chris)$ &  & \\
	& $adult(Chris)$ & & \\
	& $male(Chris)$ & &  \\
	& $\neg spinster(Chris)$ & &  \\
	& $\neg female(Chris)$ & & \\ \hline
\end{tabular} \end{center}
}
\vspace{5mm}

\newpage

\begin{figure} [hbt] 
\epsfxsize=8cm        % Define length of figure 
                      % With \epsfysize you can define width of figure
\leavevmode
\centering
\epsffile{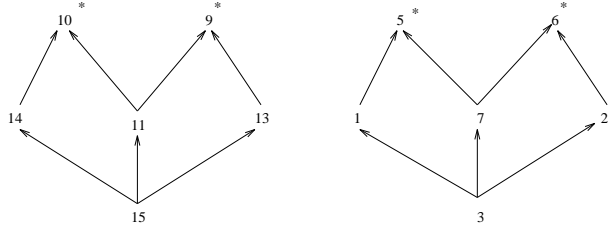}  % Works for both .ps or .eps
\caption{The felicitous models and their ordering for utterance {\em
Chris is a bachelor or a spinster.} The models labelled with an asterisk
sign (5, 6, 9, and 10) are the most optimistic ones.}  
\label{or1-yes-fig}
\end{figure}

The model schemata exhibit the ordering given in
figure~\ref{or1-yes-fig}. Among the presupposition candidates ($\neg
bachelor^d(Chris)$, $female^d(Chris)$, $adult^d(Chris)$,
$male^d(Chris)$, and $\neg spinster^d(Chris)$), only $adult^d(Chris)$
is not cancelled by some stronger information in the optimistic
felicitous models.  Therefore, {\em Chris is an adult} is the only
presupposition for the utterance.

\section{Conditional utterances}

\subsection{{\em If $\ldots$ then} --- non-cancellation}

An argument similar to that  given for disjunctive utterances can be made
for conditionals. Consider utterance~\bsent{if-not}. Its consequent
presupposes~\bsent{if-not-pres}.  The antecedent contains no
information that may cancel the presupposition. Hence, one's
expectations would be that {\em Sue came to the party} is a
presupposition of the whole conditional.  An appropriate formalization
in stratified logic is given in~\ref{if-not-eq}.
\bexample{If Mary came to the party then John will regret that Sue
came to the party. \name{if-not}}  
\bexample{ $\rhd$ Sue came to the party. \name{if-not-pres}}
\begin{equation}
\label{if-not-eq}
\left\{
\begin{array}{l}
uttered(come(Mary,party) \rightarrow regret(John, \$come(Sue,party)))
\\
(\forallu x,y,z)(regret(x,\$come(y,z)) \rightarrow come^i(y,z)) \\
(\forallu x,y,z)(\neg regret(x,\$come(y,z)) \rightarrow come^d(y,z))
\\
\end{array}
\right.
\end{equation}
Theory~\ref{if-not-eq} yields two felicitous optimistic model schemata. One
of them contains a presupposition candidate, $come^d(Sue,party)$,
that is not cancelled in the other model. Hence, {\em Sue came to
the party} is projected as a presupposition for the conditional.

\vspace{5mm} \noindent
{\small
\begin{center} \begin{tabular}{|l|l|l|l|} \hline \hline
{\em Schema \#} & {\em Indefeasible} & {\em Infelicitously defeasible} & {\em
Felicitously defeasible} \\ \hline
$m_0$ 	& $regret(John,\$come(Sue,party))$ & & \\
        & & $come^i(Sue,party)$ & $come^d(Sue,party)$ \\ \hline
$m_1$   & $\neg come(Mary,party)$ & & \\ \hline
\end{tabular} \end{center}
}
\vspace{5mm}

\subsection{{\em If $\ldots$ then} --- cancellation}

Consider utterance~\bsent{if-yes}. Its consequent
presupposes~\bsent{if-yes-pres}, but the antecedent is no longer
independent.  If the antecedent holds, then the presupposition is
overwritten; if it does not, the presupposition is
cancelled. Therefore, one would not expect in this case the
presupposition to be inherited by the whole conditional. An
appropriate formalization in stratified logic is given
in~\ref{if-yes-eq}.
\bexample{ If Mary came to the party then John will regret that she
did. \name{if-yes}}
\bexample{ ($\rhd$)  $\rhd \! \! \! \!/$ Mary came to the party.
\name{if-yes-pres}} 
\begin{equation}
\label{if-yes-eq}
\left\{
\begin{array}{l}
uttered(come(Mary,party) \rightarrow regret(John, \$come(Mary,party)))
\\
(\forallu x,y,z)(regret(x,\$come(y,z)) \rightarrow come^i(y,z)) \\
(\forallu x,y,z)(\neg regret(x,\$come(y,z)) \rightarrow come^d(y,z))
\\
\end{array}
\right.
\end{equation} 
The stratified tableau method determines two models. Model $m_0$ is
more informative than model $m_1$ but it is not weaker (see
definitions~\ref{weaker-formula} and~\ref{weaker-model}).  This is the
reason that model $m_0$ is {\em not} more optimistic than model $m_1$.
One of the felicitous optimistic models, $m_0$, contains a
presupposition candidate, $come^d(Mary,party)$, but the other one,
$m_1$, cancels it, $\neg come(Mary,party)$.  Therefore, the
presupposition is not projected.

\vspace{5mm} \noindent
{\small
\begin{center} \begin{tabular}{|l|l|l|l|} \hline \hline
{\em Schema \#} & {\em Indefeasible} & {\em Infelicitously defeasible} & {\em
Felicitously defeasible} \\ \hline
$m_0$ 	& $regret(John,\$come(Mary,party))$ & & \\
        & & $come^i(Mary,party)$ & $come^d(Mary,party)$ \\ \hline
$m_1$   & $\neg come(Mary,party)$ & & \\ \hline
\end{tabular} \end{center}
}
\vspace{5mm}

\section{Beyond complex utterances}

Most theories of presupposition over-simplify the projection
problem. They
consider~\cite{gazdar79,karttunen79,kay92,sandt92,soames82,wilson79,zeevat92}
that it is enough if a theory can successfully explain how
presuppositions are inherited in complex utterances such as
disjunctions or conditionals. No inquiry is made into the status of
presuppositions in sequence of utterances. As we have discussed in
section~\ref{gazdar}, the most puzzling consequence is that these
theories assign a dual life to presuppositions: in the initial stage,
as members of a simple or complex utterance, they are
defeasible. However, after that utterance is analyzed, there is no
possibility left of cancelling that presupposition.  But it is natural
to have presuppositions that are inferred and cancelled along a
sequence of utterances. Consider for example that Jane has two friends
--- John Smith and John McEnroe --- and that her roomate Mary met only
John Smith, a married fellow. Assume now that Jane has a conversation
with Mary in which she mentions only the name John because she is not
aware that Mary does not know that are two Johns. In this context, is
natural for Mary to get confused and to come to wrong conclusions. For
example, she may say that {\em John is not a bachelor}. She knows that
John Smith is a married male, so this makes sense for her. At this
point Jane realizes that Mary misunderstands her: all the time she was
talking about her cousin, John McEnroe. The uterance in~\bsent{utt1}
is a possible answer that Jane may give to Mary in order to clarify
the problem.
\bexample{ a. John is a not a bachelor. \\
\hspace*{12mm} b. I regret that you have misunderstood me. \\
\hspace*{12mm} c. He is only five years old. \\
\hspace*{12mm} d. You realize he cannot date women. \\ 
\hspace*{12mm} e. It is not he who is not a bachelor! \name{utt1}}
The first utterance in the sequence presupposes~\bsent{utt1-pres}.
\bexample{ $\rhd$ John is a male adult.
\name{utt1-pres}}  
Utterance~\bsent{utt1}b warns Mary that is very likely she misunderstood a
previous utterance~\bsent{utt2-pres}. The warning is conveyed by
implicature --- it is not entailed. 
\bexample{ $\rhd$ The hearer misunderstood the speaker. \name{utt2-pres}}
At this point, the hearer, Mary, starts to
believe that one of her previous utterances has been elaborated on a
false assumption, but she does not know which one. The third
utterance~\bsent{utt1}c comes to clarify the issue. It explicitly
expresses that John is not an adult. Therefore, it
cancels the early presupposition~\bsent{utt1-pres}.
\bexample{ $\rhd \! \! \! \!/$ John is an adult.
\name{utt3-pres}}  
Note that there is a gap of one statement between the generation and
the cancellation of this presupposition. In fact, it is more likely
that this presupposition has been inferred sometime earlier in the
dialog. The fourth utterance~\bsent{utt1}d emphasizes the
inappropriateness of presupposing that John is an adult. Again, the
information~\bsent{utt4-pres} is conveyed using pragmatic mechanisms,
as it was done in the second statement.
\bexample{ $\rhd$ John cannot date women.
\name{utt4-pres}} 
The last statement clarifies the issue. The speaker, Jane,
acknowledges that she sees now where the misunderstanding of the
hearer comes from. It may be the case that someone else, for example
John Smith, qualifies for not being a bachelor. This last piece of
information~\bsent{utt5-pres} is conveyed using a cleft construct.
\bexample{ $\rhd$ There is another person who is not a bachelor.
\name{utt5-pres}} 

Let us see how the above line of reasoning is reflected by our
computational method.  We consider that the initial knowledge base
contains the relevant information for interpreting the sequence of
utterances:
\begin{itemize}
\item the semantic information that bachelors are unmarried male
adults;
\item the pragmatic lexical information that those who are not
bachelors are likely to be married male adults;
\item first-order instances of the pragmatic lexical information that
factive verbs such as {\em regret} and {\em realize} presuppose their
complement; 
\item a first-order instance of the pragmatic syntactic information that
cleft constructs presuppose that there exists another agent who has
performed the corresponding action. More specifically, the last
formula in the theory says that if someone uttered that $x$ is not a
bachelor using a cleft construct, then the universe of discourse
should contain another object $y$ that has the property of being a
bachelor. The existence of this object is not entailed; therefore we
formalize it as defeasible, $E!^d$.
\end{itemize}
The existential presupposition is formalized using the predicate $E!$.
For a thorough analysis concerning existential presuppositions and
their formalization in stratified logic, the reader is referred to
chapter~\ref{existence}. 

The initial theory containing the translation of the first
utterance~\bsent{utt1}a and the appropriate knowledge is given
in theory~\ref{story1}. 

\begin{equation}
\label{story1}
\Phi_1 = 
\left\{
\begin{array}{l}
uttered(\neg bachelor(John)) \\
(\forall x)(bachelor(x) \rightarrow male(x) \wedge adult(x) \wedge
\neg married(x)) \\
(\forallu x) (\neg bachelor(x) \rightarrow married^i(x)) \\
(\forallu x) (\neg bachelor(x) \rightarrow adult^d(x)) \\
(\forallu x) (\neg bachelor(x) \rightarrow male^d(x)) \\
(\forallu x,y)(\neg regret(x, \$misunderstood(y,x)) \rightarrow
misunderstood^d(y,x)) \\
(\forallu x,y,z)(regret(x, \$misunderstood(y,x)) \rightarrow
misunderstood^i(y,z)) \\
(\forallu x,y)(\neg realize(x, \$cannot\_date\_women(y)) \rightarrow
\neg date\_women^d(y)) \\
(\forallu x,y)(realize(x, \$cannot\_date\_women(y)) \rightarrow
\neg date\_women^i(y)) \\
(\forallu x)(cleft (\$not\_bachelor(x)) \rightarrow (\exists y)(\neg
bachelor(y) \wedge E!^d(y))) \\
\end{array}
\right.
\end{equation}

The stratified semantic tableau yields one felicitous optimistic model
schema for the theory, in which one can presuppose that John is a male
adult. 

\vspace{5mm} \noindent
{\small 
\begin{center} \begin{tabular}{|l|l|l|l|} \hline \hline
{\em Schema \#} & {\em Indefeasible} & {\em Infelicitously defeasible} & {\em
Felicitously defeasible} \\ \hline
$m_0$ 	& $\neg bachelor(John)$ & & \\
        & & $married^i(John)$  &  \\ 
	& & & $adult^d(John)$ \\
	& & & $male^d(John)$ \\ \hline
\end{tabular} \end{center}
}
\vspace{5mm}

When sentence~\bsent{utt1}b is uttered, we have to examine a new theory that
is obtained by adding $uttered(regret(speaker,
\$misunderstood(hearer,speaker)))$ to theory~\ref{story1}. Therefore, the new
theory is: 
\begin{equation} 
\label{story2}
\Phi_2 =  \Phi_1 \cup  uttered(regret(speaker,
\$misunderstood(hearer,speaker))) 
\end{equation}
Theory $\Phi_2$ has 16 model schemata. Among them, eight are
felicitous. The ordering relation ($<$) leaves only four felicitous
optimistic models; none of these models cancels any pragmatic
information. Therefore, ~\bsent{utt1-pres} and ~\bsent{utt2-pres} are
projected as presuppositions of the first two statements.

Assume now that sentence~\bsent{utt1}c is uttered. The new theory,
$\Phi_3$,  has only eight model schemata, all of them
felicitous. Four model schemata are the most optimistic ones. Being an
adult is a presupposition candidate for $\Phi_3$ as well, but the
information is now cancelled in each optimistic model. Therefore, only
{\em John is a male} and {\em The hearer misunderstood the
speaker} are projected as presuppositions of the first three
sentences.
\begin{equation} 
\label{story3}
\Phi_3 =  \Phi_2 \cup  uttered(male(John) \wedge \neg adult(John)) 
\end{equation}

When the speaker utters sentence~\bsent{utt1}d, the new theory to be analyzed
is~\ref{story4}. 
\begin{equation} 
\label{story4}
\Phi_4 =  \Phi_3 \cup  uttered(realize(hearer,
\$cannot\_date\_women(John)))  
\end{equation}
Theory $\Phi_4$ has 16 model schemata, all felicitous. Four model schemata
are the most optimistic ones, and they presuppose that {\em 
John is a male}, {\em The hearer misunderstood the speaker}, and
{\em John does not date women}.

When the speaker utters the last sentence~\bsent{utt1}e, the theory to
be evaluated is: 
\begin{equation} 
\label{story5}
\Phi_5 =  \Phi_4 \cup  uttered(cleft(\neg bachelor(John))) 
\end{equation}
This has 24 felicitous model schemata and eight felicitous optimistic
model schemata. There are eight optimistic possibilities of constructing
stratified valuations for the sequence of utterances. They may differ
in the way the world is carved up: in one world the speaker can be a
bachelor, in the other he cannot. No restrictions have been
specified. However, independent of the way we choose to solve this
problem, in all eight optimistic models, the following presuppositions are
never cancelled: {\em John is a male}; {\em The hearer
misunderstood the speaker}; {\em John does not date
women}; and {\em There is someone else who is not a bachelor}.

The results obtained here offer a computational explanation for Geis's
work~\cite{geis82}. In his study of the language of TV advertising,
Geis noticed that most of the information that bombards one during the
commercial breaks is conveyed using pragmatic effects.  Implicatures
and presuppositions are the carriers of this information.  Pragmatic
inferences are defeasible, so according to Geis, a commercial that
pragmatically conveys some information to the audience does not commit
the originator of the commercial to the same extent as something
specifically uttered. One can observe that for any of the examples
discussed above, there is more than one interpretation.  For example,
utterance~\bsent{or1-yes} is satisfiable regardless of whether 
Chris is a bachelor or a spinster.  However, the pragmatic inference
that {\em Chris is an adult} is not contradicted in any optimistic
model. So even if it is defeasible information, there is a strong
tendency to believe that {\em Chris is an adult}. This is exactly the
reason that pragmatic inferences are psycholinguistically strong.

  	% solving the projection problem
\chapter{A Special Case of Pragmatic Inference: The Existential
Presupposition}
\label{existence}

It is common knowledge that a rational agent is inclined to presuppose
the existence of definite references that occur in utterances. Hearing
or uttering the examples below, a rational agent presupposes that the
cheese, children, and car physically exist.
\bexample{The cheese I bought yesterday is very bad.}
\bexample{I really don't know what to do with my children anymore.}
\bexample{Sorry I couldn't make it; my car broke on my way.}
However, day-to-day English provides an impressive number of cases
when existential presuppositions are not inferred, or when they are
defeated by some common sense knowledge (see~\cite{hirst91}
for a comprehensive study).  One can explicitly speak of
nonexistence~\bsent{nonexistence}; events and actions that do not
occur~\bsent{not occur}; existence at other times~\bsent{other times};
or fictional and imaginary objects~\bsent{fictions}.
\bexample{No one got an A+ in this course. \name{nonexistence}}
\bexample{John's party is cancelled. \name{not occur}}
\bexample{G\"odel was a brilliant mathematician. \name{other times}}
\bexample{Sherlock Holmes is smarter than any other detective.
\name{fictions}} 
Note that the simple dichotomy found in most approaches to
presupposition between existent and nonexistent objects is not enough
for a full account of natural language expressiveness. 

While trying to explain the whole phenomenon and to provide solutions
for the projection problem, linguists have often omitted any
explanation for the existential commitment of definite references or
their explanation has been a superficial one.  Similarly, philosophers
who have studied existence and nonexistence have been more concerned
with providing formal tools for manipulation of nonexistent objects
than tools to capture our common sense commitment.  This puts us in a
difficult position. From a linguistic perspective, the literature
provides a good set of theories able to more or less explain the
commitment to the presupposed truth of factives and the like but not
the existential commitment of definite references.  From a
philosophical perspective, we have quite a few theories which deal
with existence and nonexistence, but they too offer no explanation for
existential commitment. 

We are interested in a theory that is able to handle all pragmatic
phenomena in a unified manner. Therefore, we expect a solution for the
existential presuppositions to be similar with a solution for factives
and the like.  This chapter reviews some of the most significant works
that deal with nonexistence: we study their ability to reflect the
presupposition that definite referents exist, and we criticize
possible extensions of the formalisms that we study for reflecting the
types of pragmatic inferences that we have presented in
chapter~\ref{relevant-research}.

\section{On Frege-Russell-Quine nonexistence }

Frege~\shortcite{frege92} makes a sharp distinction between sense and
reference --- {\em nominatum}. According to Frege, constructions such
as {\em the heavenly body which has the greatest distance from the
earth} or {\em the series with the least
convergence}~\cite[p.~87]{frege92} certainly have some sense, but it
is doubtful they have a {\em nominatum}.
\begin{quote}
Therefore, the grasping of a particular sense does not with
certainty warrant a corresponding nominatum~\cite[p.~87]{frege92}.
\end{quote}
However, from a logical perspective, Frege rejects the words which
do not refer: {\em He who does not acknowledge the nominatum cannot
ascribe or deny a predicate to it}.  For Frege, whenever something is
asserted, the presupposition taken {\em for granted} is that the
employed proper names, simple or compound, have
nominata~\cite[p.~95]{frege92}. Therefore, it is impossible to account
for the cancellation of the pragmatic inferences. Moreover, sentences
such as~\bsent{nonexistence}, \bsent{not occur}, \bsent{other times},
or \bsent{fictions} are impossible to represent.

Russell was aware of the problems which occur when the denotation is
absent in a universe where denoting phrases {\em express} a meaning
and {\em denote} a denotation.  We follow here
Hirst's~\shortcite{hirst91} line of reasoning in presenting this
mistake.

In a Russellian ontological space, an appropriate translation of 
sentence~\bsent{dragons exist} is formula~\ref{dragons exist-eq}. 
\bexample{Dragons exist. \name{dragons exist}} 
\begin{equation}
	(\exists x)( dragon(x)) \label{dragons exist-eq}
\end{equation}
According to Russell, sentence~\bsent{dragons exist} is false because
there are no dragons: 
\begin{equation}
	(\forall x)(\neg dragon(x)) \label{dragons do not exist-eq}
\end{equation}
If someone is committed to the truth of formula~\ref{dragons do not
exist-eq}, then she is obviously committed to the falsity
of formula~\ref{dragons exist-eq}.

Let us now try to assign properties to nonexistent objects. A sentence
such as~\bsent{dragons like children} is vacuously true if there are
no dragons (formula~\ref{dragons like children-eq}), but a sentence
about a specific dragon~\bsent{my dragon likes cake} is false
(formula~\ref{my dragon likes cake-eq}).
\bexample{Dragons like children. \name{dragons like children}}
\begin{equation}
	(\forall x)(dragon(x) \rightarrow likes\_children(x))
\label{dragons like children-eq}
\end{equation}
\bexample{My dragon likes cake. \name{my dragon likes cake}}
\begin{equation}
	(\exists x)(my\_dragon(x) \wedge likes\_cake(x)) \label{my
dragon likes cake-eq}
\end{equation}
Therefore, any statement that contains references to fictional
characters is false. Even though our common sense assigns the value
{\em true} to a sentence such as~\bsent{sherlock holmes is smart},
according to the above formalism, the logical
translation~\ref{sherlock holmes is smart-eq} is {\em false}. 
\bexample{Sherlock Holmes is smart. \name{sherlock holmes is smart}}
\begin{equation}
	(\exists x)(is\_Holmes(x) \wedge smart(x)) \label{sherlock
holmes is smart-eq}
\end{equation}
If we take now the negation of~\bsent{my dragon likes cake}, i.e., 
\bsent{my dragon does not like cake}, we can see that negation does
not change the truth value of the corresponding translation. The appropriate
logical form~\ref{my dragon does not like cake-eq} remains {\em false}
because we are still committed to the nonexistence of dragons.
\bexample{My dragon does not like cake. \name{my dragon does not like cake}}
\begin{equation}
	(\exists x)(my\_dragon(x) \wedge \neg likes\_cake(x))
\label{my dragon does not like cake-eq}
\end{equation}

To circumvent these problems, Russell~\shortcite[p.~111]{russell05}
renounces taking presuppositions {\em for granted}. Instead,  he explains
the commitment to the {\em existence} of definite referents by the use
of paraphrase. Thus~\bsent{russell-ex} is paraphrased
as~\bsent{russell-ex-para}. 
\bexample{The author of Waverley was a man. \name{russell-ex}}
\bexample{One and only one entity is the author of Waverley and that
one is a man. \name{russell-ex-para}} 
\begin{equation}
(\exists x)(author\_of\_waverley(x) \wedge (\forall
y)(author\_of\_waverley(y) \rightarrow y = x) \wedge man(x))
\end{equation}
If we paraphrase a sentence such as~\bsent{buganda-ex}, {\em The King
of Buganda is (not) bald}, we will have to choose between a
representation that contains an internal negation and one that
contains an external one according to whether we know or we do
not know that there exists a king of Buganda.  Unfortunately, there is
no criterion for determining the right translation. Quine
\shortcite[p.~27]{quine49} shows how Russell's method can be extended to
include proper names, but this is done in a framework where
\begin{quote}
To say that {\em something} does not exist, or that is something which
{\em is not} is clearly a contradiction in terms; hence $(x)(x
\hspace{1mm}.exists)$  must be true~\cite[p.~150]{quine47}.
\end{quote}
Hence, Quine's solution cannot accommodate the cancellation of
presuppositions.  

Strawson's~\shortcite{strawson50} attack in the early fifties was only
the beginning of the denial of the Russellian approach to
presuppositions. Today, the paraphrase theory is no longer widely
accepted.

\section{Meinongian approaches}

Meinong's~\shortcite{meinong04} philosophical enterprise aimed to be a
theory of nonexistent objects based on mental acts. For Meinong, every
mental act is directed towards an object, but it is not necessary that
the objects of thought exist. The mental acts include knowing,
believing, judging, presupposing, thinking, etc. In a Meinong
universe, the existence of objects is no longer a tautology, as it was
for Quine: one can affirm or can deny the existence of the objects,
and more than that, one is free to assign properties to them whether
or not the objects exist. A conversational perspective of the
philosophical principles that underlie Meinong's work will be
discussed in section~\ref{meinong principles}. These principles
constitute the foundation for several logical formalisms. We review
Parsons's and Hirst's work and we analyze their suitability to reflect
the existential commitment that characterizes the use of definite
referents.

\subsection{Parsons's Theory}

Parsons's ontology consists of objects built on two different kinds of
properties and one mode of predication. {\em Nuclear} properties
include the ordinary ones: being blue, being a mountain, or being
kicked by Socrates. {\em Extranuclear} properties include existence,
being possible, being thought about by $X$.  Informally speaking, the
objects are constructed on infinite sets of properties, and they enjoy
the following features:
\begin{quote}
	1. No two objects (real or unreal) have exactly the same
nuclear properties.

	2. For any set of nuclear properties, some objects have all the
properties in that set and no other nuclear properties.
\end{quote}
Nonexistent objects are correlated with the set of properties which
does not correspond to an existing object.

%	Our main concern here is how Parsons' theory can capture the
%existential commitment. From the beginning, Parsons wants to convince
%us that {\em the restricted satisfaction principle} --- {\em any
%definite description refers to an object that satisfies the
%description} --- is not appropriate~\cite[p. 31]{parsons80}.  The only
%problem is that this principle is essential from a natural language
%perspective: it is the one that captures our commitment
%to the existence of definite references. If we put it aside, we will
%have no possibility to reflect our commitment.  Making the English
%definite description refer to an object that has internally some
%properties, including existence for example leaves us with no formal
%tool for capturing our commitment to an object that is really
%existent.

	Parsons avoids Russell's paraphrase of the definite
description by using the symbol $\iota$. For Parsons, $(\iota x) \Phi$
refers to the unique object, that satisfies $\Phi$ if there is such an
object. Otherwise, it does not refer to anything at all. For a sentence
such as~\bsent{parsons-ex1}, Parsons argues~\shortcite[p.~114]{parsons80}
that translation~\ref{parsons-ex1-eq1} will not do it, because this
translation is not committed to the existence of the man in the
doorway. So he proposes that the translation should
be~\ref{parsons-ex1-eq2}, where $E!$ stands for the existential
predicate.  
\bexample{The man in the doorway is clever. \name{parsons-ex1}}
\begin{equation}
(\iota x)((man(x) \wedge in\_the\_doorway(x)) \wedge clever(x))
\label{parsons-ex1-eq1} 
\end{equation}
\begin{equation}
(\iota x)((E!(x) \wedge man(x) \wedge in\_the\_doorway(x)) \wedge
clever(x))   \label{parsons-ex1-eq2}
\end{equation}
A proposition $(\iota x)\Phi\alpha$ is true, if $(\iota x)\Phi$ refers
to an object, and the object referred to has the property that
$\alpha$ stands for. If $(\iota x)\Phi$ fails to refer, then $(\iota
x)\Phi\alpha$ is automatically false. The problem with this is that it
embeds the existential commitment in the logical translation not as
something which is {\em implied} or presupposed, but as something {\em
said} or specified.  This is not the case with linguistic
presuppositions. Therefore, the first translation is too weak, i.e.,
unable to capture our commitment, and the second one too strong, i.e.,
the commitment becomes part of the translation.

The direct embedding of the existential presupposition in the logical
form reinforces the natural language negation ambiguity in sentences
which deal with nonexistent entities. Hence, sentence~\bsent{king does
not exist}, in the case there is no king of Buganda, can be both true
and false depending on the translation we choose:
\bexample{The king of Buganda does not exist. \name{king does not exist}}
\begin{equation}
\neg(\iota x)(E!(x) \wedge king\_of\_buganda(x))[\lambda x_{n} E!
(x_{n})] 
\end{equation}
\begin{equation}
(\iota x)(E!(x) \wedge king\_of\_buganda(x))[\lambda x_{n} \neg E!
(x_{n})] 
\end{equation}

\subsection{Hirst's Theory}

Hirst~\shortcite{hirst91} presents an
extensive study of the abilities of knowledge representation formalisms to
capture nonexistent entities, as they appear in natural
language.  After a short review of the main philosophical trends, he
shows that Russell-Quine's ontology, intensional approaches, free
logics, and possible-world formalisms are not appropriate for the study
of nonexistence.  According to Hirst, a natural language ontology
should reflect at least eight types of existence: physical existence
in the present real world, physical existence in a past world,
abstract necessary existence, existence outside a world but with
causal interaction with that world, abstract contingent existence in
the real world, existence as a concept, unactualized existence and
existence in fiction. On the basis of this rich ontology he provides a
generalization for Parsons's idea of nuclearity, and he formalizes it
using Hobbs's transparent predicates~\cite{hobbs85}. Hirst is not
interested in explaining the commitment to physical existence, so the
ontology does not provide any preference for the inclusion of a given
object in one category or another.

\section{Other approaches}

\subsection{Lejewski's unrestricted interpretation}

We were able to track  attacks on  Quine's slogan, {\em everything
exists}, back to early works of Lejewski and Hintikka.

Lejewski~\shortcite{lejewski54} builds his arguments starting from two
puzzling examples which seem to disprove the validity of the rules of
universal instantiation and existential generalization applied to
reasoning with empty noun expressions. In a universe \`{a} la Quine,
using the universal instantiation rule and the existential predicate
$E!$, from $(\forall x)(E!(x))$ we infer {\em Pegasus exists}. From the
sentence {\em Pegasus does not exist}, we infer by existential
generalization that $(\exists x)(\neg E!(x))$. However, these inferences
are objectionable to our intuitions.  In other words the logical laws:
\begin{equation}
(\forall x)(F(x)) \rightarrow F(y)	\hspace{20mm} \mbox{ Universal
Instantiation --- UI}
\end{equation}
and
\begin{equation}
F(y) \rightarrow (\exists x)F(x)	\hspace{25mm} \mbox{ Existential
Generalization --- EG}
\end{equation} 
do not hold for every interpretation of $F$ and every substitution
for the free variable.

	Seeking solutions to this problem, Lejewski turns  to the
interpretations of the quantifiers. Let us consider first a universe
with two objects, $a$ and $b$. In this universe, following Quine, we
can express the existential quantification $(\exists x)(F(x))$ as logical
disjunction $F(a) \vee F(b)$, and the universal quantification as logical
conjunction $F(a) \wedge F(b)$. In this fixed universe, introducing
noun-expressions can be regarded as a linguistic matter in the sense that
it does not affect our assumed universe, which continues to consist
only of
$a$ and $b$. However, we can use these objects. For example we can
say that $d$ satisfies the predicate $D$, or that the object $c$ does
not exist. In this universe, $(\forall x)(E!(x))$  no longer
implies that {\em $c$ exists}. The universal instantiation and
existential generalization are valid rules but are restricted to
reasoning only with $a$ and $b$. But this is not the only possible
interpretation of a quantifier. We can read them differently, such as:

$F(a) \vee F(b) \vee F(c) \vee F(d)$ for $(\exists x)(F(x))$ and

$F(a) \wedge F(b) \wedge F(c) \wedge F(d)$ for $(\forall x)(F(x))$.

	Thus $(\exists x)(\neg E!(x))$ is true because
$c$ does not exist and $(\forall x)(E!(x))$ is false. Under such an
interpretation the rules of Universal Instantiation and Existential
Generalization are valid without restrictions. Actually, Lejewski
emphasizes a dichotomy between two possible ways of interpreting the
quantifiers: a {\em restricted interpretation} under which every
component of an expression contains a noun-expression which has a {\em
nominatum}, and an {\em unrestricted interpretation} under which we
can only say that the objects are meaningful noun-expressions. This
gives two different theories of quantification.

One way  to understand expressions is in  light of the restricted
interpretation:
\begin{eqnarray}
(\forall x)(F(x)) \rightarrow F(y) \\
F(y) \rightarrow (\exists x)(F(x)) \\
(\forall x)(F(x)) \leftrightarrow \neg (\exists x)(\neg F(x))
\end{eqnarray}
In  light of the unrestricted interpretation, we can express the
same thing as:
\begin{eqnarray}
(\forall x)(E!(x)  \rightarrow F(x)) \rightarrow F(y) \\
F(y) \rightarrow (\exists x)(E!(x) \wedge F(x)) \\
(\forall x)(E!(x) \rightarrow F(x)) \leftrightarrow \neg (\exists
x)(E!(x) \wedge \neg F(x))
\end{eqnarray}

It is obvious now that under the {\em unrestricted interpretation},
the two inferences about Pegasus cannot be used as counterexamples to
disprove the validity of the rules of universal instantiation and
existential generalization.

This theory has a wonderful advantage. It no longer merges the idea of
quantification with the notion of existence, but from a
presuppositional point of view, it does not provide an explanation for
the existential commitment. A translation of a sentence such
as~\bsent{buganda-ex}, {\em The king of Buganda is bald}, under the
unrestricted interpretation is neutral with respect to the
king's existence~\ref{buganda-ex-eq}.
\begin{equation}
(\exists x)(King\_of\_Buganda(x) \wedge bald(x)) \label{buganda-ex-eq}
\end{equation}
Lejewski's universe is flat, it does not reflect our commitment to existence,
and this is not what we would like to have.

\subsection{Hintikka's existential presuppositions}

Another detractor of Russell-Quine's theory is
Hintikka~\shortcite{hintikka70,hintikka59}:
\begin{quote}
Of course, we are often likely to drop a name altogether if it turns
out to be empty. But this is because there is usually very little to
be said about what there is not, and not because an empty name is a
logical misnomer~\cite[p.~127]{hintikka59}.
\end{quote}
Hintikka's belief is that the meaningfulness and the logical status of
names and of other singular terms is not affected by the failure to
refer. Hintikka states clearly that there are existential
presuppositions embodied in the usual systems of quantification
theory~\cite[p.~130]{hintikka59}. To avoid this embedding, he
explicitly introduces Quine's slogan, {\em to exist is to be a value
of a variable}, in the Existential Generalization rule. Thus, he
replaces
\begin{equation}
f(a/x) \rightarrow (\exists x)f
\end{equation}
with
\begin{equation}
(\exists x)(x = a) \wedge f(a/x) \rightarrow (\exists x)f
\end{equation}
All the other axioms are kept unmodified. The premise $(\exists x)(x =
a)$ is all that is needed to make $a$ behave like a name with a
reference. The solution is somehow similar to the one offered by
Lejewski, if one reads the condition $(\exists x)(x = a)$ as a
predicate and interprets quantifiers under their unrestricted
interpretation. But unfortunately, as we have already mentioned, this
universe is flat also. We are free to talk about Pegasus and dragons,
but we still cannot explain our commitment to the existence of
definite referents.

\subsection{Hobbs's ontological promiscuity}

Quine~\shortcite[p.~47]{quine49}  specifies that
\begin{quote}
We seem to have a
continuum of possible ontologies here, ranging from a radical realism
at the one extreme, where even a left-hand parenthesis or the dot of
an ``i'' has some weird abstract entity as designatum, to a complete
nihilism at the other extreme.
\end{quote}
On the nihilistic extreme we find Hobbs's approach, presented
in~\cite{hobbs85} and later developed and implemented
in~\cite{hobbs93}.  Hobbs is mainly interested in solving three
classical problems: opaque adverbials, the distinction between {\em de
re} and {\em de dicto} belief reports, and the problem of identity in
intensional contexts. Hobbs's approach is constructed on Lejewski and
Hintikka's idea that the set of things we can talk about, therefore
including nonexistent things also, and the set of things we quantify
over, are equal. His language does not contain modalities, intensions,
nor even negation. His approach does not assume the existence of any
object unless this is explicitly specified or logically derivable. For
example, sentence~\bsent{ross-zeus} is represented in
formula~\ref{ross worships zeus}.
\bexample{Ross worships Zeus. \name{ross-zeus}}
\begin{equation}
        worship^\prime(E,Ross,Zeus) \wedge Exist(E) \label{ross worships zeus}
\end{equation}
The first conjunct says that $E$ is the event of worshiping Zeus by
Ross,  and the second says that $E$ exists in the real world. Hobbs
assigns a {\em transparency} property to the predicates. For 
$worship^\prime$, this property entails the existence of its second
argument in the physical world:
\begin{equation}
\forall E \forall x \forall y ((worship^\prime(E,x,y) \wedge Exist(E))
\rightarrow Exist(x)) \label{worship' transparency}
\end{equation}
Apparently, the commitment to Ross's existence is solved. {\em
Worship} is a transparent verb regarding its second argument, so in
accordance to this, Ross is existent.  Hirst~\shortcite{hirst91}
studies the implications induced by the introduction of the concept of
{\em anti-transparent} predicates into the system.  Nevertheless, this
allows one to derive nonexistence but it does not give any insights
into the presuppositional phenomenon. As one can
utter~\bsent{ross-zeus}, she can also utter~\bsent{king-zeus}.
\bexample{The king of Buganda worships Zeus. \name{king-zeus}}
For the latter sentence two things can happen assuming we know nothing
about Buganda:
\begin{itemize}
\item If we translate it as $worship^\prime(E,king\_of\_Buganda,Zeus)$,
we cannot fire the transparency axiom related to predicate $worship^\prime$,
so there is no way to explain our commitment to the existence of
Buganda's king. 
\item  If we translate it as $worship^\prime(E,king\_of\_Buganda,Zeus)
\wedge Exist(E)$ we can infer via the transparency axiom that the king
of Buganda exists. Once we do that, we are no longer able to cancel
this inference, although we may find later that Buganda has no king.
\end{itemize}
The choice of the appropriate translation is not a simple one, so from
a presuppositional perspective, Hobbs's theory will not do it either.

\subsection{Atlas's aboutness and topicality}

Atlas argues that the existential puzzles arise neither from the
mistake of taking existence as a first-order predicate, nor from
equating naming with meaning but due to a conceptual mistake, namely,
a mistake as to what existential statements are {\em about}.
According to Atlas, {\em aboutness} is a fundamental topic in
linguistics and philosophical logic, one that if it is overlooked
creates important problems. For example, even if the singular term
{\em the queen of England} is used both in~\bsent{queen1}
and~\bsent{queen2}, only the former presupposes that the singular term
is used to refer to someone, or in other words only the first
statement is {\em about} the queen of England. ``If the queen of
England were not, one should not be talking {\ul about} anyone when he
uses the term {\ul the queen of England}''~\cite[p.~377]{atlas88}.  The
latter sentence~\bsent{queen2} does not presuppose that there is a
queen of England because one would not fail to talking {\em about}
anything. 
\bexample{The queen of England raises race horses. \name{queen1}}
\bexample{It is the queen of England who raises race horses. \name{queen2}}

According to Atlas's theory, a statement $A(t)$ presupposes that {\em $t$
exists} only if $t$ is a topic noun phrase. Atlas establishes a set of
criteria for noun phrase topicality~\shortcite[p.~385]{atlas88}, but he
gives no hint of how this theory could be extended to deal with
factives or verbs of judging, and defining the notions of aboutness
and topicality for them is not trivial. Even if we did manage to do
this, such presuppositions can never be cancelled. Either they are
generated or they are not. This leads us to believe that sentences
such as~\bsent{mary-yes-no}, {\em John does not regret that Mary came
to the party because she didn't come}, cannot be appropriately
interpreted in this manner.  A good reason for circumventing the
notion of topicality is given by Lasersohn~\shortcite{lasersohn93}.

\subsection{Lasersohn's context dependency}

We interpret Lasersohn's theory~\shortcite{lasersohn93}  as a
natural extension of Labov's work on the boundaries of words and their
meaning~\cite{labov73}.  William Labov noticed that the distinction
between {\em cup} and {\em bowl} is affected more by whether the
interpreter is situated in a {\em coffee} or in a {\em mashed potatoes}
context than by factors such as the ratio of height to diameter of the
artifact. The same thing happens with sentences containing non-referring
definite descriptions. Some of them are clearly
false~\bsent{king-chair}, while others are
ambiguous~\bsent{king-france}.  
\bexample{The King of France is sitting in that chair.
\name{king-chair}} 
\bexample{The King of France is bald. \name{king-france}}
To explain this dependency on context, Lasersohn uses a semantics
constructed in terms of Data Sets~\cite{veltman81}. An ordering
relation, $\leq$, is defined on the data sets: $D \leq D'$ if $D'$ is
a consistent superset of $D$. Sentence~\bsent{king-chair} seems to be
false because {\em even if we suspend our knowledge that there is no
king of France, there is no way of consistently extending our
information to include the proposition that the King of France is
sitting in the chair}~\cite[p.~116]{lasersohn93}.  This is a subtle
observation and our contribution will subsume it. The only
inconveniences we notice are the ones related to the use of the
Russellian paraphrase in capturing the existential commitment, and the
fact that Data Sets are not computationally attractive.

\section{Nonexistence and existential presuppositions --- a solution}

\subsection{Methodological principles}
\label{meinong principles}

The approach to nonexistent objects and presuppositions that we are going
to present is constructed on the basis of a modified  set of Meinongian
principles about nonexistence. They are embedded in a stratified logic
framework in which quantifiers are taken under Lejewski's unrestricted
interpretation. The ontology is enhanced with the eight types of
existence listed by Hirst~\shortcite{hirst91}, though in this thesis, we
will deal only with physical existence, represented as $E!$,
unactualized existence, represented as $U\!E!$, existence
outside the world but with causal interaction with that world,
$EOW!$, and existence in fiction, $F!$. 

Following Rapaport's style~\shortcite{rapaport85a}, we propose a set
of methodological principles based on Meinong~\shortcite{meinong04}
that are meant to capture the ability of an intelligent agent to deal
with existence and nonexistence rather from a conversational
perspective than from a rational one.

\begin{description}
        \item[MC1.] Every uttered sentence is {\em directed} towards an
{\em object}, because every uttered sentence can be seen as a
materialization of a mental act.

        \item[MC2.] All uttered sentences exist (technically, {\em
have being}). However, this does not imply the existence of their
referents, which are {\em ausserseiend} (beyond being and non-being).

        \item[MC3.] It is not self-contradictory to deny, nor tautologous to
affirm, the existence of a referent.

        \item[MC4.] Every referent and every uttered sentence  has properties.

        \item[MC5.] The principles MC2 and MC4 are not inconsistent.

        Corollary: Even referents of an uttered sentence that do not
exist have properties. 

        \item[MC6.] (a) Every set of properties (Sosein) corresponds to the
utterance of a sentence. \\
                    (b) Every object of thought can be uttered.

        \item[MC7.] Some referents of an utterance are incomplete (undetermined with
respect to some properties).
\end{description}

In accordance with Grice~\shortcite{grice75}, we need two additional
principles: 
\begin{description}
\item[GC1.] The speaker is committed to the truth of the sentences he utters.

\item[GC2.] Using and deriving  presuppositions requires, from both speaker and
listener, a sort of ``optimism''. 
\end{description}
Principle GC1 is formalized by the translation of the uttered
sentences into classical logic formulas in which quantifiers are read
under their unrestricted interpretation. Principle GC2 is formalized
by the rules containing defeasible information that exist in the
knowledge base of the speaker and the hearer, and the notion of
optimism in the model-ordering relation. Note that a non-optimistic
interpretation of utterances will never be able to account for any of
the pragmatic inferences, because they are not explicitly uttered.

\subsection{Formalizing existential presuppositions}

We have already seen how in stratified first-order logic one can
express the pragmatic information associated with the word {\em
bachelor} (see formulas~\ref{bachelor}).

That uttering definite references imply the existence of their referents
constitutes another example of pragmatic inference.  We can capture
this either by adding a new formula~\ref{ref-eq} to our knowledge
base, and by embedding syntactic terms into the logical form, as Hobbs
did~\shortcite{hobbs85}, or by representing this defeasible commitment
explicitly in the translation of each utterance containing a definite
reference or proper noun.
\begin{equation} 
(\forallu x)({\mbox {\sf definite\_reference(x)}} \rightarrow E!^d(x))
\label{ref-eq}
\end{equation}
Both approaches exhibit the same semantic behavior, and due to the
model-ordering relation they explain our commitment to a referent's
existence (in the case that we do not know otherwise). Because ${\mbox
{\sf definite\_reference(x)}}$ is syntactic information, we depict it
using a different font, but the reader should understand that ${\mbox
{\sf x}}$ is bound by the same quantifier as $x$ is.

As a last step, we abandon the Fregean reading of the quantifiers and
we adopt Lejewski's unrestricted
interpretation~\shortcite{lejewski54}. This means that $\exists$ and
$\forall$ do not mix quantification with ontological commitment:
$(\exists x)object(x)$ does not entail the  existence of $x$,
so the set of things we can talk about equals the set of things we can
quantify over.

\subsection{What the approach can do with existent and nonexistent objects}

Assume that someone utters sentence~\bsent{buganda-ex} {\em The king
of Buganda is (not) bald.} If we know nothing about Buganda and its
king, the complete theory of this utterance and the available
knowledge in stratified logic is this: 
\begin{equation}
\left\{
\begin{array}{l}
uttered((\exists x)(king\_of\_buganda(x) \wedge {\mbox {\sf
definite\_reference(x)}} \wedge (\neg) bald(x))) \\  
(\forallu x)({\mbox {\sf definite\_reference(x)}} \rightarrow E!^d(x))
\\
\end{array} \right.
\end{equation}
This theory has one optimistic model~\ref{kob-mo} that reflects one's
commitment to the king's existence. The king's existence has the
status of felicitously defeasible information; it is derived using
knowledge of language use and, according to
definition~\ref{presupposition-def}, is a presupposition of the
utterance.
\begin{equation}
\begin{array}{l}
m = \{king\_of\_buganda(\xi_0), (\neg)bald(\xi_0)\}  
\hspace*{1mm}  \cup \hspace{1mm} \O^i \hspace{1mm}  \cup \hspace{1mm}
\{E!^d(\xi_0)\}  
\end{array}
\label{kob-mo}
\end{equation}

Knowledge about the political system of France can inhibit
the inference regarding the existence of its king in a sentence such
as~\bsent{kof-bald}.
\bexample{The king of France is (not) bald. \name{kof-bald}}
Assume that we know there is no king of France $(\neg E!)$.  A
complete formalization follows:
\begin{equation}
\left\{
\begin{array}{l}
uttered((\exists x)(king\_of\_france(x) \wedge {\mbox {\sf
definite\_reference(x)}} \wedge (\neg) bald(x))) \\ 
(\forallu x)({\mbox {\sf definite\_reference(x)}} \rightarrow E!^d(x))
\\
(\forall x)( king\_of\_france(x) \rightarrow \neg E!(x)) \\
\end{array} \right.
\end{equation}
For this theory, we obtain only one model schemata:

\vspace{5mm}
\noindent 
{\small 
\begin{center} \begin{tabular}{|l|l|l|l|} \hline \hline
{\em Schema \#} & {\em Indefeasible} & {\em Infelicitously defeasible}
&  {\em Felicitously defeasible} \\ \hline 
$m_0$ & $king\_of\_france(\xi_o)$      &  &       \\
      & $(\neg) bald(\xi_0)$  &   &     \\
      & $\neg E!(\xi_0)$ &  &  $E!^d(\xi_0)$ \\ \hline
\end{tabular} \end{center}
}
\vspace{5mm}

\noindent One can notice that the existential presupposition is now
cancelled by some background knowledge. The only way one can satisfy
the initial theory is if she has a stratified structure where $\neg
E!(\xi_0)$.  Thus, the theory yields one model:
\begin{equation} 
\begin{array}{l} 
m = \{king\_of\_france(\xi_0), (\neg)bald(\xi_0),   \neg
E!(\xi_0)\} \hspace{1mm}  \cup  \hspace{1mm} \O^i  \hspace{1mm}
\cup \hspace{1mm} \O^d    
\end{array}
\end{equation}

Asserting existence or nonexistence affects {\em defeasible
inferences} due to knowledge of language use and restricts some of the
models.  If someone utters~\bsent{kob-exists} and we know nothing
about Buganda, the translation~\ref{kob-exists-eq} gives one
model~\ref{kob-exists-mo}.
\bexample{The king of Buganda exists. \name{kob-exists}}
\begin{equation}
\left\{
\begin{array}{l}
uttered((\exists x)(king\_of\_buganda(x) \wedge {\mbox {\sf
definite\_reference(x)}}   \wedge E!(x))) \\ 
(\forallu x)({\mbox {\sf definite\_reference(x)}} \rightarrow E!^d(x))
\end{array} 
\right.
\label{kob-exists-eq}
\end{equation}
\begin{equation} 
\begin{array}{l} 
m = \{ king\_of\_buganda(\xi_0),  E!(\xi_0) \} \hspace{1mm}  \cup
\hspace{1mm} \O^i \hspace{1mm} \cup \hspace{1mm} \O^d    
\end{array} 
\label{kob-exists-mo}
\end{equation}
If we {\em know} that the king of Buganda does not exist, or in other
words we evaluate the above sentence against a knowledge base that
contains
\begin{equation}
(\forall x)(king\_of\_buganda(x) \rightarrow \neg E!(x)), 
\end{equation}  
there is no model for this theory, so the utterance is interpreted as
false.  It is noteworthy that the inconsistency appears due to
specific knowledge about the king's physical existence and not because
of a quantification convention as in classical first-order logic. On
the other hand, the negation~\bsent{king does not exist}, {\em The
king of Buganda does not exist}, 
is consistent with the knowledge base and provides
model~\ref{kob-not-exist-eq}.
\begin{equation}
 m = \{king\_of\_buganda(\xi_0), \neg E!(\xi_0)\} \hspace{1mm}  \cup
\hspace{1mm} \O^i \hspace{1mm} \cup \hspace{1mm} \O^d
\label{kob-not-exist-eq}
\end{equation}

Our approach subsumes the work done by
Lasersohn~\shortcite{lasersohn93} because sentence~\bsent{king-chair},
{\em The King of France is sitting on that chair}, is false regardless
of whether the king of France exists or not, if we know there is no one
on the chair. The corresponding theory is {\em u-inconsistent}, so it
has no stratified models.
\begin{equation}
\left\{
\begin{array}{l}
uttered(sitting(king\_of\_france,that\_chair) \wedge \\
\hspace*{5mm} {\mbox {\sf definite\_reference(king\_of\_france)}}
\wedge {\mbox {\sf definite\_reference(that\_chair)}}) \\
(\forallu x)({\mbox {\sf definite\_reference(x)}} \rightarrow E!^d(x))
\\
(\forall x)(\neg sitting(x, that\_chair)) \\
\end{array} 
\right.
\end{equation}

One can see now that the proposed method is general in the sense that
it captures all presuppositional environments.  Reconsider for example
utterance~\bsent{fel-pres}, {\em John does not regret that Mary came to the
party}. A complete formalization in stratified logic follows:
\begin{equation}
\left\{
\begin{array}{l}
uttered(\neg regret(john,\$come(mary,party)) \wedge {\mbox {\sf
definite\_reference(john)}} \wedge \\ 
\hspace*{3mm} {\mbox {\sf definite\_reference(mary)}} \wedge 
{\mbox {\sf definite\_reference(party)}}) \\
(\forallu x)({\mbox {\sf definite\_reference(x)}} \rightarrow E!^d(x))
\\
(\forallu x,y,z)(\neg regret(x, \$come(y,z))  \rightarrow come^d(y,z))
\\ 
(\forallu x,y,z)(regret(x, \$come(y,z))  \rightarrow come^i(y,z))
\end{array} \right.
\end{equation}   
The optimistic model computed by our program is this:
\begin{equation}
\begin{array}{l}
m = \{ \neg regret(john,\$come(mary,party)) \} \hspace{1mm} \cup
\hspace{1mm} \O^i \hspace{1mm} \cup \\
\hspace*{10mm}   \{E!^d(john), E!^d(mary), E!^d(party),
come^d(mary,party) \} 
\end{array}
\end{equation}
This model reflects our intuitions that Mary came to the
party and that all the definite references exist. 

If one utters now sentence~\bsent{repair}, the new model computed by
our program will reflect the fact that a presupposition has been
cancelled, even though this cancellation occurred later in the
discourse. Thus, the new optimistic model will be~\ref{repair-mo}. 
\bexample{Of course he doesn't. Mary did not come to the party.
\name{repair}} 
\begin{equation}
\begin{array}{l}
m = \{ \neg regret(john,\$come(mary,party)), \neg come(mary,party)
\} \hspace{1mm} \cup \hspace{1mm} \O^i \hspace{1mm} \cup \\
\hspace*{10mm} \{ E!^d(john), E!^d(mary), E!^d(party) \}  
\end{array}
\label{repair-mo}
\end{equation}

Our approach correctly handles references to unactualized objects such
as averted \linebreak strikes. An appropriate formalization for
sentence~\bsent{strike} is theory~\ref{strike-eq}.
\bexample{The strike was averted. \name{strike}} 
\begin{equation}
\left\{
\begin{array}{l}
uttered((\exists x)(strike(x) \wedge {\mbox {\sf definite\_reference(x)}} 
\wedge averted(x))) \\ 
(\forallu x)({\mbox {\sf definite\_reference(x)}} \rightarrow E!^d(x))
\\
(\forall x)(averted(x) \rightarrow U\!E!(x)) \\
(\forall x)(U\!E!(x) \rightarrow \neg E!(x)) \\
\end{array} \right.
\label{strike-eq}
\end{equation}
This gives one optimistic model:
\begin{equation}
\begin{array}{l} 
m = \{strike(\xi_0), averted(\xi_0), \neg
E!(\xi_0), U\!E!(\xi_0) \} \hspace*{1mm} \cup \hspace{1mm} \O^i
\hspace{1mm} \cup \hspace{1mm}  \{ E!^d(\xi_0) \} \\  
\end{array}
\end{equation}

\section{A comparison with Parsons's and Hobbs's work}

\subsection{On Parsons's evidence for his theory of nonexistence}

Parsons argues that is impossible to distinguish between the shape of
the logical form of two sentences like these, in which one subject is
fictional and the other is real: 
\bexample{Sherlock Holmes is more famous than any other detective.
\name{p1}} 
\bexample{Pel\'{e} is more famous than any other soccer player.
\name{p2}} 
In our approach, similar syntactic translations give different
semantic models when interpreted against different knowledge
bases. A complete theory for sentence~\bsent{p1} is this:
\begin{equation}
\left\{ 
\begin{array}{l}
uttered((\exists x)(sherlock\_holmes(x) \wedge {\mbox {\sf
definite\_reference(x)}}   \\  
\hspace*{3mm} \wedge (\forall y)((detective(y) \wedge (x \neq y))
\rightarrow \\
\hspace*{5mm} more\_famous(x,y))))   \\
(\forallu x)({\mbox {\sf definite\_reference(x)}} \rightarrow E!^d(x)) \\
(\forall x)( sherlock\_holmes(x) \rightarrow F!(x)) \\
(\forall x)( F!(x) \rightarrow \neg E!(x))
\end{array} 
\right.
\end{equation}
This theory gives only one model:
\begin{equation}
\begin{array}{l} 
m = \{sherlock\_holmes(\xi_0), \neg E!(\xi_0),
F!(\xi_0 ), \\
\hspace*{10mm} detective(y), more\_famous(\xi_0,y)\} \hspace*{1mm}
\cup \hspace{1mm} \O^i \hspace{1mm} \cup \hspace{1mm}  \O^d 
\end{array}  
\end{equation}
This corresponds to an object $\xi_{0}$ that does not exist in the
real world but exists as a fiction, has the property of being Sherlock
Holmes, and for any other object $y$, real or fictional that has the
property of being a detective, the object $\xi_{0}$ is more famous
than object $y$.  Of course, in this model, it is impossible to commit
ourselves to Holmes's physical existence, but is possible to talk
about him.

The theory for the second sentence~\bsent{p2} is this: 
\begin{equation}
\left\{ 
\begin{array}{l}
uttered((\exists x)(pele(x) \wedge {\mbox {\sf definite\_reference(x)}}  \wedge \\
\hspace*{3mm} (\forall y)((soccer\_player(y) \wedge  
(x \neq y)) \rightarrow \\ \hspace*{5mm} more\_famous(x,y)))) \\
(\forallu x)({\mbox {\sf definite\_reference(x)}} \rightarrow E!^d(x)) \\
\end{array} 
\right.
\end{equation}
This theory exhibits one optimistic model:
\begin{equation}
\begin{array} {l}
m =  \{pele(\xi_0), soccer\_player(y), \\ 
\hspace*{10mm} more\_famous(\xi_0,y)\} \hspace*{1mm} \cup \hspace*{1mm}
 \O^i \hspace{1mm} \cup \hspace{1mm} \{ E^d!(\xi_0) \}   \\
\end{array}
\end{equation}
Model $m$ states that the object $\xi_{0}$, being Pel\'{e}, exists in a
defeasible sense and this is the existential presupposition of the
initial utterance.

As seen, it is needless to mention the existence of specific objects
in the knowledge base.  The model-ordering relation rejects anyhow
models that are not optimistic. In this way, the commitment to Pel\'{e}'s
existence is preserved, and appears as a presupposition of the
utterance. Parsons's theory provides different logical forms for the
above sentences, but fails to avoid the commitment to nonexistent
objects.

\subsection{A comparison with Hobbs's work}

We have mentioned that Hobbs's transparency pertains to relations and
not to objects. In our approach, a sentence such as~\bsent{ross-zeus},
{\em Ross worships Zeus}, can be satisfied by a set of semantic models
that correspond to each possible combination of the existence and
nonexistence of Ross and Zeus.
\begin{equation} 
\left\{ 
\begin{array}{l}
uttered((\exists x)(\exists y)(ross(x) \wedge zeus(y) \wedge
worship(x,y) \\
\hspace*{3mm} \wedge {\mbox {\sf definite\_reference(x)}} \wedge {\mbox {\sf
definite\_reference(y)}})) \\
(\forallu x)({\mbox {\sf definite\_reference(x)}} \rightarrow E!^d(x))
\end{array} \right.
\end{equation}
Among them, only one is optimistic: the one that explains the commitment
to both  Ross's  and Zeus's existence. 
\begin{equation}
\begin{array}{l} 
m = \{ross(\xi_0), zeus(\xi_1), worship(\xi_0,\xi_1) \} 
\hspace*{1mm} \cup \hspace*{1mm} 
 \O^i \hspace{1mm} \cup \hspace{1mm} \{ E!^d(\xi_0), E!^d(\xi_1) \}
\end{array} 
\end{equation}

\noindent But let us  assume we know that there is no  entity in the
real world that enjoys the property of being Zeus, but rather one who
exists outside the real world as a god $(EOW!)$.
\begin{equation}
\left\{ 
\begin{array}{l}
uttered((\exists x)(\exists y)(ross(x) \wedge zeus(y) \wedge worship(x,y)
 \\
\hspace*{3mm} \wedge {\mbox {\sf definite\_reference(x)}} \wedge {\mbox {\sf
definite\_reference(y)}})) \\
(\forallu x)({\mbox {\sf definite\_reference(x)}} \rightarrow E!^d(x)) \\
(\forall x)(zeus(x) \rightarrow EOW!(x)) \\
(\forall x)(EOW!(x) \rightarrow \neg E!(x)) \\
\end{array} \right.
\end{equation}
This theory is no longer satisfiable by a model in which Zeus exists
as a physical entity. However, the optimistic model explains our
commitment to Ross's existence.
\begin{equation}
\begin{array}{l} 
m = \{ross(\xi_0), zeus(\xi_1), \neg E!(\xi_1), EOW!(\xi_1),
worship(\xi_0,\xi_1) \} \hspace*{1mm} \\
\hspace*{10mm}  \cup \hspace*{1mm} 
 \O^i \hspace{1mm} \cup \hspace{1mm} \{ E!^d(\xi_0) \} 
\end{array} 
\end{equation}

        % existential presuppositions
\chapter{Implementation issues}
\label{implementation issues}

Stratified logic has been implemented in Common Lisp and it makes
extensive use of two packages: {\em Screamer}~\cite{siskind91}, a set
of nondeterministic macros, and {\em Iterate}~\cite{amsterdam90}, an
iteration macro similar to {\em loop}.  The interested reader can find
our program in annex~\ref{thesis-code}. Details on nondeterministic
programming and Screamer are given by~\cite{siskind93b,siskind93a}.

\section{Representation}

Constants, variables, functions, and predicates are represented as
Lisp symbols. Stratified predicates are prefixed by {\em U-, I-,} or
{\em D-}: for example, $adult^d(x)$ is represented as \mbox{{\em
(D-adult x)}}. Terms and atomic formulas such as $come(Mary,party)$
are represented as Lisp s-expressions:
\mbox{{\em (come Mary party)}}. Utterances are represented as Lisp
s-expressions as well, as in {\em (uttered (bachelor cousin))}.
Compound formulas are represented as follows:

\vspace{5mm}
{\small 
\begin{center} \begin{tabular}{|c|c|} \hline \hline
Formula & Lisp representation \\ \hline
$\Phi \wedge \Psi$ & (and $\Phi$ $\Psi$) \\
$\Phi \vee \Psi$ & (or $\Phi$ $\Psi$) \\ 
$\neg \Phi$ & (not $\Phi$) \\
$\Phi \rightarrow \Psi$ & (implies $\Phi$ $\Psi$) \\ 
$\Phi \leftrightarrow \Psi$ & (iff $\Phi$ $\Psi$) \\ 
$\exists x \Phi$ & (exists $x$ $\Phi$) \\
$\forall x \Phi$ & (forall $x$ $\Phi$) \\
$uttered \, \Phi$ & (uttered $\Phi$) \\
$\forallu \vec{x} (\Phi \rightarrow \Psi$) & (forallutt ($\vec{x}$)
(implies $\Phi$ $\Psi$)) \\ \hline
\end{tabular} \end{center}
}
\vspace{5mm}

Any symbol that appears as the first element in a list except {\em
and, or, not, implies, iff, exists, forall, uttered}, and {\em
forallutt} is interpreted as a function symbol or predicate. The
predicates are the symbols that occur as first elements of top level
list, or as first elements of a list embedded in a logical
connective or utterance. All the other first symbols are function
symbols. 

The function symbols that are prefixed by the $\$$ sign are
used to restrict the universal quantification only to the objects that
constitute the arguments of that function. Consider the following two
formulas:
\begin{equation}
\left\{
\begin{array}{l}
Iregret(was\_bad(movie)) \\
(\forall x)(Iregret(was\_bad(x)) \rightarrow was\_bad^i(x)) \\
\end{array}
\right.
\end{equation} 
The first one is the logical translation of utterance~\bsent{Iregret}.
\bexample{I regret that the movie was bad. \name{Iregret}}
The second one captures the pragmatic inference that a factive, {\em
regret}, implies infelicitously defeasibly its complement. If we are
to apply exhaustively the universal instantiation rule, we will have
to do it for two logical objects: {\em movie} and {\em
was\_bad(movie)}. This will yield meaningless formulas such as
\begin{equation}
Iregret(was\_bad(was\_bad(movie))) \rightarrow
was\_bad^i(was\_bad(movie)) 
\end{equation} 
To avoid meaningless formulas, we prefix the function symbols that
formalizes linguistic sentences with the $\$$ sign. The convention we
use removes $\$$ sign functions from the set of objects over which an
universal quantifier instantiates. Therefore, a correct formalization
for utterance~\bsent{Iregret} and the adequate pragmatic knowledge
should be as follows:
\begin{equation}
\left\{
\begin{array}{l}
Iregret(\$was\_bad(movie)) \\
(\forall x)(Iregret(\$was\_bad(x)) \rightarrow was\_bad^i(x)) \\
\end{array}
\right.
\end{equation}
Under the convention given above, the universal instantiation rule is
applied for only one object: {\em movie}.

\section{The algorithms}

Our algorithm (see figure~\ref{main algorithm}) takes as input a set of
first-order stratified formulas $\Phi$ that represents an adequate
knowledge-base and the translation of an utterance or set of
utterances $uttered(U)$. The output of the program is a set of
felicitous optimistic model schemata that correspond to an optimistic
interpretation of the given utterances.  The program also outputs the
set of presuppositions that are carried by the uttered sentences; if
the utterances are infelicitous it reports that.

\begin{figure}[htbp]
\centering
\myboxnew{htbp}{13cm}{
\noindent{\bf Input:} $\Phi \cup uttered(U)$ 

\begin{tabbing}
Find the set ${\cal MS}$ of all model schemata for the theory
$\Phi \cup uttered(U)$. \\
{\bf if} ${\cal MS} = \O$ \= {\bf then} {\bf return} $U$ is false. \\
			     \> {\bf else} \=  Find the set ${\cal FMS}$
of felicitous model schemata. \\
			     \> \> {\bf if} ${\cal FMS} = \O$ \= {\bf
then} {\bf return} $U$ is infelicitous. \\
			     \> \> \> {\bf else} \= Find the set ${\cal
FOMS}$ of felicitous \\
			     \> \> \> \> \hspace*{10mm}  optimistic
schemata. \\
			     \> \> \> \> Collect the presuppositions. \\
			     \> \> \> \> Report the results. \\
			     \> \> {\bf endif} \\
{\bf endif} \\
\end{tabbing}
} % end mybox
\caption{The Main Algorithm}
\label{main algorithm}
\end{figure}

\subsection{Finding the model schemata}

Finding the model schemata is the most difficult task. The solution
generalizes Siskind's nondeterministic approach for constructing
first-order tableaux~\shortcite{siskind91}. The implementation makes
use of the following data structures:
\begin{description}
\item[$\Gamma$] is the queue of formulas that remain to be processed.
\item[$M$] is a list of true and negated stratified formulas that
constitute the model schema constructed for the initial set $\Gamma$.
\item[$\Delta$] is a list of pairs $(formula,const)$ that monitors the
constants on which a universal formula has been already instantiated.
\item[$I$] is a list of pairs $(utterance, (\vec{cont}))$ that monitors
the constants on which a pragmatic rule should be applied.
\end{description}

We also use the following functions:
\begin{description}
\item[$push\_formula(\alpha,\Gamma)$] pushes formula $\alpha$ on the queue
$\Gamma$;
\item[$pop\_formula(\Gamma)$] pops the first formula from the queue $\Gamma$;
\item[$ground\_terms()$] returns all the ground terms that occur at a
specific moment on a branch;
\item[$uttered\_terms(\alpha)$]  returns all the ground terms that occur at a
specific moment on an utterance about $\alpha$.
\end{description}

To find all the model schemata we construct an exhausted tableau for
the initial theory. Each exhausted open branch yields a model schema
by collecting the positive and negative atomic formulas on it.

\begin{figure}[htbp]
\centering

\myboxnew{htbp}{13cm}{

\begin{tabbing}
$\Gamma := \Phi \cup uttered(U)$; \\
$M := \O$; \\
${\bf while}$ \= $\neg empty(\Gamma) {\bf do}$ \\
	\> $\alpha := pop\_formula(\Gamma)$; \\
	\> ${\bf case}$ \= $\alpha$  \\
	\> \> $\alpha_1 \wedge \alpha_2$: $push\_formula(\alpha_1,\Gamma)$
and $push\_formula(\alpha_2,\Gamma)$; \\
	\> \> $\alpha_1 \vee \alpha_2$: $push\_formula(\alpha_1,\Gamma)$
or $push\_formula(\alpha_2,\Gamma)$; \\
	\> \> $\alpha_1 \rightarrow \alpha_2$: $push\_formula(\neg
\alpha_1 \vee \alpha_2,\Gamma)$; \\
	\> \> $\vdots$ \\
	\> \> $\alpha_1^i, \neg \alpha_1^i, \alpha_1^d, or \neg
\alpha_1^d$, and $\alpha_1$ is atomic: $M := M \cup \alpha$; \\
	\> \> $\alpha^u$ is atomic: {\bf if} $\neg \alpha^u \in M$ \=
{\bf then} {\bf Fail}; \\
	\> \> \> {\bf else} $M := M \cup \alpha^u$; \\
	\> \> $\neg \alpha^u$ is atomic: {\bf if} $\alpha^u \in M$ \=
{\bf then} {\bf Fail};  \\
	\> \> \> {\bf else} $M := M \cup \{\neg \alpha^u\}$; \\
	\> \> $(\exists x) \alpha_1$:
$push\_formula(\alpha_1(x/gensym()),\Gamma)$; \\
	\> \> $(\forall x) \alpha_1$: \= $T := ground\_terms()$ ; \\
	\> \> \> {\bf for} $t \in T$ {\bf do}
$push\_formula(\alpha_1(x/t),\Gamma)$; \\
	\> \> \> $push\_formula((\forall x) \alpha_1,\Gamma)$; \\
	\> \> \> {\bf endfor} \\
	\> \> $(\forallu x) \alpha_1$: \= $T := uttered\_terms()$ ; \\
	\> \> \> {\bf for} $t \in T$ {\bf do}
$push\_formula(\alpha_1(x/t),\Gamma)$; \\
	\> \> \> $push\_formula((\forallu x) \alpha_1,\Gamma)$; \\
	\> \> \> {\bf endfor} \\
	\> {\bf endcase} \\
{\bf endwhile} \\	
\end{tabbing}
} % end mybox
\caption{The Algorithm for Building a Model Schema}
\label{building schemata}
\end{figure}

Algorithm~\ref{building schemata} terminates when there are no
formulas left on $\Gamma$.  Infinite loops due to universal
instantiation are avoided by keeping track in $\Delta$ of the
constants each universal formula has been instantiated. When an
attempt is made to instantiate a universally quantified formula on the
same constants, that formula is removed from $\Gamma$. Alternatives in
constructing the tableau are explored using {\em Screamer}'s
nondeterministic construct {\em either}  that allows one to
keep only one branch in memory at a time. To collect all the model
schemata, we have used the {\em all-values} macro.

\subsection{Finding the felicitous model schemata}

If all models for an utterance are infelicitous, i.e., they are not
{\em i-satisfiable} (see definition~\ref{infelicity-def}), the
utterance is infelicitous. The felicitous model schemata are obtained
by preserving from the set ${\cal MS}$ only the model schemata that
are felicitous, i.e., the ones in which no infelicitously defeasible
information is cancelled. The algorithm is $O(n)$, where $n$ is the
number of model schemata.

\subsection{Finding the optimistic model schemata}

The algorithm is $O(n^2)$ and it compares the list of models ${\cal
FMS}$ two by two and eliminates the ones that are not optimistic with
respect to definition~\ref{optimistic-models}.

\subsection{Collecting the presuppositions} 

We have seen that presuppositions are associated with the defeasible
inferences that are triggered by pragmatic maxims. Therefore, in order
to collect the presuppositions for a given set of utterances, all
defeasible inferences are considered as presupposition candidates. The
ones that are not cancelled in any optimistic model are projected as
presuppositions for the initial set of utterances.

  	% implementation issues
\chapter{Conclusions}

\section{Contributions}

This thesis has provided a principled approach to the notion of {\em
infelicity} and shown how pragmatic infelicities, i.e., those that
appear when {\em infelicitously defeasible pragmatic inferences} are
cancelled, can be signalled.  We exploited a fundamental difference
between the semantic content of an utterance and the inferences that can
be drawn from this content, and the pragmatic content and its afferent
inferences. First-order logic and the notion of entailment seem
appropriate for characterizing the first class because semantic
information is undefeasible. Since implicatures and presuppositions are
not explicitly uttered, one can argue that {\em all} pragmatic
inferences are defeasible.  First-order logic cannot accommodate the
notion of nonmonotonic reasoning; hence, in order to account for the
pragmatic phenomena a formalism able to handle defeasible information is
needed.

The crucial observation we made is that a taxonomy that has a finer
granularity is needed if one wants to consider more subtle aspects of
natural language interaction. Hence, we noted that pragmatic information
may be inferred with different degrees of strength or commitment: some
pragmatic inferences are felicitous to defeat, while others, are
infelicitously defeasible.  Figure~\ref{pragmatic-strength} shows the
properties that pragmatic inferences have with respect to their
cancelability. The reader should interpret the taxonomy we propose only
as a proof that there are differences between the degree of commitment that
characterizes different types of pragmatic inferences. No claims are
made that the classification we give is definitive or exhaustive.
Future research could reveal that a finer granularity is needed. However,
figure~\ref{pragmatic-strength} constitutes a strong argument for
considering a taxonomy of natural language inferences with at least
three layers:
\begin{itemize}
\item The {\em undefeasible layer} characterizes the semantic
information and all the inferences that derive from it.
\item The {\em infelicitously defeasible layer} characterizes the
pragmatic inferences that are infelicitous to defeat.
\item The {\em felicitously defeasible layer} characterizes the
pragmatic inferences that are felicitous to defeat. 
\end{itemize}

\begin{figure}[htbp]
\centering
\begin{small}
\begin{tabular}{|l|l|c|} \hline \hline
{\em Pragmatic inferences} & {\em The strength of pragmatic} &
{\em Example} \\
 & {\em inferences with respect to} & \\
 & {\em  their cancelability} & \\ \hline
Implicatures derived using the & Infelicitously defeasible &
Sentence~\bsent{infel1} \\ 
maxim of quality & & \\  \hline 
Implicatures derived using the  & Felicitously defeasible &
Sentence~\bsent{quantity-maxim} \\ 
maxim of quantity & & \\ \hline
Implicatures derived using the & Felicitously defeasible &
Sentence~\bsent{relevance-maxim} \\
maxims of relevance & &  \\ \hline
Implicatures derived using the & Infelicitously defeasible &
Sentence~\bsent{infel2} \\ 
maxims of manner (order) & & \\ \hline
Particularized implicatures & Felicitously defeasible &
Sentence~\bsent{particularized-implicature} \\ \hline
Floating implicatures & Felicitously defeasible &
Sentence~\bsent{floating-implicature} \\ \hline
Scalar implicatures & Felicitously defeasible &
Sentence~\bsent{scalar-implicature} \\ \hline
Clausal implicatures & Felicitously defeasible &
Sentence~\bsent{clausal-implicature} \\ \hline
Presuppositions in affirmative sentences & Infelicitously defeasible &
Sentence~\bsent{infel3} \\ \hline
Presuppositions in negative sentences & Felicitously defeasible &
Sentence~\bsent{mary-yes-no} \\ \hline
\end{tabular}
\end{small}
\caption{The strength of pragmatic inferences with respect to their
cancelability} 
\label{pragmatic-strength}
\end{figure}

In order to deal with these three layers, we have devised a new logical
framework, called {\em stratified logic}. Stratified logic allows one to
account not only for the nonmonotonic aspects that are specific to
pragmatic reasoning, but also for the determination of infelicitous
utterances.  The framework yields a formal definition for the notion of
{\em infelicitous utterance} and {\em presupposition}, and a tractable
algorithm for computing interpretations for utterances, for determining
their associated presuppositions, and for signalling infelicities. The
proposed method is general, i.e., it handles simple and complex
utterances, sequences of utterances, and it gives a uniform explanation
for the presuppositions carried by definite references and the ones
determined by other presupposition environments. The solution for the
existential presuppositions is not subject to the classical logic
puzzles. 

We have contrived our framework having in mind that the result ought to
be computationally tractable. A generalization of the Beth tableau
technique provides the theoretical foundation for our implementation.
Stratified logic is implemented in Common Lisp and it makes extensive
use of the  nondeterministic facilities of the Screamer
system~\cite{siskind91}. The result is a program that takes as input a
set of stratified formulas that constitute the necessary semantic and
pragmatic knowledge and the logical translation of an utterance or set
of utterances and that computes a set of {\em optimistic}
interpretations for the given utterances. The program computes for each
set of utterances the associated presuppositions, and signals when an
infelicitous sentence has been uttered.

\section{Future work}

There are a number of important directions in which one can expand this
research:

\paragraph{{\bf Consider agents with different knowledge bases:}} There are no
restrictions that prevent our system from considering that the
participants in a conversation do not share the same knowledge. If one
does so, we expect that she will be able to distinguish between the
pragmatic inferences that pertain to different participants in a
conversation; to understand better the way they interact; to see how
semantic and pragmatic information is transferred between them; and to
determine where semantic or pragmatic misunderstandings occur.

\paragraph{{\bf Extend the notion of {\em infelicity} to syntax,
semantics, and speech acts:}} We believe that a principled account can
be given for the other types of infelicities using the same ideas we
have presented in this thesis.  For example, in order to signal a
syntactic infelicity, such as the one that occurs when the noun in a
sentence does not have the same number as the corresponding verb,
one will have to loosen the syntactic rules. The same methodology that
we used when we split the classical notion of truth into undefeasible,
infelicitously defeasible, and felicitously defeasible truth should be
applicable at this level as well. The number agreement rule will be
assigned an {\em infelicitously defeasible} strength, so that the
syntactic analysis will proceed, but the infelicity is signalled.

Formalizing infelicities in speech acts seems to be a more complex issue
because one will have to accommodate the different levels of truth
within a theory of action. In a universe that changes over time it is
not even clear {\em when} an infelicity occurs. Assume for example that
John utters sentence~\bsent{last-one} during a conversation he has with
Robert.
\bexample{I promise I'll play badminton with you tomorrow. See you at 8
o'clock at the gym. \name{last-one}} 
If John is not able to get up in time, his promise will be infelicitous,
but it is not very clear when did his utterance become infelicitous. At
the moment he uttered it?  At the moment John did not manage to get
up?  At the moment Robert realized that John is not going to show up? Or
at the moment when they met the next day and Robert reminded John that
he had waited for him.

\paragraph{{\bf Study the relation between stratified logic and other
knowledge representation formalisms:}} So far, we have shown only that
stratified logic is an appropriate tool for modelling the pragmatic
inferences and for signalling infelicities. It would be interesting to
see exactly what the place is of stratified logic with respect to the
other frameworks that are concerned with nonmonotonic reasoning, what
the problems are that stratified logic is able to solve, and what its
drawbacks are.

\setstretch{1.00} % for the appendix
\appendix
\chapter{Computing pragmatic inferences and infelicities}
\label{thesis-code}

\begin{footnotesize}
\begin{verbatim}

;;; Stratified Semantic Tableaux


(in-package :screamer-user)


;;;;;;;;;;;;;;;;;;;;;;;;;;;;;;;;;;;;;;;;;;;
;;; Handyman's corner - general 
;;;;;;;;;;;;;;;;;;;;;;;;;;;;;;;;;;;;;;;;;;;

;;; explode a symbol into a list of characters

(defun explode(symbol)
  (map 'list
       #'(lambda(x) (intern (string x)))
       (symbol-name symbol)))


;;; implode a list of characters into a symbol

(defun implode(lista)
        (read-from-string
                (coerce (map 'list
                        #'(lambda(x) (coerce x 'character))
                        lista) 'string)))


;;; collapse a list three times
;;; ((((a b)) (((((a) c d))))) -> (a b (a) c d)


(defun collapse-list-thrice (l)
  (let ((rez nil))
    (mapc 
     #'(lambda(x)
         (mapc 
          #'(lambda(y) 
              (mapc
               #'(lambda(z) (push z rez))
               y))
          x))
     l)
    rez))


(defun collapse-list-twice (l)
  (let ((rez nil))
         (mapc 
          #'(lambda(y) 
              (mapc
               #'(lambda(z) (push z rez))
               y))
          l)
    rez))

;;; flatten a list
;;; ((A) (((b) x))) -> (A b x)

(defun flatten(l)
  (cond ((null l) nil)
        ((atom (first l)) (cons (first l) (flatten (rest l))))
        (t (append (flatten (first l)) (flatten (rest l))))))

;;; remove special symbols from a list ($)

(defun remove-special (l)
  (remove-if #'(lambda(a) (if (eq (first (explode a)) '$) t nil))
             l))

;;; 

(defun is-list-in (l1 l2)
  (cond ((null l1) t)
        (t (and 
            (member (car l1) l2)
            (is-list-in (rest l1) l2)))))


;;;;;;;;;;;;;;;;;;;;;;;;;;;;;;;;;;;;;;;;;;;;;;;;;;;;;;;;;;;;;;;
;;; Handyman's corner - functions specific to stratified logic
;;;;;;;;;;;;;;;;;;;;;;;;;;;;;;;;;;;;;;;;;;;;;;;;;;;;;;;;;;;;;;;


(defun atomic-formula-type (phi)
  (when (equal (first phi) 'not) (setq phi (second phi)))
  (let* ((tip1 (first (explode (first phi))))
         (tip2 (second (explode (first phi))))
         (tip (implode (list tip1 tip2))))
    (case tip
          (I- 'I)
          (D- 'D)
          (otherwise 'U))))


(defun remove-predicate-type(pred)
  (let ((pred-list (explode pred)))
    (if (and 
           (or (equal (car pred-list) 'U)
               (equal (car pred-list) 'I)
               (equal (car pred-list) 'D))
           (equal (second pred-list) '-))
          (implode (rest (rest pred-list)))
          pred)))

      
(defun change-strength-atomic-formula (phi strength)
  (let* ((f (if (equal (first phi) 'not) (second phi) phi))
         (negated? (if (equal (first phi) 'not) t nil))
         ; (tip (atomic-formula-type phi))
         (cleaned-pred (remove-predicate-type (first f))))
    (unless (equal strength 'U) 
            (setq cleaned-pred 
                  (implode (append (list strength '-)
                                   (explode cleaned-pred)))))
    (if negated? 
        `(not ,(cons cleaned-pred (rest f)))
        (cons cleaned-pred (rest f)))))
         


;;;;;;;;;;;;;;;;;;;;;;;;;;;;;;;;;;;;;;;;;;;;;;;;;;
;;; Stratified semantic tableaux - or a variation 
;;; on a Siskindian theme
;;;;;;;;;;;;;;;;;;;;;;;;;;;;;;;;;;;;;;;;;;;;;;;;;;


(defvar *gamma* nil)         ; the queue of formulas that remain
                             ; to be processed

(defvar *model* nil)         ; a list of true and negated atomic formulas
                             ; constituting the model constructed for the
                             ; initial set *gamma*

(defvar *my-gensym-counter* -1) ; the index of the last symbol created by gensym

(defvar *delta* nil)         ; used during the universal instantiation

(defvar *instantiation* nil) ; contains an association list that 
                             ; memorizes the utterances and their 
                             ; associated constants

(defun empty? ()             ; returns T if the queue of formulas is empty
  (null *gamma*))


(defun pop-formula()
  (when (empty?) (break "Attempt to pop an empty queue"))
  (let ((phi  (first *gamma*)))
    (local (setf *gamma* (rest *gamma*)))
    phi))


(defun push-end-formula(phi)
  (if (empty?)
      (local (setf *gamma* (list phi)))
      (local (setf *gamma* (append *gamma* (list phi))))))


(defun push-formula(phi)
  (local (setf *gamma* (cons phi *gamma*))))


(defun negate(phi)
  (if (and (listp phi) (= (length phi) 2) (eq (first phi) 'not))
      (second phi)
      `(not ,phi)))


(defun add-to-model(phi &optional (utterance? nil))
  (when (and 
         (member (negate phi) *model* :test #'equal)
         (equal (atomic-formula-type phi) 'U))
        (fail)) ; fails only if a U-contradiction occurs
    (unless (member phi *model* :test #'equal)
                 (local (setf *model* (cons phi *model*))))
    (let ((temp nil)
          (phi (if (eq (first phi) 'not) (second phi) phi)))
     (when utterance?
          (if (setf temp (assoc (first phi) *instantiation*))
              (local (setf (second temp)
                           (cons (remove-special (flatten (rest phi)))
                                 (second temp))))
              (local (setf *instantiation* 
                         (cons (list (first phi) 
                                     (list (remove-special 
                                            (flatten (rest phi)))))
                               *instantiation*))))))
    *instantiation*)
    



(defun empty-model () '())

(defun subst-term (s v f)  ; substitutes term s for all free occurences
                           ; of v in the formula f.    f(v->s)
 (case (first f)
   (and `(and ,@(iterate (for g in (rest f)) (collect (subst-term s v g)))))
   (or `(or ,@(iterate (for g in (rest f)) (collect (subst-term s v g)))))
   (not `(not ,(subst-term s v (second f))))
   (implies `(implies ,(subst s v (second f)) ,(subst s v (third f))))
   (iff `(iff ,(subst s v (second f)) ,(subst s v (third f))))
   (exists (if (eq v (second f))
               f
               `(exists ,(second f) ,(subst-term s v (third f)))))
   (forall (if (eq v (second f))
               f
               `(forall ,(second f) ,(subst-term s v (third f)))))
   (otherwise (subst s v f :test #'eq))))




(defun ground-term? (s variables)
 (or (and (symbolp s) 
          (not (member s variables :test #'equal)))
     (and (consp s)
          (every #'(lambda (u) (ground-term? u variables)) (rest s)))))



(defun a-subterm-of (s)
 (either
  s
  (progn (unless (consp s) (fail)) (a-subterm-of (a-member-of (rest s))))))



(defun a-ground-subterm-of (f &optional variables)
 (if (member f variables :test #'eq) (fail))
 (if (symbolp f)
     f
     (case (first f)
       (uttered (a-ground-subterm-of (second f) variables))
       (and (a-ground-subterm-of (a-member-of (rest f)) variables))
       (or (a-ground-subterm-of (a-member-of (rest f)) variables))
       (not (a-ground-subterm-of (second f) variables))
       (implies (a-ground-subterm-of (either (second f) (third f)) variables))
       (iff (a-ground-subterm-of (either (second f) (third f)) variables))
       (exists (a-ground-subterm-of (third f) (cons (second f) variables)))
       (forall (a-ground-subterm-of (third f) (cons (second f) variables)))
       (forallutt (a-ground-subterm-of (third f) (append (second f) variables)))
       (otherwise (let ((s (a-subterm-of (a-member-of (rest f)))))
                   (unless (ground-term? s variables) (fail))
                   s)))))


(defun a-ground-term()
  (a-ground-subterm-of (either (a-member-of *gamma*) (a-member-of *model*))))


(defun ground-terms()
  (remove-duplicates (all-values (a-ground-term))))


(defun make-constant-symbol()
  (local (setf *my-gensym-counter* (+ 1 *my-gensym-counter*)))
  (implode (coerce (format nil "C~D" *my-gensym-counter*) 'list)))


(defun loop-over(count)
;  (format t "~% count ~D " count)
;  (format t " ~% --- ~A ~% ---- ~A ~% ---- ~A ~% ----  ~% --- ~D ~%~%~% "
;             *gamma* *model* *delta* *instantiation*  *my-gensym-counter*)
;  (BREAK)
    (unless (or (zerop count) (empty?))
      (let* ((phi (pop-formula)))
        (case (first phi)
           (uttered ; it is an utterance
             (setf phi (second phi))
             (case (first phi)
               (not 
                (case (first (second phi))
                   (not (push-formula (list 'uttered (second (second phi)))))
                   (and (push-formula (list 'uttered 
                                        `(or ,@(iterate
                                         (for g in (rest (second phi)))
                                         (collect `(not ,g)))))))
                   (or (push-formula (list 'uttered 
                                         `(and ,@(iterate
                                          (for g in (rest (second phi)))
                                          (collect `(not ,g)))))))
                   (implies (push-formula (list 'uttered
                                            `(and ,(second (second phi)) 
                                             (not ,(third (second phi)))))))
                   (iff (push-formula (list 'uttered 
                                         `(iff ,(second (second phi))
                                            (not ,(third (second phi)))))))
                   (exists (push-formula (list 'uttered
                                            `(forall ,(second (second phi))
                                              (not ,(third (second phi)))))))
                   (forall (push-formula (list 'uttered 
                                            `(exists ,(second (second phi))
                                              (not ,(third (second phi)))))))
                   (otherwise (add-to-model phi t)))  ; ! no rule for 
                                                ; (not (forallutt form))
                (loop-over (length *gamma*)))
               (and (mapc #'(lambda(x) (push-formula (list 'uttered x)))
                              (rest phi))
                    (loop-over (length *gamma*)))
               (or  (push-formula (list 'uttered (a-member-of (rest phi))))
                    (loop-over (length *gamma*)))
               (implies (push-formula (list 'uttered
                                       `(or (not ,(second phi)) ,(third phi))))
                        (loop-over (length *gamma*)))
               (iff (push-formula (list 'uttered
                                    `(and (implies ,(second phi) ,(third phi))
                                    (implies ,(third phi) ,(second phi)))))
                (loop-over (length *gamma*)))
               (exists (push-formula (list 'uttered
                                     (subst (make-constant-symbol) ; add this! 
                                        (second phi)
                                        (third phi))))
                   (loop-over (length *gamma*)))
               (forall (let ((again? nil)
                             (gt (ground-terms)))
                         (cond ((null gt)
                                (push-formula (list 'uttered
                                                 (subst (make-constant-symbol)
                                                   (second phi)
                                                   (third phi))))
                                (global (setf again? t)))
                           (t
                              (mapc 
                                #'(lambda(x)
                                   (unless (or 
                                            ; is not yet expanded
                                            (member (list phi x) *delta*
                                                   :test #'equal)
                                            ; is not a special function
                                            (and 
                                             (listp x)
                                             (equal `$ 
                                                 (first (explode (first x))))))
                                     (local (setf *delta* (cons (list phi x) 
                                                                *delta*)))
                                     (push-formula (list 'uttered
                                                       (subst x
                                                          (second phi)
                                                          (third phi))))
                                     (global (setf again? t))))
                                gt)))
                     (push-end-formula (list 'uttered phi))
                     (if again? 
                         (loop-over (length *gamma*))
                         (loop-over (- count 1)))))
               (otherwise
                (add-to-model phi t)
                (loop-over (length *gamma*)))))
           (not 
             (case (first (second phi))
               (not (push-formula (second (second phi))))
               (and (push-formula `(or ,@(iterate
                                         (for g in (rest (second phi)))
                                         (collect `(not ,g))))))
               (or (push-formula `(and ,@(iterate
                                          (for g in (rest (second phi)))
                                          (collect `(not ,g))))))
               (implies (push-formula `(and ,(second (second phi)) 
                                             (not ,(third (second phi))))))
               (iff (push-formula `(iff ,(second (second phi))
                                        (not ,(third (second phi))))))
               (exists (push-formula `(forall ,(second (second phi))
                                              (not ,(third (second phi))))))
               (forall (push-formula `(exists ,(second (second phi))
                                              (not ,(third (second phi))))))
               (otherwise (add-to-model phi)))  ; ! no rule for 
                                                ; (not (forallutt form))
             (loop-over (length *gamma*)))
           (and (mapc #'push-formula (rest phi))
                (loop-over (length *gamma*)))
           (or  (push-formula (a-member-of (rest phi)))
                (loop-over (length *gamma*)))
           (implies (push-formula `(or (not ,(second phi)) ,(third phi)))
                    (loop-over (length *gamma*)))
           (iff (push-formula `(and (implies ,(second phi) ,(third phi))
                                    (implies ,(third phi) ,(second phi))))
                (loop-over (length *gamma*)))
           (exists (push-formula (subst (make-constant-symbol) ; add this! 
                                        (second phi)
                                        (third phi)))
                   (loop-over (length *gamma*)))
           (forall (let ((again2? nil)
                         (gt (ground-terms)))
                     (cond ((null gt)
                              (push-formula (subst (make-constant-symbol)
                                                   (second phi)
                                                   (third phi)))
                              (global (setf again2? t)))
                           (t
                              (mapc 
                                #'(lambda(x)
                                   (unless (or 
                                            ; is not yet expanded
                                            (member (list phi x) *delta*
                                                   :test #'equal)
                                            ; is not a special function
                                            (and 
                                             (listp x)
                                             (equal `$ 
                                                 (first (explode (first x))))))
                                     (local (setf *delta* (cons (list phi x) 
                                                                *delta*)))
                                     (push-formula (subst x
                                                          (second phi)
                                                          (third phi)))
                                     (global (setf again2? t))))
                                gt)))
                     (push-end-formula phi)
                     (if again2? 
                         (loop-over (length *gamma*))
                         (loop-over (- count 1)))))
           (forallutt (let* ((again3? nil)
                             (param (second phi))
                             (antecedent (second (third phi)))
                             (ant (if (equal (first antecedent) 'not) 
                                      (first (second antecedent))
                                      (first antecedent)))
                             (gtc (assoc ant  *instantiation*))
                             (gt (if gtc (second gtc)  nil)))
                        (unless (null gt)
                         (unless (member (list phi gt) *delta* :test #'equal)
                            (local (setf *delta* (cons (list phi gt) *delta*)))
                            (let ((newphi phi))
                              (mapc 
                               #'(lambda(xx)
                                   (mapc 
                                    #'(lambda (p val)
                                        (setf newphi (subst val p newphi)))
                                    param
                                    xx)
                                   (push-formula (third newphi))
                                   (setf newphi phi)
                                   (global (setf again3? t)))
                               gt))
                            (global (setf again3? t))))
                        (push-end-formula phi)
                        (if again3?
                            (loop-over (length *gamma*))
                            (loop-over (- count 1)))))
           (otherwise (add-to-model phi)
                      (loop-over (length *gamma*)))))))



(defun a-model-of (sigma)
  (global (setf *gamma* nil)
          (setf *model* nil)
          (setf *delta* nil)
          (setf *instantiation* nil)
          (setf *my-gensym-counter* -1))
  (mapc #'push-end-formula sigma)
  (loop-over (length *gamma*))
  *model*)


(defun entails? (sigma phi)
  (all-values (a-model-of (cons (negate phi) sigma))))

(defun find-models (sigma)
  (remove-duplicates (all-values (a-model-of sigma)) :test #'equal))



;;;;;;;;;;;;;;;;;;;;;;;;;;;;;;;;;;;;;;;;;;;;;;;;
;;; Model specific functions - the basics
;;;;;;;;;;;;;;;;;;;;;;;;;;;;;;;;;;;;;;;;;;;;;;;;



(defun arrange-model (formulas)
  (let ((models '()))
    (mapc
     #'(lambda(x) 
         (let* ((Ufo (change-strength-atomic-formula x 'U))
                (negd? (if (equal (first x) 'not) 'yes 'no))
                (fftype (atomic-formula-type x))
                (pUfo (if (equal (first x) 'not) (second Ufo) Ufo)))
           (unless (member pUfo models :test #'equal  :key #'first)
               (push (list pUfo nil nil nil nil nil nil) models))
           (case fftype
             (U 
              (case negd?
                (yes 
                 (setf (third (first (member pUfo models 
                                            :test #'equal :key #'first))) t))
                (otherwise
                 (setf (second (first (member pUfo models 
                                            :test #'equal :key #'first))) t))))
             (I
              (case negd?
                (yes 
                 (setf (fifth (first (member pUfo models 
                                            :test #'equal :key #'first))) t))
                (otherwise
                 (setf (fourth (first (member pUfo models 
                                            :test #'equal :key #'first))) t))))
             (D
              (case negd?
                (yes 
                       (setf (seventh (first (member pUfo models
                                            :test #'equal :key #'first))) t))
                (otherwise
                 (setf (sixth (first (member pUfo models 
                                            :test #'equal :key #'first))) t))))
             (otherwise
              (break "Unknown predicate type")))))
     formulas)
    models))
              


(defun get-all-models2 (formulas) ; the main function
  (let ((all-models nil)
        (felicitous-models nil)
        (felicitous-minimal-models nil))
    (mapc #'(lambda(x) (push (arrange-model x) all-models)) 
          (find-models formulas))
    (print 'MODELS)
    (print-models all-models)
    (if (null all-models)
        (format t "~% The utterance is false")
      (progn
        (print-model-ordering all-models #'smaller-content-3)
        (setf felicitous-models
              (remove-if #'null 
                         (mapcar 
                          #'(lambda(x) (is-the-model-felicitous? x))
                          all-models)))
        (if (null felicitous-models)
            (format t "~% The utterance is infelicitous")
          (progn
            (print 'FELICITOUS-MODELS)
            (print-models felicitous-models)
            (print-model-ordering felicitous-models #'smaller-content-3)
            (setf felicitous-minimal-models 
                  (keep-minimal-models felicitous-models #'smaller-content-3))
          (print 'FELICITOUS-OPTIMISTIC-MODELS)
          (print-models felicitous-minimal-models) 
          (print 'PRESUPPOSITIONS)
          (mapc #'(lambda(x) (print x)) 
                (collect-presuppositions felicitous-minimal-models))))))
    (terpri) 
    (terpri) 
    t
    ))


(defun print-one-model(model)
  (mapc #'(lambda(x)
            (format t "~% ~A" (first x))
            (when (second x) (format t "~41T*"))
            (when (third x) (format t "~47T*"))
            (when (fourth x) (format t "~56T*"))
            (when (fifth x) (format t "~62T*"))
            (when (sixth x) (format t "~71T*"))
            (when (seventh x) (format t "~77T*")))
        model))

(defun print-models (models)
  (let ((no 0))
    (mapc #'(lambda(x) 
              (format t "~% Model number ~D" no)
              (unless (is-the-model-felicitous? x)
                    (format t " is infelicitous."))
              (format t "~% Formulas ~40TUndefeasible ~55TInfelicitous ~70TDefeasible")
              (format t "~% ~40TAff ~46TNeg ~55TAff ~61TNeg ~70TAff ~76TNeg")
              (incf no)
              (print-one-model x)
0              (format t "~%~%"))
          models)))



;;;;;;;;;;;;;;;;;;;;;;;;;;;;;;;;;;;;;;;;;;;;;;;;;;;;;;
;;; Model ordering functions
;;;;;;;;;;;;;;;;;;;;;;;;;;;;;;;;;;;;;;;;;;;;;;;;;;;;;;


(defun print-model-ordering (models ordering-function)
  (let ((counter1 0))
    (mapc  
     #'(lambda(m1)
         (do* ((go-over models (rest go-over))
               (counter2 0 (+ counter2 1))
               (m2 (when (listp go-over) (first go-over))
                   (when (listp go-over) (first go-over))))
              ((null go-over))
              (unless                       ; I do not compare the same models 
               (or (null m2)
                   (equal (member m1 models)
                          (member m2 models)))
               (when (funcall ordering-function m1 m2)
                     (format t "~% m~D > m~D" counter1 counter2))))
         (setf counter1 (+ counter1 1)))
   models)))





;;; O(n^2) algorithm
(defun keep-minimal-models (models ordering-function)
  (mapc 
   #'(lambda(m1)
       (block pupu
       (do* ((go-over models (rest go-over))
             (m2 (when (listp go-over) (first go-over))
                 (when (listp go-over) (first go-over))))
            ((null go-over))
            (unless                        ; I do not compare the same models 
             (or (null m2)
                 (equal (member m1 models)
                        (member m2 models)))
             (when (funcall ordering-function m2 m1)
                   (setf (first (member m1 models :test #'equal)) nil)
                   (return-from pupu))))))
;            (print models)))
   models)
  (remove-if #'null models))



(defun smaller-content-3(m1 m2)
  (block pupu
    (mapc 
     #'(lambda(x)
         (let ((y (if (member (first x) m1 :test #'equal :key #'first) 
                      (first (member (first x) m1 :test #'equal :key #'first))
                      nil))
               (aff2? nil)
               (aff1? nil)
               (where1? nil)
               (where2? nil)
               (bool 'yes)
               (cancel1? nil)
               (cancel2? nil)
               (where 1))
            (when (null y) (return-from pupu nil)) ; x does not belong to m1
            (mapc #'(lambda(val2 val1)
                        (when (and val1 (null aff1?)) 
                              (setf aff1? bool))
                        (when (and val1 (not (equal aff1? bool)))
                              (setf cancel1? t))
                        (when val1
                              (push where where1?))
                        (when (and val2 (null aff2?)) 
                              (setf aff2? bool))
                        (when (and val2 (not (equal aff2? bool)))
                              (setf cancel2? t))
                        (when val2
                              (push where where2?))
                        (setf bool (if (eq bool 'yes) 'no 'yes))
                        (setf where (+ where 1)))
                  (rest x) (rest y))
            (unless (and (not cancel1?) cancel2?)
                    (when (not (eq aff1? aff2?)) (return-from pupu nil))
                    (when (and
                           (not (equal where1? where2?))
                           (is-list-in where1? where2?))
                          (return-from pupu nil))
                    (when (< (apply #'max where1?) (apply #'max where2?))
                          (return-from pupu nil)))))
     m2)
    t))



;;;;;;;;;;;;;;;;;;;;;;;;;;;;;;;;;;;;;;;;;;;;;;;
;;; Analyzing the results
;;;;;;;;;;;;;;;;;;;;;;;;;;;;;;;;;;;;;;;;;;;;;;;


(defun is-the-model-felicitous? (model)
  (block pupu
         (mapc
          #'(lambda(x)
              (when (or
                     (and (second x) (fifth x))
                     (and (third x) (fourth x))
                     (and (fourth x) (fifth x)))
                    (return-from pupu nil)))
          model)
         model))


(defun collect-facts (models)
 (remove-duplicates 
  (remove-if-not #'consp 
                 (collapse-list-thrice models)) :test #'equal))  

(defun collect-presup-candidates (models)
 (remove-duplicates 
  (let ((rez nil))
    (mapc 
     #'(lambda(x)
         (when (or (sixth x) (seventh x))
               (push (first x) rez)))
     (collapse-list-twice models))
    rez)
  :test #'equal))


(defun collect-presuppositions (models)
  (let ((presuppositions nil))
    (mapc
     #'(lambda(fact)
         (block pupu
                (let ((affirmative? nil)) ; yes, no, or nil
                  (mapc 
                   #'(lambda(model)
                       (let ((form (first (member fact model 
                                           :test #'equal :key #'first))))
                        (when form
                         (when (and ; contradiction
                                (or (second form) (fourth form) (sixth form))
                                (or (third form) (fifth form) (seventh form)))
                               (return-from pupu))                              
                         (when (and ; contradiction
                                (or (second form) (fourth form) (sixth form))
                                (eq affirmative? 'no))
                               (return-from pupu))
                         (when (and ; contradiction
                                (or (third form) (fifth form) (seventh form))
                                (eq affirmative? 'yes))
                               (return-from pupu))
                         (when (and ; set the polarity of the formula
                                (or (second form) (fourth form) (sixth form))
                                (or (null affirmative?)
                                    (eq affirmative? 'yes)))
                               (setf affirmative? 'yes))
                         (when (and ; set the polarity of the formula
                                (or (third form) (fifth form) (seventh form))
                                (or (null affirmative?)
                                    (eq affirmative? 'no)))
                               (setf affirmative? 'no)))))
                   models)
                  (if affirmative?
                      (push fact presuppositions)
                    (push `(not ,fact) presuppositions)))))
     (collect-presup-candidates models))
    presuppositions))
                        
  
\end{verbatim}
\end{footnotesize}

\chapter{Stratified logic at work --- a set of examples}

This appendix provides a comprehensive set of examples that shows how
stratified logic works.  Each example contains
\begin{itemize}
\item the utterance or utterances that are analyzed;
\item their logical translation in stratified logic and the relevant
semantic and pragmatic knowledge;
\item the result of the program having as input the corresponding
theory. 
\end{itemize}

\section{Simple presuppositions}

\subsection{My cousin is not a bachelor}

\begin{scriptsize}
\begin{verbatim}

(defun bac1 ()
  (get-all-models2 
   '((uttered (not (bachelor cousin)))
     (forall x (implies (bachelor x) (and (male x) (adult x) (not (married x))))) 
     (forall x (implies (not (bachelor x)) (I-married x)))
     (forallutt (x) (implies (not (bachelor x)) (D-male x)))
     (forallutt (x) (implies (not (bachelor x)) (D-adult x))))))


SCREAMER-USER(28): (bac1)

MODELS 
 Model number 0 is infelicitous.
 Formulas                               Undefeasible   Infelicitous   Defeasible
                                        Aff   Neg      Aff   Neg      Aff   Neg
 (BACHELOR COUSIN)                             *
 (MARRIED COUSIN)                              *        *
 (MALE COUSIN)                           *                             *
 (ADULT COUSIN)                          *                             *


 Model number 1
 Formulas                               Undefeasible   Infelicitous   Defeasible
                                        Aff   Neg      Aff   Neg      Aff   Neg
 (BACHELOR COUSIN)                             *
 (MARRIED COUSIN)                                       *
 (MALE COUSIN)                                                         *
 (ADULT COUSIN)                                                        *


FELICITOUS-MODELS 
 Model number 0
 Formulas                               Undefeasible   Infelicitous   Defeasible
                                        Aff   Neg      Aff   Neg      Aff   Neg
 (BACHELOR COUSIN)                             *
 (MARRIED COUSIN)                                       *
 (MALE COUSIN)                                                         *
 (ADULT COUSIN)                                                        *


FELICITOUS-OPTIMISTIC-MODELS 
 Model number 0
 Formulas                               Undefeasible   Infelicitous   Defeasible
                                        Aff   Neg      Aff   Neg      Aff   Neg
 (BACHELOR COUSIN)                             *
 (MARRIED COUSIN)                                       *
 (MALE COUSIN)                                                         *
 (ADULT COUSIN)                                                        *


PRESUPPOSITIONS 
(ADULT COUSIN) 
(MALE COUSIN) 

\end{verbatim}
\end{scriptsize}

\subsection{John does not regret that Mary came to the party}

\begin{scriptsize}
\begin{verbatim}

(defun jm1()
  (get-all-models2
   '((uttered (not (regret john ($come-party mary))))
     (forallutt (x y) (implies (not (regret x ($come-party y))) 
                               (D-come-party y)))
     (forallutt (x y) (implies (regret x ($come-party y)) 
                               (I-come-party y))))))

SCREAMER-USER(32): (jm1)

MODELS 
 Model number 0
 Formulas                               Undefeasible   Infelicitous   Defeasible
                                        Aff   Neg      Aff   Neg      Aff   Neg
 (REGRET JOHN ($COME-PARTY MARY))              *
 (COME-PARTY MARY)                                      *              *


 Model number 1
 Formulas                               Undefeasible   Infelicitous   Defeasible
                                        Aff   Neg      Aff   Neg      Aff   Neg
 (REGRET JOHN ($COME-PARTY MARY))              *
 (COME-PARTY MARY)                                                     *


FELICITOUS-MODELS 
 Model number 0
 Formulas                               Undefeasible   Infelicitous   Defeasible
                                        Aff   Neg      Aff   Neg      Aff   Neg
 (REGRET JOHN ($COME-PARTY MARY))              *
 (COME-PARTY MARY)                                      *              *


 Model number 1
 Formulas                               Undefeasible   Infelicitous   Defeasible
                                        Aff   Neg      Aff   Neg      Aff   Neg
 (REGRET JOHN ($COME-PARTY MARY))              *
 (COME-PARTY MARY)                                                     *


FELICITOUS-OPTIMISTIC-MODELS 
 Model number 0
 Formulas                               Undefeasible   Infelicitous   Defeasible
                                        Aff   Neg      Aff   Neg      Aff   Neg
 (REGRET JOHN ($COME-PARTY MARY))              *
 (COME-PARTY MARY)                                      *               *


PRESUPPOSITIONS 
(COME-PARTY MARY) 

\end{verbatim}
\end{scriptsize}

\section{Presupposition cancellation}

\subsection{John does not regret that Mary came to the party because she didn't come}

\begin{scriptsize}
\begin{verbatim}

(defun jm3()
  (get-all-models2
   '((uttered (and (not (come-party mary)) 
                   (not (regret john ($come-party mary)))))
     (forallutt (x y) (implies (not (regret x ($come-party y))) 
                               (D-come-party y)))
     (forallutt (x y) (implies (regret x ($come-party y)) 
                               (I-come-party y))))))

SCREAMER-USER(34): (jm3)

MODELS 
 Model number 0 is infelicitous.
 Formulas                               Undefeasible   Infelicitous   Defeasible
                                        Aff   Neg      Aff   Neg      Aff   Neg
 (REGRET JOHN ($COME-PARTY MARY))              *
 (COME-PARTY MARY)                             *        *              *


 Model number 1
 Formulas                               Undefeasible   Infelicitous   Defeasible
                                        Aff   Neg      Aff   Neg      Aff   Neg
 (REGRET JOHN ($COME-PARTY MARY))              *
 (COME-PARTY MARY)                             *                       *


FELICITOUS-MODELS 
 Model number 0
 Formulas                               Undefeasible   Infelicitous   Defeasible
                                        Aff   Neg      Aff   Neg      Aff   Neg
 (REGRET JOHN ($COME-PARTY MARY))              *
 (COME-PARTY MARY)                             *                       *


FELICITOUS-OPTIMISTIC-MODELS 
 Model number 0
 Formulas                               Undefeasible   Infelicitous   Defeasible
                                        Aff   Neg      Aff   Neg      Aff   Neg
 (REGRET JOHN ($COME-PARTY MARY))              *
 (COME-PARTY MARY)                             *                       *


PRESUPPOSITIONS 

\end{verbatim}
\end{scriptsize}

\section{Signalling infelicities}

\subsection{John regrets that Mary came to the party, but she didn't come}

\begin{scriptsize}
\begin{verbatim}

(defun jm4()
  (get-all-models2
   '((uttered (and (not (come-party mary)) 
                   (regret john ($come-party mary))))
     (forallutt (x y) (implies (not (regret x ($come-party y))) 
                               (D-come-party y)))
     (forallutt (x y) (implies (regret x ($come-party y)) 
                               (I-come-party y))))))

SCREAMER-USER(35): (jm4)

MODELS 
 Model number 0 is infelicitous.
 Formulas                               Undefeasible   Infelicitous   Defeasible
                                        Aff   Neg      Aff   Neg      Aff   Neg
 (REGRET JOHN ($COME-PARTY MARY))        *
 (COME-PARTY MARY)                             *        *              *


 Model number 1 is infelicitous.
 Formulas                               Undefeasible   Infelicitous   Defeasible
                                        Aff   Neg      Aff   Neg      Aff   Neg
 (REGRET JOHN ($COME-PARTY MARY))        *
 (COME-PARTY MARY)                             *        *

 The utterance is infelicitous

\end{verbatim}
\end{scriptsize}

\section{Solving the projection problem}

From now on we will reproduce only the felicitous optimistic models
and the associated presuppositions for each utterance or set of
utterances.

\subsection{My cousin is  a bachelor or a spinster {\em or---cancellation}}

\begin{scriptsize}
\begin{verbatim}

(defun bac2 ()
  (get-all-models2 
   '((uttered (or (bachelor cousin) (spinster cousin)))
     (implies (or (bachelor cousin) (spinster cousin))
              (or (not (D-bachelor cousin)) (not (D-spinster cousin))))
     (forall x (implies (bachelor x) (and (male x) (adult x) (not (married x)))))
     (forallutt (x) (implies (not (bachelor x)) (I-married x)))
     (forallutt (x) (implies (not (bachelor x)) (D-male x)))
     (forallutt (x) (implies (not (bachelor x)) (D-adult x)))
     (forall x (implies (spinster x) (and (female x) (adult x) (not (married x)))))
     (forallutt (x) (implies (not (spinster x)) (I-married x)))
     (forallutt (x) (implies (not (spinster x)) (D-female x)))
     (forallutt (x) (implies (not (spinster x)) (D-adult x)))
     (forall x (iff (male x) (not (female x)))))))

SCREAMER-USER(29): (bac2)

FELICITOUS-OPTIMISTIC-MODELS 
 Model number 0
 Formulas                               Undefeasible   Infelicitous   Defeasible
                                        Aff   Neg      Aff   Neg      Aff   Neg
 (SPINSTER COUSIN)                       *
 (BACHELOR COUSIN)                             *                             *
 (MARRIED COUSIN)                              *
 (ADULT COUSIN)                          *
 (FEMALE COUSIN)                         *                             *
 (MALE COUSIN)                                 *


 Model number 1
 Formulas                               Undefeasible   Infelicitous   Defeasible
                                        Aff   Neg      Aff   Neg      Aff   Neg
 (SPINSTER COUSIN)                       *
 (BACHELOR COUSIN)                             *                             *
 (MARRIED COUSIN)                              *
 (FEMALE COUSIN)                         *
 (ADULT COUSIN)                          *                             *
 (MALE COUSIN)                                 *


 Model number 2
 Formulas                               Undefeasible   Infelicitous   Defeasible
                                        Aff   Neg      Aff   Neg      Aff   Neg
 (BACHELOR COUSIN)                       *
 (MARRIED COUSIN)                              *
 (ADULT COUSIN)                          *
 (MALE COUSIN)                           *                             *
 (SPINSTER COUSIN)                             *                             *
 (FEMALE COUSIN)                               *


 Model number 3
 Formulas                               Undefeasible   Infelicitous   Defeasible
                                        Aff   Neg      Aff   Neg      Aff   Neg
 (BACHELOR COUSIN)                       *
 (MARRIED COUSIN)                              *
 (MALE COUSIN)                           *
 (ADULT COUSIN)                          *                             *
 (SPINSTER COUSIN)                             *                             *
 (FEMALE COUSIN)                               *


PRESUPPOSITIONS 
(ADULT COUSIN) 

\end{verbatim}
\end{scriptsize}

\subsection{If Mary came to the party then John will regret that Sue came
to the party --- {\em if ... then ... no---cancellation}}

\begin{scriptsize}
\begin{verbatim}

(defun jms1()
  (get-all-models2
   '((uttered (implies (come-party mary) 
                       (regret john ($come-party sue))))
     (forallutt (x y) (implies (not (regret x ($come-party y))) 
                               (D-come-party y)))
     (forallutt (x y) (implies (regret x ($come-party y)) 
                               (I-come-party y))))))

SCREAMER-USER(36): (jms1)

FELICITOUS-OPTIMISTIC-MODELS 
 Model number 0
 Formulas                               Undefeasible   Infelicitous   Defeasible
                                        Aff   Neg      Aff   Neg      Aff   Neg
 (REGRET JOHN ($COME-PARTY SUE))         *
 (COME-PARTY SUE)                                       *              *


 Model number 1
 Formulas                               Undefeasible   Infelicitous   Defeasible
                                        Aff   Neg      Aff   Neg      Aff   Neg
 (COME-PARTY MARY)                             *


PRESUPPOSITIONS 
(COME-PARTY SUE) 

\end{verbatim}
\end{scriptsize}

\section{Existential presuppositions}
\subsection{Ross worships Zeus (we know nothing about Zeus)}

\begin{scriptsize}
\begin{verbatim}

(defun ross1()
  (get-all-models
   '((exists x 
             (exists y 
                     (and (ross x) 
                          (definite-reference x)
                          (zeus y)
                          (definite-reference y)
                          (worships x y))))
     (forall x (implies (definite-reference x) (D-exists! x))))))

SCREAMER-USER(24): (ross1)

FELICITOUS-OPTIMISTIC-MODELS 
 Model number 0
 Formulas                               Undefeasible   Infelicitous   Defeasible
                                        Aff   Neg      Aff   Neg      Aff   Neg
 (WORSHIPS C0 C1)                        *
 (DEFINITE-REFERENCE C1)                 *
 (ZEUS C1)                               *
 (DEFINITE-REFERENCE C0)                 *
 (ROSS C0)                               *
 (EXISTS! C1)                                                          *
 (EXISTS! C0)                                                          *


PRESUPPOSITIONS 
(EXISTS! C0) 
(EXISTS! C1) 


\end{verbatim}
\end{scriptsize}

\subsection{Ross worships Zeus (we know {\em everything} about Zeus)}

\begin{scriptsize}
\begin{verbatim}

(defun ross2()
  (get-all-models
   '((exists x 
             (exists y 
                     (and (ross x) 
                          (definite-reference x)
                          (zeus y)
                          (definite-reference y)
                          (worships x y))))
     (forall x (iff (ross x) (not (zeus x))))
     (forall x (implies (zeus x) (not (exists! x))))
     (forall x (implies (zeus x) (eow! x)))
     (forall x (implies (definite-reference x) (D-exists! x))))))

FELICITOUS-OPTIMISTIC-MODELS 
 Model number 0
 Formulas                               Undefeasible   Infelicitous   Defeasible
                                        Aff   Neg      Aff   Neg      Aff   Neg
 (WORSHIPS C0 C1)                        *
 (DEFINITE-REFERENCE C1)                 *
 (ZEUS C1)                               *
 (DEFINITE-REFERENCE C0)                 *
 (ROSS C0)                               *
 (ROSS C1)                                     *
 (ZEUS C0)                                     *
 (EOW! C1)                               *
 (EOW! C0)                               *
 (EXISTS! C1)                                  *                       *
 (EXISTS! C0)                                                          *

PRESUPPOSITIONS 
(EXISTS! C0) 

\end{verbatim}
\end{scriptsize}

\section{Sequence of utterances}

\subsection{ John is a not a bachelor. I regret that you have
misunderstood me.  He is only five years old. You realize he cannot
date women. It is not him who is not a bachelor!}

\begin{scriptsize}
\begin{verbatim}

(defun story5()
  (get-all-models2
   '((uttered (not (bachelor john)))
     (uttered (regret speaker ($misunderstood hearer speaker)))
     (uttered (and (male john) (not (adult john))))
     (uttered (realize hearer ($cannot-date-women john)))
     (uttered (cleft ($not-bachelor john)))
     (forall x (implies (bachelor x) (and (male x) (adult x) (not (married x)))))
     (forallutt (x) (implies (not (bachelor x)) (I-married x)))
     (forallutt (x) (implies (not (bachelor x)) (D-male x)))
     (forallutt (x) (implies (not (bachelor x)) (D-adult x)))
     (forallutt (x y) (implies (not (regret x ($misunderstood y x))) 
                               (D-misunderstood y x)))
     (forallutt (x y) (implies (regret x ($misunderstood y x)) 
                               (I-misunderstood y x)))
     (forallutt (x y) (implies (not (realize x ($cannot-date-women y))) 
                               (not (D-date-women y))))
     (forallutt (x y) (implies (realize x ($cannot-date-women y)) 
                               (not (I-date-women y))))
     (forallutt (x) (implies (cleft ($not-bachelor x))
                               (exists y (and (not (bachelor y)) 
                                               (D-exists! y)))))))) 

FELICITOUS-OPTIMISTIC-MODELS 
 Model number 0
 Formulas                               Undefeasible   Infelicitous   Defeasible
                                        Aff   Neg      Aff   Neg      Aff   Neg
 (BACHELOR JOHN)                               *
 (REGRET SPEAKER 
      ($MISUNDERSTOOD HEARER SPEAKER))   *
 (REALIZE HEARER 
      ($CANNOT-DATE-WOMEN JOHN))         *
 (CLEFT ($NOT-BACHELOR JOHN))            *
 (MARRIED SPEAKER)                             *
 (ADULT SPEAKER)                         *
 (MALE SPEAKER)                          *
 (MARRIED HEARER)                              *
 (ADULT HEARER)                          *
 (MALE HEARER)                           *
 (MARRIED JOHN)                                       *
 (MALE JOHN)                             *                             *
 (ADULT JOHN)                                  *                       *
 (MISUNDERSTOOD HEARER SPEAKER)                         *              *
 (DATE-WOMEN JOHN)                                                           *


 Model number 1
 Formulas                               Undefeasible   Infelicitous   Defeasible
                                        Aff   Neg      Aff   Neg      Aff   Neg
 (BACHELOR JOHN)                               *
 (REGRET SPEAKER 
       ($MISUNDERSTOOD HEARER SPEAKER))  *
 (REALIZE HEARER 
       ($CANNOT-DATE-WOMEN JOHN))        *
 (CLEFT ($NOT-BACHELOR JOHN))            *
 (MARRIED SPEAKER)                             *
 (ADULT SPEAKER)                         *
 (MALE SPEAKER)                          *
 (MARRIED HEARER)                              *
 (ADULT HEARER)                          *
 (MALE HEARER)                           *
 (MARRIED JOHN)                                         *
 (MALE JOHN)                             *                             *
 (ADULT JOHN)                                  *                       *
 (MISUNDERSTOOD HEARER SPEAKER)                         *              *
 (DATE-WOMEN JOHN)                                            *
 (EXISTS! C0)                                                          *
 (BACHELOR C0)                                 *
 (MARRIED C0)                                  *
 (ADULT C0)                              *
 (MALE C0)                               *


 Model number 2
 Formulas                               Undefeasible   Infelicitous   Defeasible
                                        Aff   Neg      Aff   Neg      Aff   Neg
 (BACHELOR JOHN)                               *
 (REGRET SPEAKER 
       ($MISUNDERSTOOD HEARER SPEAKER))  *
 (REALIZE HEARER 
       ($CANNOT-DATE-WOMEN JOHN))        *
 (CLEFT ($NOT-BACHELOR JOHN))            *
 (MARRIED SPEAKER)                             *
 (ADULT SPEAKER)                         *
 (MALE SPEAKER)                          *
 (BACHELOR HEARER)                             *
 (MARRIED JOHN)                                         *
 (MALE JOHN)                             *                             *
 (ADULT JOHN)                                  *                       *
 (MISUNDERSTOOD HEARER SPEAKER)                         *              *
 (DATE-WOMEN JOHN)                                                           *


 Model number 3
 Formulas                               Undefeasible   Infelicitous   Defeasible
                                        Aff   Neg      Aff   Neg      Aff   Neg
 (BACHELOR JOHN)                               *
 (REGRET SPEAKER 
       ($MISUNDERSTOOD HEARER SPEAKER))  *
 (REALIZE HEARER 
       ($CANNOT-DATE-WOMEN JOHN))        *
 (CLEFT ($NOT-BACHELOR JOHN))            *
 (MARRIED SPEAKER)                             *
 (ADULT SPEAKER)                         *
 (MALE SPEAKER)                          *
 (BACHELOR HEARER)                             *
 (MARRIED JOHN)                                         *
 (MALE JOHN)                             *                             *
 (ADULT JOHN)                                  *                       *
 (MISUNDERSTOOD HEARER SPEAKER)                         *              *
 (DATE-WOMEN JOHN)                                           *
 (EXISTS! C0)                                                          *
 (BACHELOR C0)                                 *
 (MARRIED C0)                                  *
 (ADULT C0)                              *
 (MALE C0)                               *

 Model number 4
 Formulas                               Undefeasible   Infelicitous   Defeasible
                                        Aff   Neg      Aff   Neg      Aff   Neg
 (BACHELOR JOHN)                               *
 (REGRET SPEAKER 
       ($MISUNDERSTOOD HEARER SPEAKER))  *
 (REALIZE HEARER 
       ($CANNOT-DATE-WOMEN JOHN))        *
 (CLEFT ($NOT-BACHELOR JOHN))            *
 (BACHELOR SPEAKER)                            *
 (MARRIED HEARER)                              *
 (ADULT HEARER)                          *
 (MALE HEARER)                           *
 (MARRIED JOHN)                                         *
 (MALE JOHN)                             *                             *
 (ADULT JOHN)                                  *                       *
 (MISUNDERSTOOD HEARER SPEAKER)                         *              *
 (DATE-WOMEN JOHN)                                                           *


 Model number 5
 Formulas                               Undefeasible   Infelicitous   Defeasible
                                        Aff   Neg      Aff   Neg      Aff   Neg
 (BACHELOR JOHN)                               *
 (REGRET SPEAKER 
       ($MISUNDERSTOOD HEARER SPEAKER))  *
 (REALIZE HEARER 
       ($CANNOT-DATE-WOMEN JOHN))        *
 (CLEFT ($NOT-BACHELOR JOHN))            *
 (BACHELOR SPEAKER)                            *
 (MARRIED HEARER)                              *
 (ADULT HEARER)                          *
 (MALE HEARER)                           *
 (MARRIED JOHN)                                         *
 (MALE JOHN)                             *                             *
 (ADULT JOHN)                                  *                       *
 (MISUNDERSTOOD HEARER SPEAKER)                         *              *
 (DATE-WOMEN JOHN)                                            *
 (EXISTS! C0)                                                          *
 (BACHELOR C0)                                 *
 (MARRIED C0)                                  *
 (ADULT C0)                              *
 (MALE C0)                               *


 Model number 6
 Formulas                               Undefeasible   Infelicitous   Defeasible
                                        Aff   Neg      Aff   Neg      Aff   Neg
 (BACHELOR JOHN)                               *
 (REGRET SPEAKER 
       ($MISUNDERSTOOD HEARER SPEAKER))  *
 (REALIZE HEARER 
       ($CANNOT-DATE-WOMEN JOHN))        *
 (CLEFT ($NOT-BACHELOR JOHN))            *
 (BACHELOR SPEAKER)                            *
 (BACHELOR HEARER)                             *
 (MARRIED JOHN)                                         *
 (MALE JOHN)                             *                             *
 (ADULT JOHN)                                  *                       *
 (MISUNDERSTOOD HEARER SPEAKER)                         *              *
 (DATE-WOMEN JOHN)                                                           *


 Model number 7
 Formulas                               Undefeasible   Infelicitous   Defeasible
                                        Aff   Neg      Aff   Neg      Aff   Neg
 (BACHELOR JOHN)                               *
 (REGRET SPEAKER 
       ($MISUNDERSTOOD HEARER SPEAKER))  *
 (REALIZE HEARER 
       ($CANNOT-DATE-WOMEN JOHN))        *
 (CLEFT ($NOT-BACHELOR JOHN))            *
 (BACHELOR SPEAKER)                            *
 (BACHELOR HEARER)                             *
 (MARRIED JOHN)                                         *
 (MALE JOHN)                             *                             *
 (ADULT JOHN)                                  *                       *
 (MISUNDERSTOOD HEARER SPEAKER)                         *              *
 (DATE-WOMEN JOHN)                                            *
 (EXISTS! C0)                                                          *
 (BACHELOR C0)                                 *
 (MARRIED C0)                                  *
 (ADULT C0)                              *
 (MALE C0)                               *


PRESUPPOSITIONS 
(EXISTS! C0) 
(MISUNDERSTOOD HEARER SPEAKER) 
(MALE JOHN) 
(NOT (DATE-WOMEN JOHN)) 

\end{verbatim}
\end{scriptsize}

%\bibliographystyle{named}
%\bibliography{bibfile}

\begin{thebibliography}{}

\bibitem[\protect\citeauthoryear{Amsterdam}{1990}]{amsterdam90}
J.~Amsterdam.
\newblock The Iterate Manual.
\newblock Technical Report A.I. Memo No. 1236, M.I.T. Artificial Intelligence
  Laboratory, October 1990.

\bibitem[\protect\citeauthoryear{Atlas}{1988}]{atlas88}
J.D. Atlas.
\newblock What are negative existence statements about?
\newblock {\em Linguistics and Philosophy}, 11:373--394, 1988.

\bibitem[\protect\citeauthoryear{Austin}{1962}]{austin62}
J.L. Austin.
\newblock {\em How to Do Things with Words}.
\newblock Harvard University Press, 1962.

\bibitem[\protect\citeauthoryear{Bell and Machover}{1986}]{bell86}
J.L. Bell and M.~Machover.
\newblock {\em A Course in Mathematical Logic}.
\newblock Elsevier Science Publishers B.V., 1986.

\bibitem[\protect\citeauthoryear{Brewka}{1994}]{brewka94}
G.~Brewka.
\newblock Reasoning about priorities in default logic.
\newblock In {\em Proceedings of the Twelfth National Conference on Artificial
  Intelligence}, pages 940--945, 1994.

\bibitem[\protect\citeauthoryear{Chellas}{1980}]{chellas80}
B.F. Chellas.
\newblock {\em Modal Logic: an Introduction}.
\newblock Cambridge University Press, 1980.

\bibitem[\protect\citeauthoryear{Delgrande}{1994}]{delgrande94}
J.P. Delgrande.
\newblock A preference-based approach to default reasoning: Preliminary report.
\newblock In {\em Proceedings of the Twelfth National Conference on Artificial
  Intelligence}, pages 902--908, 1994.

\bibitem[\protect\citeauthoryear{Frege}{1892}]{frege92}
G.~Frege.
\newblock \"{U}ber sinn und bedeutung.
\newblock {\em Z. Philos. Philos. Kritik}, 100:373--394, 1892.
\newblock reprinted as: On Sense and Nominatum, In Feigl H. and Sellars W.,
  editors, {\em Readings in Philosophical Analysis}, pages 85--102,
  Appleton-Century-Croft, New York, 1947.

\bibitem[\protect\citeauthoryear{Gazdar}{1979}]{gazdar79}
G.J.M. Gazdar.
\newblock {\em Pragmatics: Implicature, Presupposition, and Logical Form}.
\newblock Academic Press, 1979.

\bibitem[\protect\citeauthoryear{Geis}{1982}]{geis82}
M.L. Geis.
\newblock {\em The Language of Television Advertising}.
\newblock Academic Press, 1982.

\bibitem[\protect\citeauthoryear{Ginsberg}{1988}]{ginsberg88}
M.L. Ginsberg.
\newblock Multivalued logics: A uniform approach to reasoning in artificial
  intelligence.
\newblock {\em Computational Intelligence}, 4:265--316, 1988.

\bibitem[\protect\citeauthoryear{Gordon and Lakoff}{1975}]{gordon75}
D.~Gordon and G.~Lakoff.
\newblock Conversational postulates.
\newblock In Cole P. and Morgan J.L., editors, {\em Syntax and Semantics,
  Speech Acts}, volume~3, pages 83--106. Academic Press, 1975.

\bibitem[\protect\citeauthoryear{Green}{1989}]{green89}
G.M. Green.
\newblock {\em Pragmatics and Natural Language Understanding}.
\newblock Lawrence Erlbaum Associates, Inc., Publishers, 1989.

\bibitem[\protect\citeauthoryear{Green}{1990}]{green90}
N.~Green.
\newblock Normal state implicature.
\newblock In {\em Proceedings 28th Annual Meeting of the Association for
  Computational Linguistics}, pages 89--96, 1990.

\bibitem[\protect\citeauthoryear{Green}{1992}]{green92}
N.~Green.
\newblock Conversational implicatures in indirect replies.
\newblock In {\em Proceedings 30th Annual Meeting of the Association for
  Computational Linguistics}, pages 64--71, 1992.

\bibitem[\protect\citeauthoryear{Grice}{1975}]{grice75}
H.P. Grice.
\newblock Logic and conversation.
\newblock In Cole P. and Morgan J.L., editors, {\em Syntax and Semantics,
  Speech Acts}, volume~3, pages 41--58. Academic Press, 1975.

\bibitem[\protect\citeauthoryear{Grice}{1978}]{grice78}
H.P. Grice.
\newblock Further notes on logic and conversation.
\newblock In Cole P., editor, {\em Syntax and Semantics, Pragmatics}, volume~9,
  pages 113--127. Academic Press, 1978.

\bibitem[\protect\citeauthoryear{Hintikka}{1959}]{hintikka59}
K.J.J. Hintikka.
\newblock Existential presuppositions and existential commitments.
\newblock {\em Journal of Philosophy}, 56:125--137, 1959.

\bibitem[\protect\citeauthoryear{Hintikka}{1970}]{hintikka70}
K.J.J. Hintikka.
\newblock Existential presuppositions and uniqueness presuppositions.
\newblock In Lambert K., editor, {\em Philosophical Problems in Logic}, pages
  20--55. Humanities Press, New York, 1970.

\bibitem[\protect\citeauthoryear{Hirschberg}{1985}]{hirschberg85}
J.B. Hirschberg.
\newblock A theory of scalar implicature.
\newblock Technical Report MS-CIS-85-56, Department of Computer and Information
  Science, University of Pennsylvania, 1985.

\bibitem[\protect\citeauthoryear{Hirst}{1991}]{hirst91}
G.~Hirst.
\newblock Existence assumptions in knowledge representation.
\newblock {\em Artificial Intelligence}, 49:199--242, 1991.

\bibitem[\protect\citeauthoryear{Hobbs \bgroup \em et al.\egroup
  }{1993}]{hobbs93}
J.R. Hobbs, M.~Stickel, D.~Appelt, and P.~Martin.
\newblock Interpretation as abduction.
\newblock {\em Artificial Intelligence}, 63:69--142, 1993.

\bibitem[\protect\citeauthoryear{Hobbs}{1985}]{hobbs85}
J.R. Hobbs.
\newblock Ontological promiscuity.
\newblock In {\em Proceedings 23rd Annual Meeting of the Association for
  Computational Linguistics}, pages 61--69, 1985.

\bibitem[\protect\citeauthoryear{Horn}{1972}]{horn72}
L.R. Horn.
\newblock {\em On the Semantic Properties of Logical Operators in English}.
\newblock PhD thesis, University of California, Los Angeles, 1972.

\bibitem[\protect\citeauthoryear{Horton}{1987}]{horton87}
D.L. Horton.
\newblock Incorporating agents' beliefs in a model of presupposition.
\newblock Master's thesis, Dept. of Computer Science, University of Toronto,
  1987.
\newblock Tech. Report CSRI-201, Computer Systems Research Institute,
  University of Toronto.

\bibitem[\protect\citeauthoryear{Horton and Hirst}{1988}]{horton88}
D.~Horton and G.~Hirst.
\newblock Presuppositions as beliefs.
\newblock In {\em Proceedings of the International Conference on Computational
  Linguistics, COLING}, pages 255--260, 1988.

\bibitem[\protect\citeauthoryear{Kaplan}{1982}]{kaplan82}
J.~Kaplan.
\newblock Cooperative responses from a portable natural language database query
  system.
\newblock In Brady M. and Berwick R.C., editors, {\em Computational Models of
  Discourse}, pages 167--208. The MIT Press, 1982.

\bibitem[\protect\citeauthoryear{Karttunen and Peters}{1979}]{karttunen79}
L.~Karttunen and S.~Peters.
\newblock Conventional implicature.
\newblock In Oh~C.K. and Dinneen D.A, editors, {\em Syntax and Semantics,
  Presupposition}, volume~11, pages 1--56. Academic Press, 1979.

\bibitem[\protect\citeauthoryear{Karttunen}{1971}]{karttunen71}
L.~Karttunen.
\newblock Implicative verbs.
\newblock {\em Language}, 47:340--358, 1971.

\bibitem[\protect\citeauthoryear{Karttunen}{1974}]{karttunen74}
L.~Karttunen.
\newblock Presupposition and linguistic context.
\newblock {\em Theoretical Linguistics}, 1:3--44, 1974.

\bibitem[\protect\citeauthoryear{Katz and Langendoen}{1976}]{katz76}
J.J. Katz and D.T. Langendoen.
\newblock Pragmatics and presupposition.
\newblock {\em Language}, 52:1--17, 1976.

\bibitem[\protect\citeauthoryear{Kay}{1992}]{kay92}
P.~Kay.
\newblock The inheritance of presuppositions.
\newblock {\em Linguistics \& Philosophy}, 15:333--379, 1992.

\bibitem[\protect\citeauthoryear{Kifer and Lozinskii}{1992}]{kifer92a}
M.~Kifer and E.L. Lozinskii.
\newblock {A} logic for reasoning with inconsistency.
\newblock {\em Journal of Automated Reasoning}, 9(2):179--215, November 1992.

\bibitem[\protect\citeauthoryear{Kifer and Subrahmanian}{1992}]{kifer92b}
M.~Kifer and V.S. Subrahmanian.
\newblock Theory of generalized annotated logic programming and its
  applications.
\newblock {\em Journal of Logic Programming}, 12(4):335--368, April 1992.

\bibitem[\protect\citeauthoryear{Labov}{1973}]{labov73}
W.~Labov.
\newblock The boundaries of words and their meaning.
\newblock In Bailey C.J. and Shuy R., editors, {\em New Ways of Analyzing
  Variation in English}, pages 340--373. Georgetown University Press,
  Washington DC, 1973.

\bibitem[\protect\citeauthoryear{Lasersohn}{1993}]{lasersohn93}
P.~Lasersohn.
\newblock Existence presuppositions and background knowledge.
\newblock {\em Journal of Semantics}, 10:113--122, 1993.

\bibitem[\protect\citeauthoryear{Lejewski}{1954}]{lejewski54}
C.~Lejewski.
\newblock Logic and existence.
\newblock {\em British Journal for the Philosophy of Science}, 5:104--119,
  1954.

\bibitem[\protect\citeauthoryear{Levinson}{1983}]{levinson83}
S.C. Levinson.
\newblock {\em Pragmatics}.
\newblock Cambridge University Press, 1983.

\bibitem[\protect\citeauthoryear{McCarthy}{1980}]{mccarthy80}
J.~McCarthy.
\newblock Circumscription --- a form of nonmonotonic reasoning.
\newblock {\em Artificial Intelligence}, 13:27--40, 1980.

\bibitem[\protect\citeauthoryear{McCarthy}{1986}]{mccarthy86}
J.~McCarthy.
\newblock Applications of circumscription to formalizing common-sense
  knowledge.
\newblock {\em Artificial Intelligence}, 28:89--116, 1986.

\bibitem[\protect\citeauthoryear{Meinong}{1904}]{meinong04}
A.~Meinong.
\newblock \"{U}ber gegenstandstheorie.
\newblock In Meinong A., editor, {\em Untersuchungen zur Gegenstandstheorie und
  Psychologie}. Barth, Leipzig, 1904.
\newblock reprinted in: The theory of objects, Chisholm R.M. editor, {\em
  Realism and the Background of Phenomenology}, pages 76--117. Free Press,
  Glencoe, IL, 1960.

\bibitem[\protect\citeauthoryear{Mercer and Reiter}{1982}]{mercer82}
R.E. Mercer and R.~Reiter.
\newblock The representation of presuppositions using defaults.
\newblock In {\em Proceedings of the Fourth Biennal Conference of the Canadian
  Society for the Computational Studies of Intelligence (CSCSI/SCEIO)}, pages
  103--107, 1982.

\bibitem[\protect\citeauthoryear{Mercer}{1987}]{mercerphd}
R.E. Mercer.
\newblock {\em A Default Logic Approach to the Derivation of Natural Language
  Presuppositions}.
\newblock PhD thesis, Department of Computer Science, University of British
  Columbia, 1987.

\bibitem[\protect\citeauthoryear{Mercer}{1988a}]{mercer88b}
R.E. Mercer.
\newblock Solving some persistent presupposition problems.
\newblock In {\em Proceedings of the International Conference on Computational
  Linguistics, COLING}, pages 420--425, 1988.

\bibitem[\protect\citeauthoryear{Mercer}{1988b}]{mercer88a}
R.E. Mercer.
\newblock Using default logic to derive natural language presuppositions.
\newblock In {\em Proceedings of the Seventh Biennal Conference of the Canadian
  Society for the Computational Studies of Intelligence (CSCSI/SCEIO)}, pages
  14--21, 1988.

\bibitem[\protect\citeauthoryear{Mercer}{1990}]{mercer90}
R.E. Mercer.
\newblock Deriving natural language presuppositions from complex conditionals.
\newblock In {\em Proceedings of the Eighth Biennal Conference of the Canadian
  Society for the Computational Studies of Intelligence (CSCSI/SCEIO)}, pages
  114--120, 1990.

\bibitem[\protect\citeauthoryear{Mercer}{1991}]{mercer91}
R.E. Mercer.
\newblock Presuppositions and default reasoning: A study in lexical pragmatics.
\newblock In Pustejovski J. and Bergler S., editors, {\em ACL SIG Workshop on
  Lexical Semantics and Knowledge Representation}, pages 224--237, 1991.

\bibitem[\protect\citeauthoryear{{Nait Abdallah}}{1989}]{areski88}
A.~{Nait Abdallah}.
\newblock An extended framework for default reasoning.
\newblock In Csirik J., Demetrovics J., and Gecseg F., editors, {\em
  International Conference on Fundamentals of Computation Theory}, pages
  339--348. Springer LNCS 380, 1989.

\bibitem[\protect\citeauthoryear{{Nait Abdallah}}{1991}]{areski91a}
A.~{Nait Abdallah}.
\newblock Kernel knowledge versus belt knowledge in default reasoning: a
  logical approach.
\newblock In Dehne F., Fiala F., and Koczkodaj W., editors, {\em International
  Conference on Computing and Information}, pages 675--686. Springer LNCS 497,
  1991.

\bibitem[\protect\citeauthoryear{Parsons}{1980}]{parsons80}
T.~Parsons.
\newblock {\em Nonexistent Objects}.
\newblock Yale University Press, New Haven, CT, 1980.

\bibitem[\protect\citeauthoryear{Quine}{1947}]{quine47}
W.V.O. Quine.
\newblock {\em Mathematical Logic}.
\newblock Cambridge, 1947.

\bibitem[\protect\citeauthoryear{Quine}{1949}]{quine49}
W.V.O. Quine.
\newblock Designation and existence.
\newblock In Feigl H. and Sellars W., editors, {\em Readings in Philosophical
  Analysis}, pages 44--51. Appleton-Century-Croft, New York, 1949.

\bibitem[\protect\citeauthoryear{Rapaport}{1985}]{rapaport85a}
W.J. Rapaport.
\newblock Meinongian semantics for propositional semantic networks.
\newblock In {\em Proceedings 23rd Annual Meeting of the Association for
  Computational Linguistics}, pages 43--48, 1985.

\bibitem[\protect\citeauthoryear{Reiter}{1980}]{reiter80}
R.~Reiter.
\newblock A logic for default reasoning.
\newblock {\em Artificial Intelligence}, 13:81--132, 1980.

\bibitem[\protect\citeauthoryear{Reiter}{1990}]{reiter90}
E.~Reiter.
\newblock The computational complexity of avoiding conversational implicatures.
\newblock In {\em Proceedings 28th Annual Meeting of the Association for
  Computational Linguistics}, pages 97--104, 1990.

\bibitem[\protect\citeauthoryear{Russell}{1905}]{russell05}
B.~Russell.
\newblock On denoting.
\newblock {\em Mind n.s.}, 14:479--493, 1905.
\newblock reprinted in: Feigl H. and Sellars W. editors, {\em Readings in
  Philosophical Analysis}, pages 103--115. Appleton-Century-Croft, New York,
  1949.

\bibitem[\protect\citeauthoryear{Sandt}{1992}]{sandt92}
{R.A. van der} Sandt.
\newblock Presupposition projection as anaphora resolution.
\newblock {\em Journal of Semantics}, 9:333--377, 1992.

\bibitem[\protect\citeauthoryear{Siskind and McAllester}{1993a}]{siskind93b}
J.M. Siskind and D.A. McAllester.
\newblock Nondeterministic Lisp as a substrate for constraint logic
  programming.
\newblock In {\em Proceedings of the Twelfth National Conference on Artificial
  Intelligence}, pages 133--138, 1993.

\bibitem[\protect\citeauthoryear{Siskind and McAllester}{1993b}]{siskind93a}
J.M. Siskind and D.A. McAllester.
\newblock Screamer: A portable efficient implementation of nondeterministic
  Common Lisp.
\newblock Technical Report IRCS-93-03, University of Pennsylvania, Institute
  for Research in Cognitive Science, July 1 1993.

\bibitem[\protect\citeauthoryear{Siskind}{1991}]{siskind91}
J.M. Siskind.
\newblock Screaming yellow zonkers.
\newblock Technical report, M.I.T. Artificial Intelligence Laboratory,
  September 1991.

\bibitem[\protect\citeauthoryear{Smullyan}{1970}]{smullyan70}
R.M. Smullyan.
\newblock {\em First-Order Logic}.
\newblock Springer-Verlag, New York, 1970.

\bibitem[\protect\citeauthoryear{Soames}{1979}]{soames79}
S.~Soames.
\newblock A projection problem for speaker presuppositions.
\newblock {\em Linguistic Inquiry}, 10(4):623--666, Fall 1979.

\bibitem[\protect\citeauthoryear{Soames}{1982}]{soames82}
S.~Soames.
\newblock How presuppositions are inherited: A solution to the projection
  problem.
\newblock {\em Linguistic Inquiry}, 13(3):483--545, Summer 1982.

\bibitem[\protect\citeauthoryear{Soames}{1989}]{soames89}
S.~Soames.
\newblock Presupposition.
\newblock In Gabbay D. and Guenthner F., editors, {\em Handbook of
  Philosophical Logic --- Topics in the Philosophy of Language}, pages
  553--617. D. Reidel Publishing Company, 1989.

\bibitem[\protect\citeauthoryear{Stalnaker}{1973}]{stalnaker73}
R.C. Stalnaker.
\newblock Presuppositions.
\newblock {\em Journal of Philosophical Logic}, 2:447--457, 1973.

\bibitem[\protect\citeauthoryear{Strawson}{1950}]{strawson50}
P.F. Strawson.
\newblock On referring.
\newblock {\em Mind}, 59:320--344, 1950.

\bibitem[\protect\citeauthoryear{Velman}{1981}]{veltman81}
F.~Velman.
\newblock Data semantics.
\newblock In Groenendijk~J.A.G. et. al., editor, {\em Formal Methods in the
  Study of Language}. Mathematisch Centrum, Amsterdam, 1981.

\bibitem[\protect\citeauthoryear{Walker}{1992}]{walker92}
M.A. Walker.
\newblock Redundancy in collaborative dialogue.
\newblock In {\em Proceedings of the International Conference on Computational
  Linguistics, COLING}, pages 345--351, 1992.

\bibitem[\protect\citeauthoryear{Weischedel}{1979}]{weischedel79}
R.M. Weischedel.
\newblock A new semantic computation while parsing: Presupposition and
  entailment.
\newblock In Oh~C.K. and Dinneen D.A, editors, {\em Syntax and Semantics,
  Presupposition}, volume~11, pages 155--182. Academic Press, 1979.

\bibitem[\protect\citeauthoryear{Wilson and Sperber}{1979}]{wilson79}
D.~Wilson and D.~Sperber.
\newblock Ordered entailments: An alternative to presuppositional theories.
\newblock In Oh~C.K. and Dinneen D.A, editors, {\em Syntax and Semantics,
  Presupposition}, volume~11, pages 299--324. Academic Press, 1979.

\bibitem[\protect\citeauthoryear{Zeevat}{1992}]{zeevat92}
H.~Zeevat.
\newblock Presupposition and accommodation in update semantics.
\newblock {\em Journal of Semantics}, 9:379--412, 1992.

\end{thebibliography}

\end{document}